\documentclass[runningheads]{llncs}
\usepackage{graphicx}
\usepackage{comment}
\usepackage{amsmath,amssymb} % define this before the line numbering.
\usepackage{color}
\usepackage[utf8]{inputenc} % allow utf-8 input
\usepackage[T1]{fontenc}    % use 8-bit T1 fonts
\usepackage{booktabs}       % professional-quality tables
\usepackage{amsfonts}       % blackboard math symbols
\usepackage{nicefrac}       % compact symbols for 1/2, etc.
\usepackage{microtype}      % microtypography
\usepackage{times}
\usepackage{epsfig}
\usepackage{subfigure}
\usepackage{multirow}
\usepackage[para]{footmisc}
\usepackage{wrapfig,lipsum,booktabs}

\usepackage{dsfont}
\usepackage{algorithm}
\usepackage{algpseudocode}

\usepackage{here}
\usepackage{multirow}
\usepackage{bigints}
\usepackage{xargs}                      % Use more than one optional parameter in a new commands
\usepackage{enumerate}

\usepackage{ mathrsfs }
\usepackage{kbordermatrix}% http:/

\if 0

\newtheorem{lemma}{Lemma}

\theoremstyle{definition}
\newtheorem{definition}{Definition}

\newtheorem{remark}{Remark}

\fi

\newcommand{\mB}{\mathcal{B}}
\newcommand{\mC}{\mathcal{C}}

\newcommand{\mS}{\mathcal{S}}

\newcommand{\mM}{\mathcal{M}}

\newcommand{\mL}{\mathcal{L}}
\newcommand{\mK}{\mathcal{K}}
\newcommand{\mN}{\mathcal{N}}

\renewcommand{\tilde}{\widetilde}

\newcommand{\argmin}{\operatornamewithlimits{argmin}}

\newcommand\restr[2]{{% we make the whole thing an ordinary symbol
		\left.\kern-\nulldelimiterspace % automatically resize the bar with \right
		#1 % the function
		\vphantom{\big|} % pretend it's a little taller at normal size
		\right|_{#2} % this is the delimiter
}}

\usepackage{hyperref}
\usepackage{array}
\newcolumntype{N}{>{\centering\arraybackslash}m{.5in}}
\newcolumntype{G}{>{\centering\arraybackslash}m{\dimexpr2in+6\tabcolsep}}
\newcommand{\rarrow}[1]{\buildrel #1 \over \longrightarrow}

\begin{document}
% \linenumbers
\pagestyle{headings}
\mainmatter
\def\ECCVSubNumber{6515}  % Insert your submission number here

\title{Simplicial Complex based Point Correspondence between Images warped onto Manifolds  } % Replace with your title

%\begin{comment}
\titlerunning{ECCV-20 submission ID \ECCVSubNumber} 
\authorrunning{ECCV-20 submission ID \ECCVSubNumber} 
\author{Anonymous ECCV submission}
\institute{Paper ID \ECCVSubNumber}
%\end{comment}
%******************


\titlerunning{Simplicial Complex based Point Correspondence between warped Images}
\author{Charu Sharma\inst{} \and
Manohar Kaul\inst{}}
\authorrunning{Charu Sharma and Manohar Kaul}
\institute{Indian Institute of Technology Hyderabad, India
\email{\{cs16resch11007,mkaul\}@iith.ac.in}}

%******************
\maketitle

\begin{abstract}
Recent increase in the availability of warped images projected onto a manifold (e.g., omnidirectional spherical images), coupled with the success of higher-order assignment methods, has sparked an interest in the search for improved higher-order matching algorithms on warped images due to projection. 
Although currently, several existing methods ``flatten" such 3D images to use planar graph / hypergraph matching methods, they still suffer from severe distortions and other undesired artifacts, which result in inaccurate matching. Alternatively, current planar methods cannot be trivially extended to effectively match points on images warped onto manifolds.
Hence, matching on these warped images persists as a formidable challenge.
In this paper, we pose the assignment problem as finding a bijective map between two graph induced simplicial complexes, which are higher-order analogues of graphs.
%, whose vertices are landmark points embedded on the smooth surface image and edges are geodesics on the smooth surface connecting these vertices.
We propose a constrained quadratic assignment problem (QAP) that matches each $p$-skeleton of the simplicial 
complexes, iterating from the highest to the lowest dimension.   
The accuracy and robustness of our approach are illustrated on both synthetic and real-world spherical / warped (projected) images with known ground-truth correspondences. 
We significantly outperform existing state-of-the-art spherical matching methods on a diverse set of datasets.

\end{abstract}

\section{Introduction}

There exists a longstanding line of research on finding bijective correspondences (i.e., assignments / matchings\footnote{\emph{assignment} and \emph{matching} are used interchangeably in this paper.}) 
between two sets of visual features. Notable applications include stereo matching~\cite{goesele2007multi}, structure from motion (SfM)~\cite{szeliski2010computer}, and image registration~\cite{shen2002hammer}, to name a few.
Traditionally, when matching points between multiple images of a fixed environment from various viewpoints, most approaches recover matchings and relative camera geometry (e.g. fundamental matrix) using a robust technique such as RANSAC~\cite{hartley2003multiple}. On the other hand, when matching between different instances of the same category, 
graph matching methods~\cite{ZhouD16} using \emph{unary}
and \emph{pairwise} constraints have been successfully utilized. More recently, graph matching has been subsumed by \emph{hypergraph matching} using 
\emph{higher-order} constraints~\cite{duchenne2011tensor,lzss_iccv11}. An important appeal of higher-order matching methods is their ability to coherently match compact local geometric features from the source space to similar compact regions in the target space, despite the presence of noise, outliers, and incomplete data, 
thus achieving accurate  matches that are also \emph{local structure-preserving} in nature.

%Traditionally, this assignment problem has been cast as a \emph{graph matching} problem using \emph{unary}
%and \emph{pairwise} constraints~\cite{ZhouD16}, later subsumed by \emph{hypergraph matching} using 
%\emph{higher-order} constraints~\cite{duchenne2011tensor,lzss_iccv11}. An important appeal of higher-order matching methods is their ability to coherently match compact local geometric features from the source space to similar compact regions in the target space, thus achieving accurate  matches that are also \emph{structure-preserving} in nature.

The recent proliferation of spherical images (e.g., omnidirectional and panoramic images captured from cameras mounted on drones and autonomous vehicles) and more generally, images warped onto manifolds with non-trivial curvatures,
%\footnote{We restrict our attention to smooth surfaces with zero genus for ease of geodesic computation and ruling out degenerate surfaces.}, 
has sparked a heightened interest in assignment algorithms on such datasets due to the challenges they present in terms of curvature, both uniform and non-uniform~\cite{starck2005spherical,kaminsky2009alignment,zeng20173dmatch,yang2013go}. 
Although assignment problems have been well studied for decades in computer vision, a majority of the work has only
focused on matching points between \emph{planar (flat) images}. Therefore, matching points on images with warping transformations which fall into the category of projective parametric models remains a challenging task, mainly due to the introduction of undesirable artifacts like severe distortions in pairwise distances between landmark points, non-linear distortions in local geometries, noise, illumination, blur, and occlusions~\cite{azevedo2019,coors2018}, on flattening. 

When dealing with matchings on curved geometries, primarily two types of methods are employed. 
Some putative matchings are computed to estimate a \emph{fundamental matrix}~\cite{hartley2003multiple,Cyganek2007} that captures the \emph{epipolar geometry} of the 3D image. \emph{Stereo rectification}~\cite{bradski2008learning} uses this fundamental matrix to re-project the two images on the same flat plane with row images aligned in parallel, followed by a re-matching to improve matching accuracy.
Alternatively, \emph{geometric alignment} on the fundamental matrix is used to \emph{verify} and distinguish 
\emph{inliers} from \emph{outliers}, so that outliers can be pruned post matching to further boost accuracy~\cite{szeliski2010computer}.
Elements warped on the curved manifold cannot be metrically sampled in such methods and hence severe distortions are introduced~\cite{Colombo2004}, 
which is also consistent with the findings in our empirical studies.

\textbf{Applications} An interesting and noteworthy application of higher-order matching on spherical images arises in the area of biomedical imaging, especially in \emph{retinal imaging} using \emph{optical coherence tomography} (OCT). To investigate a wider \emph{field of view}, 3D \emph{fundus} images of the eye 
are captured, matched, and ``stitched" together to form an \emph{OCT montage}~\cite{Li2011,Pauly2008}. 
This matching operation must additionally preserve \emph{regions of interest} such as the \emph{optic cup / disc, 
fovea, macula, vessels, and microaneurysms}, to name a few~\cite{sengupta2018}. In addition to the standard noise, 
occlusion, and artifacts in these OCT fundus images, the data also suffers from data shifts due to axial eye motions 
and unpredictability between \emph{eye positions} and \emph{instrument alignment} across various 
scans~\cite{Li2011}. Therefore, OCT datasets cannot easily be matched using rigid 3D transformations.
Such images are not limited to merely spherical ones, but also arise in more general warped images due to projection. For instance, 3D 
sonograms depict the cervix as a \emph{conic frustum (truncated cone)}~\cite{ahmed2017} and clustered nanofluid 
microflow patterns in elastic micro-tubes are tracked via matching between \emph{cylindrical} images in a 
time-lapse~\cite{Sung2015}.  

\textbf{Related work}
Previous works on spherical matching mainly focus on producing good feature descriptors and can broadly be classified as \emph{planar} and \emph{spherical} feature extractors.
\emph{Planar feature extractors} like
SIFT~\cite{lowe2004distinctive}, SURF~\cite{bay2006surf}, ORB~\cite{rublee2011orb}, BRISK~\cite{leutenegger2011brisk}, and FREAK~\cite{alahi2012freak} extract descriptors either on an \emph{unwrapped 
	equirectangular version of an omnidirectional image} or directly on the 2D (flat) representation of the 
spherical image.
In contrast, \emph{spherical feature extractors} reduce the distortion due to planar embeddings, by 
taking into account the underlying geodesic distances on the sphere while computing descriptors. 
These methods include ones based on spectral analysis, spherical harmonics (SIFTS~\cite{cruz2012scale}), and projection on geodesic grids (SPHORB~\cite{zhao2015sphorb}, BRISKS~\cite{guan2017brisks}). 

\textbf{Our method}
In this paper, we focus on exploiting the intrinsic higher-order geometric relationships between landmark points on images warped onto curved manifolds. We capture these higher-order connections by constructing a combinatorial topological structure (simplicial complex) which is induced by a graph, whose \emph{vertices} are the landmark points embedded on the warped image and whose \emph{edges} are \emph{geodesic curves} between selected vertex pairs. Next, we pose the assignment problem as a multi-dimensional \emph{quadratic assignment problem (QAP)} between two graph-induced simplicial complexes. 

\textbf{Our contributions} 
(i) To the best of our knowledge, we are the first to propose higher-order matching of  landmark points on warped images projected onto curved manifolds (including for example spheres, ellipsoids, cylinders, and cones).
(ii) In an attempt to break away from other works which solely focus on flat or spherical images, we propose a novel graph induced simplicial complex that efficiently captures higher order structures in a succinct manner, considering the inherent properties of the underlying manifold on which the landmark points are embedded. 
(iii) We uniquely formulate the assignment problem as a multi-dimensional combinatorial matching between two graph induced simplicial complexes, propose a novel algorithm to solve it, and analyze the time-complexity of our algorithm.
(iv) Finally, to illustrate the robustness of our proposed method, we perform extensive experiments 
by comparing to planar matching methods, both \emph{existing} and \emph{extended by us} as \emph{naive baselines} for matching of landmarks on manifolds.
%covering \emph{multimodal matching} and \emph{ablative studies}.
% between \emph{spherical-spherical, spherical-planar, unwrapped spherical-unwrapped 
%spherical}, and \emph{unwrapped spherical-planar} images. 
We compare our method against existing \emph{graph matching} and \emph{spherical matching} (both boosted using \emph{rectification} 
and \emph{verification} 
techniques)~\cite{leutenegger2011brisk,rublee2011orb,zhao2015sphorb,ZhouD16,zhou2013deformable,sharma2018solving,duchenne2011tensor} on warped images and interestingly observe that 
not only does our method significantly outperform these matching methods on warped images onto curved manifolds (with up to $49.7 \%$ matching error reduction), but it also outperforms existing planar matching algorithms on ``flat" planar images too (with up to $42.2\%$ matching error reduction), due to the ability to naturally capture higher-order relationships by the simplicial complex.

\section{Preliminaries}
\label{sec:prelim}
In this section, we introduce our notation and provide the necessary background for our 
higher-order assignment algorithm on curved manifolds. 
We begin by introducing certain standard definitions followed by our problem definition.

Let $\mM$ denote a curved manifold. %, i.e., with zero genus and no boundaries.
On a plane, the shortest distance between any two points is a straight line, i.e., a 
curve whose derivative to its tangent vectors is zero. We extend this notion of a ``straight line" to curved manifolds by defining the shortest path (on $\mM$) between its endpoints $u$ and $v$ placed on $\mM$, as a geodesic curve $\gamma(u,v)$.

\paragraph{\textbf{Simplicial complex}} We begin by providing some general definitions before we can formally define a simplicial complex. More background can be found in~\cite{munkres1984}.

Given a set $V=\{v_0,\dots,v_n\}$ of $(n+1)$ affinely independent points in $\mathds{R}^{n+1}$, a $n$-\emph{dimensional simplex} (also called $n$-simplex) $\sigma^{(n)}$ with \emph{vertices} $V$ is the \emph{convex hull} of $V$, i.e., more formally
\[
\sigma^{(n)} = \left\{(k_0,k_1, \dots, k_n) \in  \mathds{R}^{n+1}  \mid \sum_{i=0}^{n} k_i=1, \text{  }k_i \geq 0 \text{ }\forall i  \right\}
\]
The dimension of $n$-simplex $\sigma^{(n)}$ is denoted by $dim(\sigma^{(n)})$.
For example, a \emph{point / vertex} ($0$-simplex), an \emph{edge} ($1$-simplex), and a \emph{triangle} ($2$-simplex) 
are represented as $\sigma^{(0)}$, $\sigma^{(1)}$, and $\sigma^{(2)}$, respectively.

For $0 \leq i \leq n$, the $i$-th \emph{facet} $f_i$ of the $n$-simplex $\sigma^{(n)}$ is the 
$(n-1)$-simplex $\sigma^{(n-1)}$, whose vertices are those underlying $\sigma^{(n)}$, except the $i$-th vertex. 
For example, a $2$-simplex (triangle) has three $1$-simplices (edges) as \emph{facets}.
The \emph{boundary} $\partial \sigma^{(n)}$ of the $n$-simplex $\sigma^{(n)}$ is $\bigcup_{i=0}^n f_i$.

Finally, a \emph{simplicial complex} $\mK$ is a set of simplices that satisfy the following conditions: 

(i). Any face of a simplex in $\mK$ is a simplex in $\mK$ and
(ii). Intersection of distinct simplices $\sigma_i$ and $\sigma_j$ in $\mK$, is a \emph{common face} of 
both $\sigma_i$ and $\sigma_j$\footnote{For ease of notation, we drop the dimension superscript and index subscript for a simplex when it is understood from context. }.

The $p$-\emph{skeleton} $\mK^{(p)} \subset \mK$ is formed by the set of $k$-simplices
$\sigma^{(k)}$, where $k \leq p$. 
Additionally, we denote by $\mK_k$ the set of $k$-simplices in $\mK$. 
The dimension $dim(\mK)$ of a simplicial complex $\mK$ is the maximum of the dimensions 
of its constituent simplices.

\textbf{Problem definition}
Our problem consists of first constructing \emph{geometric simplicial complexes} 
between landmark points given on curved manifolds, followed by finding an 
optimal (i.e., least cost) assignment between a pair of such geometric simplicial complexes by 
matching simplices of the same dimension, one dimension at a time.  

More formally, Let $P$ and $P'$ denote two sets of \emph{landmark points} on curved manifolds $\mM$ and $\mM'$, respectively. We construct \emph{geometric simplicial complexes} $\mK$ and 
$\mK'$ whose set of vertices ($0$-simplices) are $P$ and $P'$. The edges/arcs ($1$-simplices) in $\mK$ and $\mK'$ are given by geodesics between select few pairs of vertices, from their corresponding vertex sets.

Given two simplicial complexes $\mK$ and $\mK'$, we assume without loss of generality, that the number of simplices of each corresponding dimension are equal in both complexes.
Then, our goal is to find a set of $h$ bijective \emph{matching functions}
$\{ m_k  \}_{k=0}^h : \mK \rarrow{} \mK' $ that match the set of $k$-simplices in $\mK$ (i.e., $\mK_k$) to $k$-simplices in $\mK'$ (i.e., $\mK'_k$), for dimensions $k=0 \dots h$, to minimize the overall objective function
\begin{align}
\argmin_{m_0,\dots,m_h} \sum_{k=0}^{h} \sum_{i=1}^{|\mK_k|} c(\sigma^{(k)}_i, m_k(\sigma^{(k)}_i ))
\end{align}
where $c(\cdot,\cdot)$ is the \emph{geometric matching cost} between a $k$-simplex 
$\sigma^{(k)}$ in $\mK$ to a $k$-simplex $m(\sigma^{(k)})$ in $\mK'$ and 
simplicial complex dimension $h=\min (dim(\mK),dim(\mK'))$. 

Unlike formulations proposed in graph matching methods~\cite{ZhouD16}, where only node and pairwise geometric relations are considered, our \emph{combinatorial optimization} formulation takes into consideration higher-order geometric constraints, which better excludes ambiguous matchings.

In subsequent sections, we show how we construct such geometric simplicial complexes from 
the landmark points on curved manifolds (Section~\ref{sec:sor}), followed by a detailed explanation of our assignment algorithm (Section~\ref{sec:assignment}).

\section{Building a Simplicial Complex on a Curved Manifold}
\label{sec:sor}
In this section, inspired by the work of Dey et. al.~\cite{Dey2015}, we similarly construct a \emph{graph-induced} simplicial complex, which is built upon a graph connecting the landmark points. We begin by describing the process of constructing the \emph{underlying graph}.

\textbf{Graph construction} Let $(P,g)$ denote the set of landmark points $P$ with a metric $g$ that denotes the geodesic distance between a pair of points on $\mM$.
Additionally, let the $k$-neighborhood $\mN_k(u)$ denote the set of $k$ nearest neighbors of landmark point $u \in P$ (inclusive of $u$) on manifold $\mM$ according to the geodesic metric $g$.

Considering all ordered pairs $(u,v)$, where $u,v \in P$, an undirected \emph{edge/arc} is introduced between points $u$ and $v$, when their corresponding $k$-neighborhoods $\mN_k(u)$ and $\mN_k(v)$ have a non-empty intersection, i.e., $\mN_k(u) \cap \mN_k(v) \neq \emptyset $. All such edges are collected into a set denoted by $E$.
This completes the construction of our underlying graph $G=(P,E)$. 
Observe that the vertex set (landmarks) $P$ 
form the $0$-skeleton $\mK^{(0)}(G)$ and 
the sets $E$ and $P$ together form the $1$-skeleton $\mK^{(1)}(G)$, 
of our graph-induced simplicial complex that we will denote by $\mK(G)$. 

Recall that a $n$-clique in a graph is a complete subgraph between $n$ vertices, i.e., it consists of $n$ vertices and 
$n \choose 2$ edges. 

\textbf{Graph-induced complex} $\mK(G)$ is defined as the simplicial complex where a $n$-simplex 
$\sigma^{(n)} = \{p_1, p_2, \dots, p_{n+1}  \}$ is in $\mK(G)$, if and only if there exists a $(n+1)$-clique $\{p_1, p_2, \dots, p_{n+1}  \} \subseteq P$ in the underlying graph $G=(P,E)$. In words, the \emph{cliques} of the underlying graph $G=(P,E)$ form the \emph{simplices} in $\mK(G)$ because cliques satisfy both conditions of being a simplicial complex (which can be trivially verified). 
In order to be used in our assignment algorithm, we must represent the graph-induced simplicial complex $\mK(G)$ as a set of \emph{boundary} matrices, which we present next.

\noindent\textbf{Matrix representation of $\mK(G)$}: Given $\mK(G)$ and its $p$-skeleton 
$\mK^{(p)}(G)$ that contains cliques upto size $p+1$, we represent it as a \emph{boundary matrix} $M_p \in \mathds{Z}^{n \times m}$ defined as 
\renewcommand{\kbldelim}{(}% Left delimiter
\renewcommand{\kbrdelim}{)}% Right delimiter
\[
M_p = \kbordermatrix{
	& \sigma^{(p)}_1  & \dots & \sigma^{(p)}_m \\
	\tau^{(p-1)}_1  & a_{11}  & \dots & a_{1m} \\
%	\tau^{(p-1)}_n & a_{21} & a_{22} & \dots & a_{2m} \\
	\vdots & \vdots  & \ddots & \vdots \\
	\tau^{(p-1)}_n & a_{n1} & \dots & a_{nm} 
}
\]
where $a_{ij} = 1$ if and only if the $i$-th $(p-1)$-simplex $\tau^{(p-1)}_i$ is a \emph{facet} of the $j$-th 
$p$-simplex $\sigma^{(p)}_j$, otherwise $a_{ij}=0$. 
Then, the boundary of a $j$-th $p$-simplex is given by $\partial_p \sigma^{(p)}_j = \sum_{i=1}^{n} a_{ij} \tau^{(p-1)}_i$.

Observe that the $p$-th boundary matrix $M_p$ captures all possible relationships between 
$p$-simplices and their $(p-1)$-simplex boundaries (or facets). 
Boundary matrix $M_p$ is made for each $p$-skeleton and therefore $\mK(G)$ is expressed as a 
set of boundary matrices $\{ M_p  \}_{p=1}^h$, where $h = dim(\mK(G))$.

\begin{remark}
	Our underlying graph $G$ already contains as a \emph{subgraph} a simple $k$-nearest neighbor graph which is constructed by introducing edges between a vertex in question and its $k$ nearest neighbors. Therefore, our underlying graph $G$ has more edges and thus has a higher likelihood to form higher-order relations between vertices. 
	On the other hand, while the Delaunay triangulation is simple to compute and is a good vehicle for extracting topology of sampled spaces, its size becomes prohibitively large for reasonable computations and thus adversely affects the QAP matching algorithm. 

	In summary, our underlying graph $G$ which is inspired by the \emph{Vietoris-Rips} complex construction provides a good \emph{proximity structure}, which is neither \emph{too sparse} (like simple $k$-NN graphs) or \emph{too dense} (like Delaunay triangulated graphs) and encodes useful higher-order information about local relations of landmark points in $P$.
\end{remark}

%----------------------------------------------------------------------------------------------------------------
\section{Assignment Algorithm}
\label{sec:assignment}
Recall our problem definition (Section~\ref{sec:prelim}) of trying to find a set of assignments / matching functions between two graph-induced simplicial complexes $\mK(G)$ and $\mK(G')$. Here, we outline the details of our 
assignment algorithm.

%\textbf{Our method}

Given a boundary matrix $M_p \in \mathds{Z}^{n \times m}$ that represents a $p$-skeleton $\mK^{(p)}(G)$, we first
capture the \emph{geodesic neighborhood geometry} of simplices in $M_p$.
We begin by defining an \emph{adjacency operator} $\sim$ between two simplices followed by a definition of a \emph{neighborhood of a simplex}. This neighborhood of a simplex is then elegantly captured by \emph{affine weight vectors}, which are later used in the matching algorithm. 
\begin{definition}[adjacency relation]
	Given two simplices $\sigma^{(d)}$ and $\sigma'^{(d')}$, each of arbitrary dimension $d$ and $d'$, we consider them to be \emph{adjacent}
	to one another if and only if they share a \emph{common simplex}. 
	We denote this adjacency relation by $\sigma^{(d)} \sim \sigma'^{(d')}$. The dimension of the common simplex can take values from $0$ to $\min( d,d')$.
\end{definition}
For example, two $2$-simplices / triangles $\sigma^{(2)}$ and $\sigma'^{(2)}$ could either be connected at a common
$0$-simplex / vertex or share a common $1$-simplex / edge; both cases would result in the simplices being \emph{adjacent}, i.e., $\sigma^{(2)} \sim \sigma'^{(2)}$.

\textbf{Simplex neighborhood} The boundary matrix $M_p$'s columns encode $p$-simplices $\sigma^{(p)}_1, \dots, \sigma^{(p)}_m$ and its 
rows encode $(p-1)$-simplices $\tau^{(p-1)}_1, \dots, \tau^{(p-1)}_n$. 
The computation of the neighborhood $\mathfrak{N}(\cdot)$ for $p$-simplices and $(p-1)$-simplices differ slightly.
The neighborhood of a $p$-simplex consists of $p$-simplices (same dimension) and $(p-1)$-simplices (one dimension lower) that are adjacent to it. While, the neighborhood of a $(p-1)$-simplex consists of $(p-1)$-simplices (same dimension) and $p$-simplices (one dimension higher) that are adjacent to it.
More formally, the neighborhood of the $i$-th $p$-simplex $\sigma^{(p)}_i$ is
\begin{align}
\mathfrak{N}( \sigma^{(p)}_i) = \lbrace \sigma^{(p)}_j \mid \sigma^{(p)}_j \sim \sigma^{(p)}_i  \rbrace 
\cup \lbrace  \tau^{(p-1)}_j \mid   \tau^{(p-1)}_j  \sim \sigma^{(p)}_i     \rbrace \nonumber
\end{align}
and the neighborhood of the $i$-th $(p-1)$-simplex $\tau^{(p-1)}_i$ is
\begin{align}
\nonumber
\mathfrak{N}( \tau^{(p-1)}_i) =& \lbrace \tau^{(p-1)}_j \mid \tau^{(p-1)}_j \sim \tau^{(p-1)}_i \rbrace \\ \nonumber
&\cup \lbrace  \sigma^{(p)}_j \mid   \sigma^{(p)}_j  \sim \tau^{(p-1)}_i     \rbrace \nonumber
\end{align}
Such neighborhoods are computed for all the $p$- and $(p-1)$-simplices in $M_p$, where $i \neq j$. 

\textbf{Affine weight vectors} For a $p$-simplex $\sigma^{(p)}$, let $ \mB( \sigma^{(p)} )$ denote the set of all the \emph{barycenters} 
$\{b_1,\dots,b_{|\mathfrak{N}( \sigma^{(p)})|}  \} $
of the simplices in the neighborhood $\mathfrak{N}( \sigma^{(p)})$.  
Then, $\sigma^{(p)}$ is represented as an \emph{affine combination} of the barycenters in $\mB( \sigma^{(p)} )$, i.e., \\
$\sum_{i=1}^{ | \mB( \sigma^{(p)}) |  } \alpha_i b_i $, where $\sum_{i=1}^{| \mB( \sigma^{(p)}) |} \alpha_i = 1$ (i.e., weights $\alpha_i$'s must sum to $1$).
Therefore, $\sigma^{(p)}$ is expressed as an  \emph{affine weight vector} $\alpha(\sigma^{(p)})$ of dimension $(n+m)$, with $ | \mB( \sigma^{(p)}) | $ 
positions corresponding to $\mathfrak{N}( \sigma^{(p)})$ filled with non-empty affine weights and the rest set to zero.
Such an affine weight vector is computed for every simplex of dimension $p$ and $(p-1)$ contained in $M_p$.

Following a similar method as shown in~\cite{li2013object} for the choice of a \emph{locally invariant affine weight vector}, we extend their method to simplices and among all possible affine representations of a simplex, we chose to use \emph{least squares} 
to guarantee minimal error under L2-norm, and furthermore it assigns
non-zero weights to each of its adjacent simplex barycenters,
thereby better capturing the local geometric properties in its neighborhood.
\begin{remark}
	The affine weight vectors act as \emph{locally affine invariant descriptors} that can handle complex and natural transformations of the underlying manifold $\mM$. Additionally, it allows for much fewer variables and can be much more easily linearized in the subsequent QAP formulation. Furthermore, the inclusion of barycenters from neighborhoods of each simplex act as higher-order geometric constraints that easily excludes ambiguous matchings.
	In comparison, simple matching models that rely on just a distance matrix with pairwise geodesic distances on the manifold are not invariant to local and global affine transformations and completely disregard higher-order relationships.
\end{remark}

\textbf{Cost matrix construction} Next, we describe the construction of a \emph{cost matrix} that is needed to compute assignments between $M_p \in \mathds{Z}^{n \times m}$ and $M'_p \in \mathds{Z}^{n' \times m'}$. 
We begin by constructing two cost matrices $\mC^{(p-1)} \in \mathds{R}^{n \times n'}$ and 
$\mC^{(p)} \in \mathds{R}^{m \times m'}$ to measure the Euclidean distance between the affine weight vectors of 
$(p-1)$-simplices and the Euclidean distance between the affine weight vectors of $p$-simplices, respectively.

More specifically, 
$c_{ii'}^{(p-1)} = \lVert \alpha(\tau^{(p-1)}_i) - \alpha(\tau^{(p-1)}_{i'} ) \lVert_2$, 
measures the 
Euclidean distance between the affine weight vectors of the $i$-th $(p-1)$-simplex of $M_p$ 
and the $i'$-th $(p-1)$-simplex of $M'_p$, while 
$c_{kk'}^{(p)} = \lVert \alpha(\sigma^{(p)}_k) - \alpha(\sigma^{(p)}_{k'}) \lVert_2$, 
measures the Euclidean distance between the affine weight vectors of the $k$-th $p$-simplex of $M_p$ 
and the $k'$-th $p$-simplex of $M'_p$. 

Similar to the \emph{affinity matrix} construction in~\cite{ZhouD16},
we combine both the cost matrices in a single geodesic-cost matrix $\mL^{(p)} = ( l_{  ii',jj' } ) \in \mathds{R}^{nn' \times mm'}$ as
\[ 
l^{(p)}_{ii',jj'} =
\begin{cases} 
c_{ii'}^{(p-1)}  & i=j \text{ , } i'=j'  \\
c_{kk'}^{(p)}   & i \neq j \text{ , } i ' \neq j'  \text{ , } a_{ik} a_{jk} a'_{i'k'} a'_{j'k'} = 1  \\
0 & \text{otherwise }
\end{cases}
\]
The \emph{diagonal} and \emph{off-diagonal} entries of matrix $\mL^{(p)} $ capture the \emph{Euclidean distances between the affine weight vectors of $(p-1)$-simplices} and the \emph{Euclidean distances between the 
affine weight vectors of $p$-simplices}, respectively. Therefore, our QAP can now be formulated as 
\begin{equation}
\label{eq:opt}
\begin{aligned}
& \argmin_{X_1,\dots,X_h}
& &   \sum_{p=1}^{h} vec(X_p)^T \mL^{(p)} vec(X_p)  \\
& \text{subject to}
& & \forall p \leq h, \mathds{1}^T X_p = \mathds{1},  X_p^T \mathds{1} = \mathds{1}
\end{aligned}
\end{equation}
% Figure for matching spheres.

where $X_p$ is a permutation matrix and $vec(X_p)$ is it's vector representation.
Our solution to Equation~\ref{eq:opt} is concisely outlined in Algorithm~\ref{alg:matching}. As we solve a QAP from highest to lowest dimension $p$-skeleton, 
we track the $(p-1)$-simplices whose matchings are \emph{induced} by higher order simplex matches.
On finding $(p-1)$-simplices that have the lowest cost and cannot be improved by solving a lower level 
QAP, we eliminate such simplices, causing the size of the matrix to shrink in subsequent iterations, leading to substantial speedups. Also, we use a \emph{spectral relaxation} proposed by Lordeneu et. al.~\cite{leordeanu2005spectral} 
to solve our QAP efficiently.
%\begin{comment}

\begin{algorithm}[tbp]
	\caption{Matching graph induced simplicial complexes}
	\label{alg:matching}
	\begin{flushleft}
		\textbf{Input:} $\mK(G) = \{  M_p \}_{p=1}^h$ and
		$\mK(G') = \{ M'_p   \}_{p=1}^h$
	\end{flushleft}
	\begin{algorithmic}[1]
		%\REPEAT
		\For{ $p = h \dots 1$}
%		\STATE $(m,n)$ := (\#rows, \#columns) in $M_p$
		% Get affine vectors
	
		\State Build cost matrix $\mL^{(p)}$ for $M_p$ and $M'_p$
	(*account for $\mL^{(p-1)}$)

%			\FOR{ $i = 1 \dots m$} 
%				\STATE $\alpha( \sigma^{(i)} ) := \left[  \underbrace{\alpha_1 , \dots, \alpha_n }_{ (p-1)\text{-simplex} } ,  
%				\underbrace{\alpha_{m+1}^{(p)} , \dots, \alpha_{2m}^{(p)}}_{ p\text{-simplex}}    \right]$
%				\STATE $\alpha := \alpha \cup \{ \alpha_i \}$  \COMMENT{affine weight vectors}			
%			\ENDFOR
		% Repeat steps
%		\STATE Repeat steps $3$--$6$ on $M'_p$ to get $\alpha'$.
		% Build a cost matrix
%		\STATE Build cost matrix $C^{(p)}$ from $\alpha$ and $\alpha'$ (*account for $\mL^{(p-1)}$)
		\State $X^{*}_p := $ \textbf{Solve QAP }($M_p,M'_p,\mL^{(p)} $)	
		\State $\mL^{(p-1)} := $ Build cost matrix of $(p-1)$-faces \\from successful $p$-simplex matches.
		\EndFor
	\end{algorithmic}
	\begin{flushleft}
		\textbf{Return:}  $ \{ X^{*}_1, \dots, X^{*}_h \}$ \text{ \# set of permutation matrices}
	\end{flushleft}
\end{algorithm}
%\end{comment}
\begin{figure}[tbp]
	\centering
	\includegraphics[width=0.35\textwidth,height=4cm]{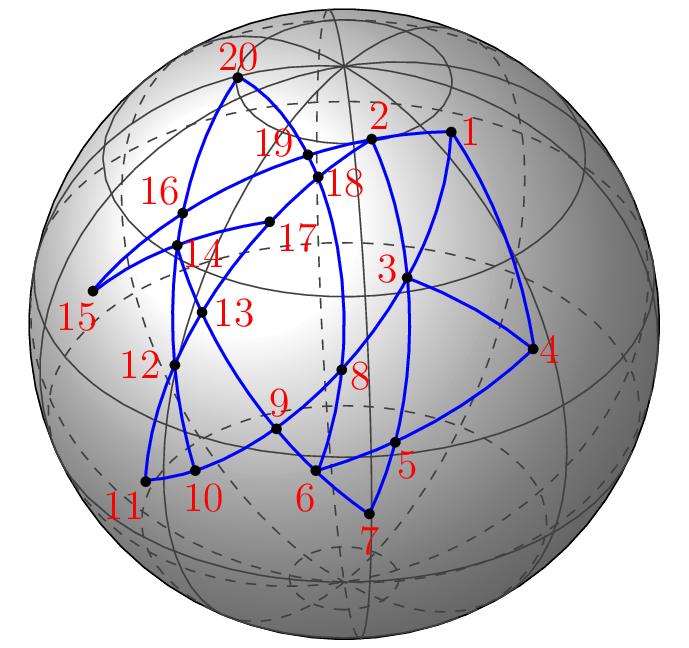}
	\includegraphics[width=0.35\textwidth,height=4cm]{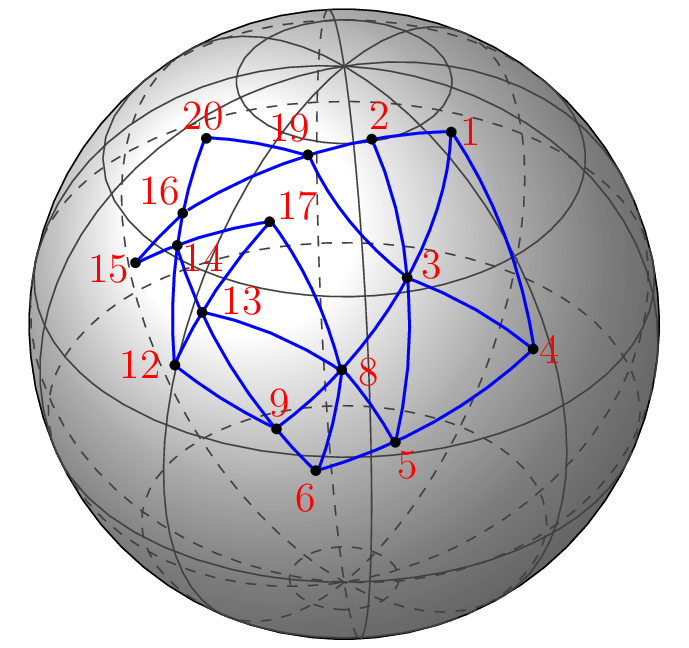}	
	\caption{ Pair of spheres with simplicial complexes constructed between the landmark points on the spheres along with assignments between cliques.}
	%\vspace{-5mm}
	\label{fig:ExampleFig}.
\end{figure}
\begin{table}
	\centering
	\scriptsize
	%\vspace{-0.5cm}
	\caption{Matchings of $3,2$-cliques of simplicial complexes $\mK$ and $\mK'$ shown in Figure~\ref{fig:ExampleFig}.}
	\vspace{1mm}
	%{\scriptsize
	%\hfill{}
	\setlength\tabcolsep{3.5pt}
	\begin{tabular}{lll}
		\toprule
		\multicolumn{1}{l}{\textbf{$k$-Clique}} & \multicolumn{2}{l}{\hspace{1cm}\textbf{Matching between $\mK$ \& $\mK'$}} \\
		
		%\textbf{$20^{\circ}$}&\textbf{$20^{\circ}$}& \multicolumn{7}{|c|}{\textbf{Hello}}\\
		\cline{1-3}
		%\hline
		\hline
		%3-Cliques & \{3,4,5\}, \{5,6,7\} & \{3,4,5\}, \{5,6,8\}\\
		$3$-Cliques & $(1,2,3)$, $(1,3,4)$, $(3,4,5)$, & $(1,2,3)$, $(1,3,4)$, $(3,4,5)$, \\
		% &  &  \\
		% &  &  \\
		& $(2,18,19)$, $(5,6,7)$, $(6,8,9)$, & $(2,3,19)$, $(5,6,8)$, $(6,8,9)$,\\
		% &  &  \\
		% &  & \\
		& $(12,13,14)$, $(13,14,17)$, & $(12,13,14)$, $(13,14,17)$,\\
		%& \{14,15,16\} & \{14,15,16\} \\
		& $(14,15,16)$, $(16,19,20)$. & $(14,15,16)$, $(16,19,20)$. \\

		%FGM-D & 3.0 & 5.0 & 18.0 & 36.0 & 15.0 & 27.0 & 0.0 & 0.0 & 0.0 & 0.0 & 6.0 & 11.0 & 0.0 &0.0\\
		\hline
		%2-Cliques & \{3,4\}, \{3,5\}, \{4,5\}, & \{3,4\}, \{3,5\}, \{4,5\},\\
		%	  & \{5,6\}, \{5,7\}, \{6,7\} & \{5,6\}, \{5,8\}, \{$\emptyset$\}\\
		%&\hspace{0.2cm} \{6,8\}, \{6,9\}, \{8,9\}  &\hspace{0.2cm} \{6,8\}, \{6,9\}, \{8,9\} \\
		$2$-Cliques & $(1,2)$, $(1,3)$, $(2,3)$, $(1,4)$,& $(1,2)$, $(1,3)$, $(2,3)$, $(1,4)$,\\
		& $(3,4)$, $(3,5)$, $(4,5)$, $(2,19)$,& $(3,4)$, $(3,5)$, $(4,5)$, $(2,19)$,\\
		& $(18,19)$, $(5,6)$, $(5,7)$, $(6,8)$, & $(3,19)$, $(5,6)$, $(5,8)$, $(6,8)$,\\
		%& ,& \\
		%& \{5,6\}, \{5,7\}, \{6,8\} & \{5,6\}, \{5,8\}, \{6,8\}\\
		& $(6,9)$, $(8,9)$, $(12,13)$, $(12,14)$, & $(6,9)$, $(8,9)$, $(12,13)$, $(12,14)$,\\
		& $(13,14)$, $(13,17)$, $(14,17)$,& $(13,14)$, $(13,17)$, $(14,17)$, \\
		& $(14,15)$, $(14,16)$, $(15,16)$,& $(14,15)$, $(14,16)$, $(15,16)$,\\
		%& & \\
		& $(16,19)$, $(16,20)$, $(19,20)$, & $(16,19)$, $(16,20)$, $(19,20)$, \\
		& $(3,8)$, $(8,18)$, $(9,10)$, $(9,13)$, & $(3,8)$, $(8,13)$, $(9,12)$, $(9,13)$, \\
		&  $(17,18)$. & $(17,8)$. \\
		\bottomrule
	\end{tabular}
	%\hfill{}
	\vspace{-6mm}	
	\label{tb:ExampleTable}
\end{table}
\paragraph{\textbf{Example}} We illustrate with an example the bijective assignment produced by our algorithm between 
cliques / simplices of a pair of graph-induced \emph{spherical} simplicial complexes, as shown in Figure~\ref{fig:ExampleFig}.  
We consider two simplicial complexes $\mK$ and $\mK'$ each embedded on $\mS^2$, with $20$ and $16$ vertices, respectively. Matching of corresponding $3$-cliques and $2$-cliques are mentioned in the Table~\ref{tb:ExampleTable}. Matching between vertices ($1$-cliques) is shown by marking them with the same label on both spheres.
%\vspace{-1cm}

\paragraph{\textbf{Time complexity analysis}} The major cost incurred by our algorithm arises from matching cliques between two simplicial complexes. Therefore, we first derive an upper bound on the number of cliques that need to be matched as follows (proof in supplementary notes). 
\begin{lemma}
	\label{lemma:clique_count}
	Let $\mK(G)$ represent the simplicial complex induced from graph $G$ with $n$ and $m$ 
	number of vertices and edges, respectively. Let $h$ denote the maximum order of cliques
	in $G$ and $\delta$ be the maximum degree of a vertex in $G$.
	Then, the total number of $k$-cliques in $\mK(G)$ for $k = (1, \dots ,h)$, are at most
	\[
	n + \frac{2m}{ \delta (\delta +1)  } \left[    \min \left\lbrace  (\delta+1)^h +1, \left(  \frac{e(\delta+1)}{h}         \right)^h  \right\rbrace   - \delta -2   \right]	
	\]
\end{lemma}
Neglecting lower order terms, the number of cliques are of order $O(n+ m(\delta^{h-2} - \delta) )$. We 
know that the \emph{spectral relaxation} proposed by Lordeneu et. al.~\cite{leordeanu2005spectral} has a complexity of $O(n^{3/2})$, where $n$ is the number of points to match on each side. Our higher order matching of cliques
then has a time complexity of $O( \{ n+ m(\delta^{h-2} - \delta) \}^{3/2})$. 
In practice, for maximum order of cliques, $h=3$ (triangles) and $h=4$ (tetrahedrons), observe that the complexity 
drops to $O(n^{3/2})$ and $O( \{  n+m \delta^2  \}^{3/2} ) $, respectively, which is very efficient.

\section{Experiments}

For our experiments, we considered \emph{synthetic} and \emph{real-world} datasets that cover both 
\emph{spherical} and \emph{planar} images.
Spherical images can broadly be categorized as: \emph{parabolic omnidirectional} ($360^{\circ}$), \emph{fish-eye}, and 
\emph{panoramic} images.
%Figure~\ref{fig:ImageTransform} shows examples of each type of spherical image. 
Note that our matching algorithm does not require any calibration parameters of cameras.
\if 0
Our experiments comprise of two main categories, namely: \emph{multimodal matching} and 
\emph{ablative studies}. In multimodal matching, we conducted matching experiments on \emph{spherical/SoR-spherical/SoR} 
images and \emph{spherical-planar} images.
\fi

%To evaluate our matching algorithm, we first \emph{extended} the factorized graph matching (FGM) algorithm by feeding it a $k$-NN graph based on geodesic distances between points to serve as our naive baseline method (called ``FGM+geodesic").  

To evaluate our matching algorithm, we compared against three main categories. 
(i) \emph{Planar matching methods extended with geodesic metric on 3D manifolds}: Here, we \emph{extended} the factorized graph matching (FGM)~\cite{ZhouD16} algorithm by feeding it a $k$-NN graph based on geodesic distances between points to serve as our naive baseline method (called ``FGM+geodesic"). The rest of the methods were feature-descriptor based.
(ii) \emph{Planar matching methods on 2D projected (unwrapped\footnote{\emph{unwrapped} implies planar projection of a spherical image with minimal distortion~\cite{cruz2012scale}~\cite{guan2017brisks}.}) manifolds}. 
(iii) \emph{Planar matching methods on 2D planar images}: Here, we proposed a \emph{flat} 
version of our algorithm with Euclidean distance as the underlying metric (called ``OurPlanar") to work on flat 2D images.  

%Our experiments comprise of three main categories as: (ii) Our method vs. planar matching methods with geodesic metric on 3D smooth surfaces, (ii) Our method vs. planar matching methods on 2D-projected smooth surfaces 
%(unwrapped\footnote{\emph{unwrapped} implies planar projection of a spherical image~\cite{cruz2012scale}~\cite{guan2017brisks}.}), and (iii) Our method vs. planar matching methods on 2D planar images .

\begin{figure}
	\centering
	\vspace{-5mm}
	\makebox[\linewidth]{
	\subfigure[]{%
		\label{fig:Vase1}%
		\includegraphics[trim={0 0 0 0},clip,width=30mm,height=45mm]{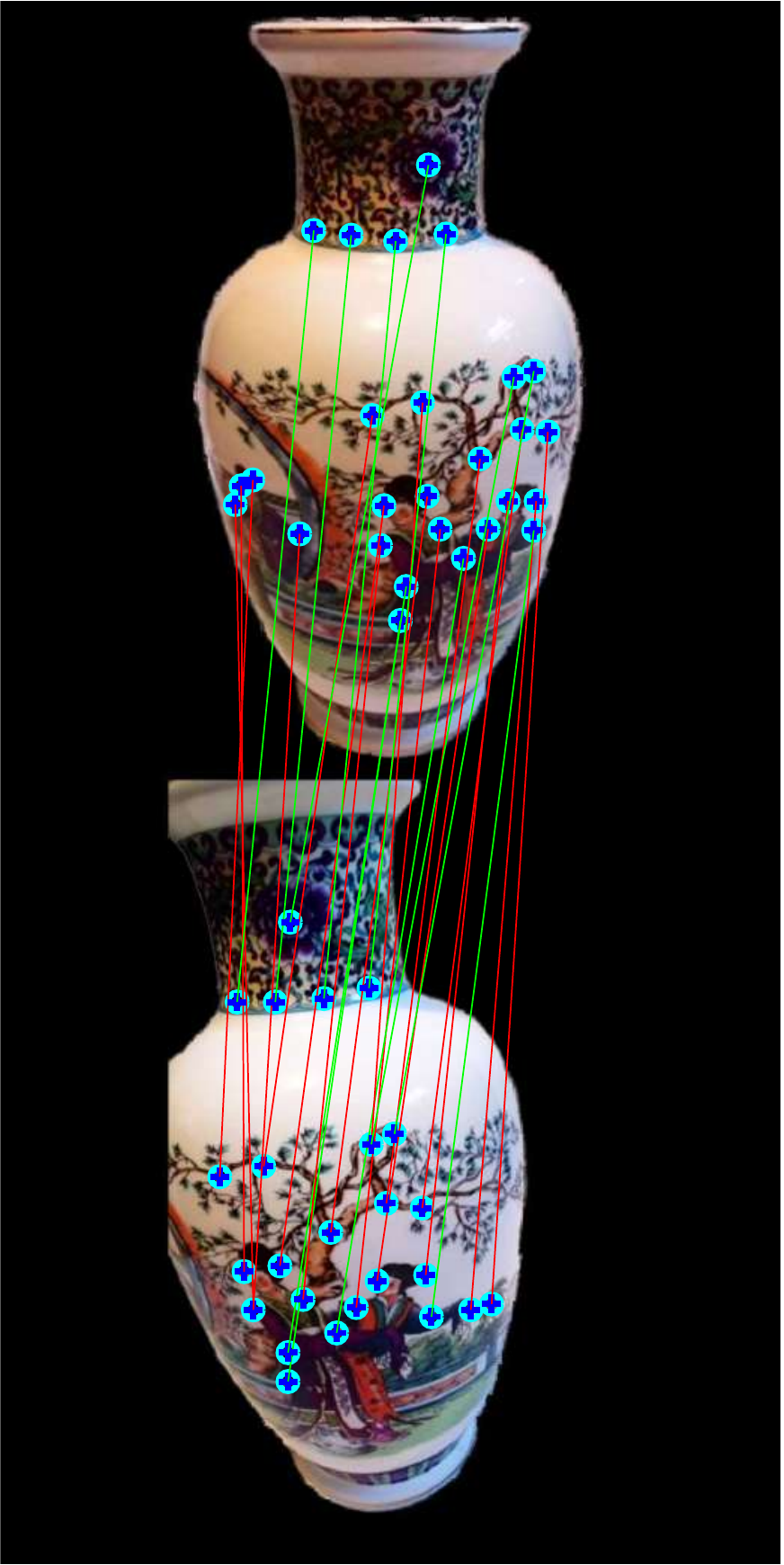}}%
	%\vspace{-0.2cm}
	\qquad
	\subfigure[]{%
		\label{fig:Vase2}%
		\includegraphics[trim={0 0 0 0},clip,width=30mm,height=45mm]{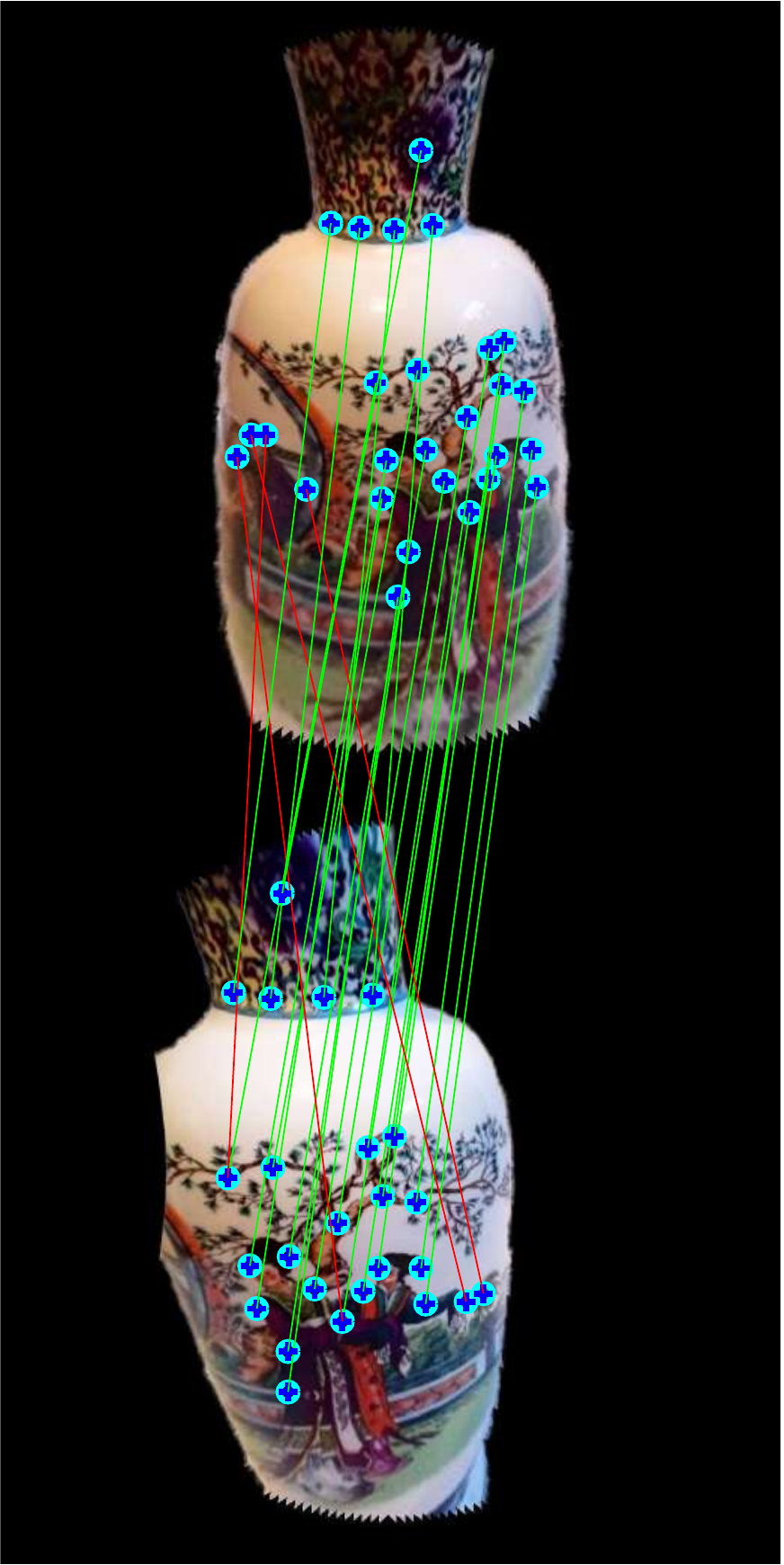}}%
	%\vspace{-2mm}
		\qquad
	\subfigure[]{%
		\label{fig:Vase3}%
		\includegraphics[trim={0 0 0 0},clip,width=30mm,height=45mm]{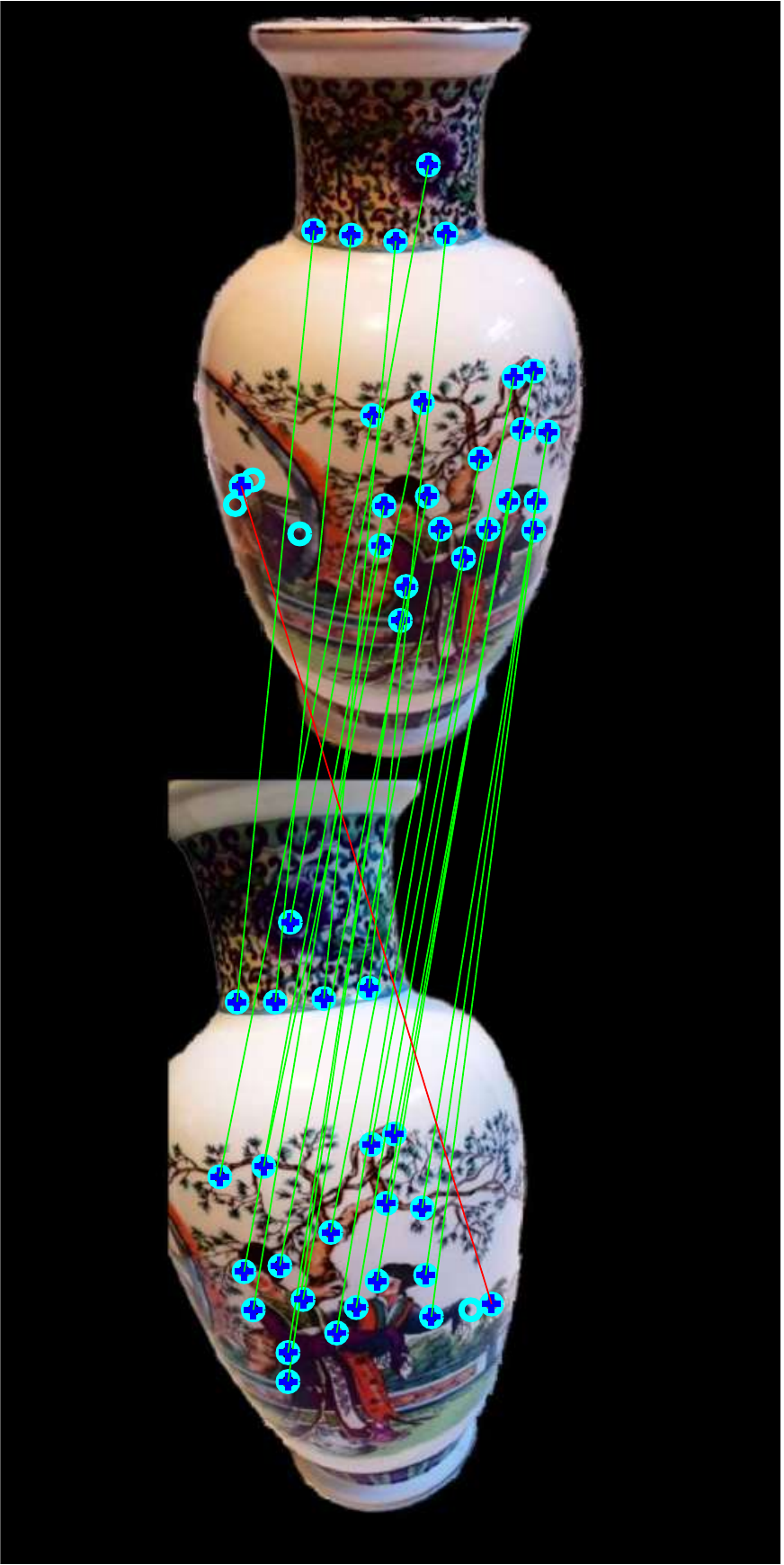}}%
	}
	\vspace{-5mm}
	\caption{Instances of matchings between (a) \emph{Chinese vase} images for Tensor based method, (b) flat version of
		\emph{Chinese vase} images for Tensor based method, and (c) \emph{Chinese vase} images for our method. Green/red lines show correct/incorrect matches respectively. Isolated points show no matches.}
	\label{fig:VaseMatch}
\end{figure}

\begin{figure}
	\centering
	\vspace{-10mm}
	\makebox[\linewidth]{
	\subfigure[]{%
		\label{fig:Retinal1}%
		\includegraphics[trim={0 0 0 0},clip,width=55mm,height=22mm]{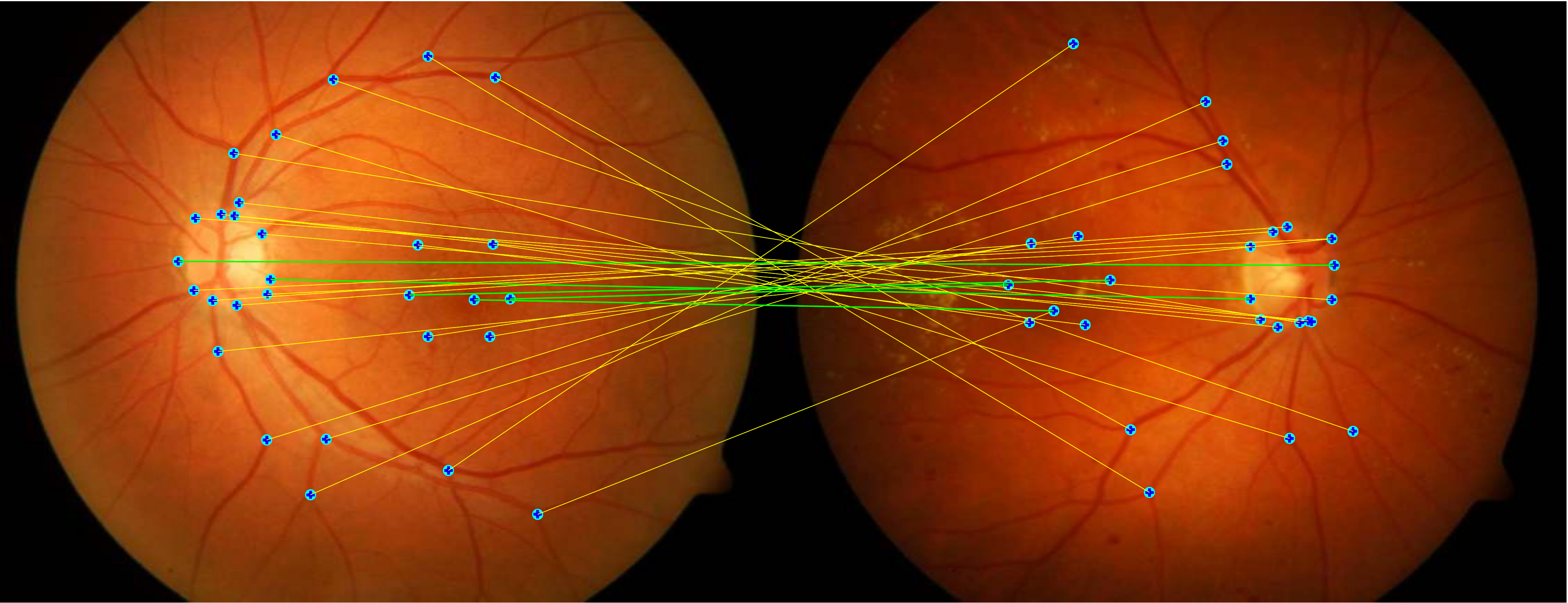}}%
	%\vspace{-0.2cm}
	\qquad
	\subfigure[]{%
		\label{fig:Retinal2}%
		\includegraphics[trim={0 0 0 0},clip,width=55mm,height=22mm]{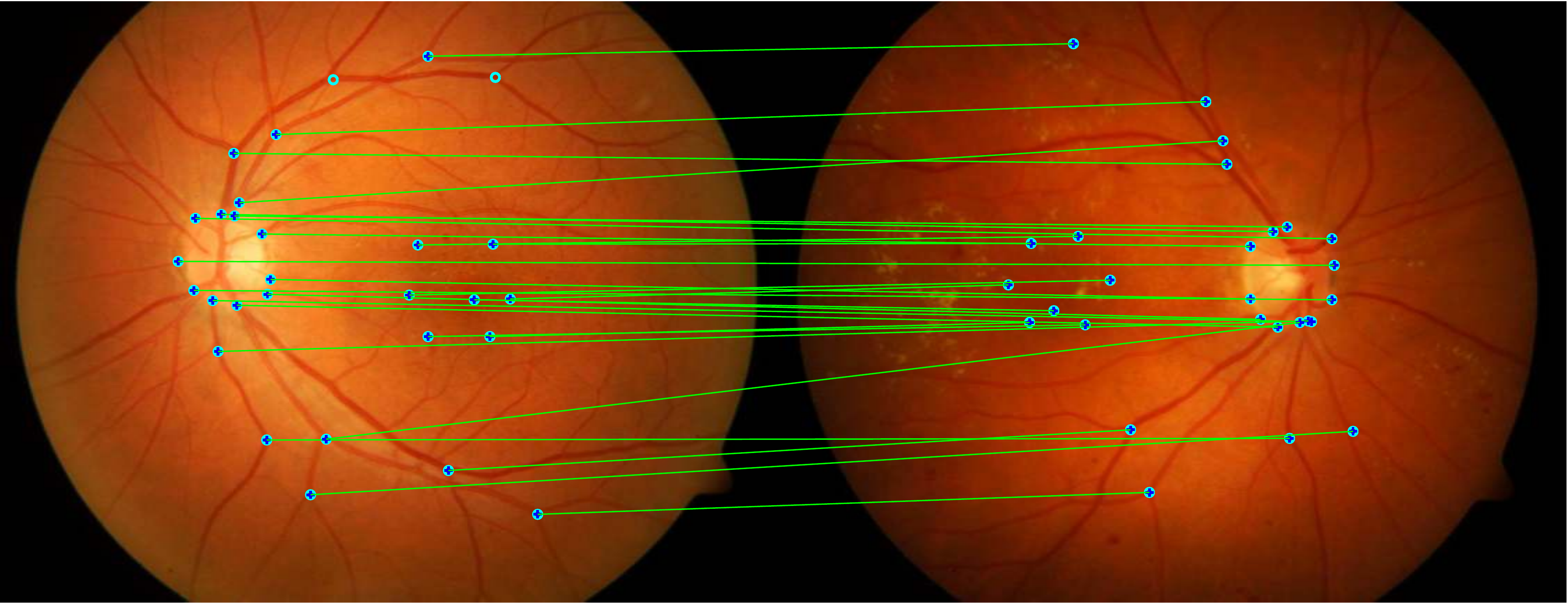}}%
	}
	\vspace{-5mm}
	\caption{Instances of matchings between (a) \emph{Fundus} images for Tensor based method, (b) \emph{Fundus} images for our method. Green/yellow lines show correct/incorrect matches respectively. Isolated points show no matches.}
	\label{fig:Retinal}
\end{figure}

Furthermore, we conducted additional matching experiments on 
\emph{spherical/warped-planar} images  and \emph{unwrapped spherical/warped-planar} images whose results are shown in our supplementary material. We also perform experiments using RANSAC~\cite{hartley2003multiple} for geometric verification and rectification.
In our ablative studies, we analyze the robustness of our algorithm under affine transformations (rotation, reflection, scaling, and shear), under the effect of two noise models (in supplementary material) proposed in~\cite{feizi2016spectral}, and the effects of randomly removing some landmarks to simulate \emph{missing completely at random} (MCAR) phenomena (in supplementary material). 
Additional experiments on rotation for SUN360 dataset and Desktop dataset and the effects of varying $k$ values in a neighborhood of landmark points, varying the radius of the sphere and manifold, and varying the number of landmark points for matching are also shown in supplementary material.
\if 0
\begin{figure}[tbp]
	\makebox[\linewidth]{
		\centering
		\subfigure{%
			\label{fig:zero11}%
			\includegraphics[width=30mm,height=25mm]{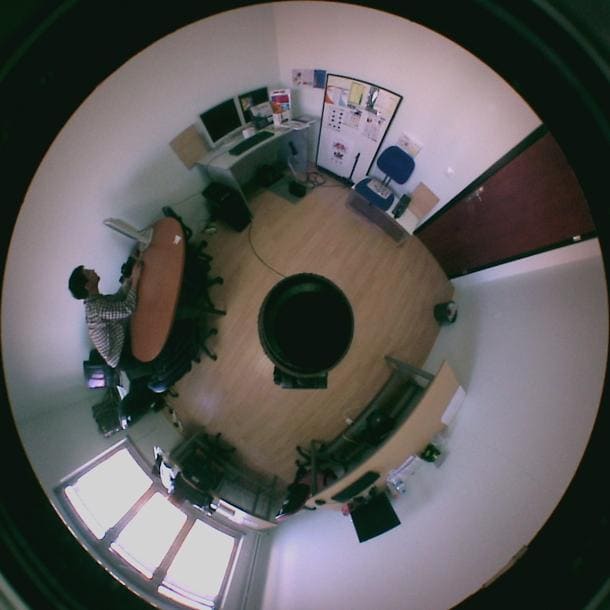}}%
		\qquad
		\subfigure{%
			\label{fig:first11}%
			\includegraphics[width=30mm,height=25mm]{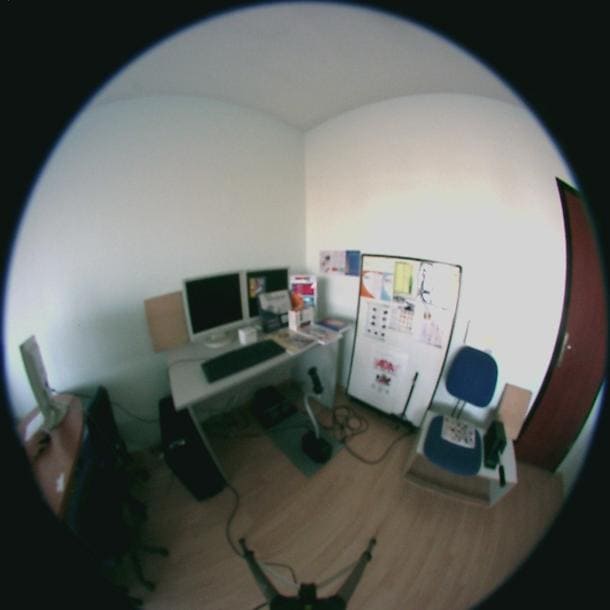}}%
	}
	%\vspace{-0.5cm}
	\centering
	%\vspace{-0.2cm}
	\qquad
	\subfigure{%
		\label{fig:second11}%
		\includegraphics[width=67mm,height=25mm]{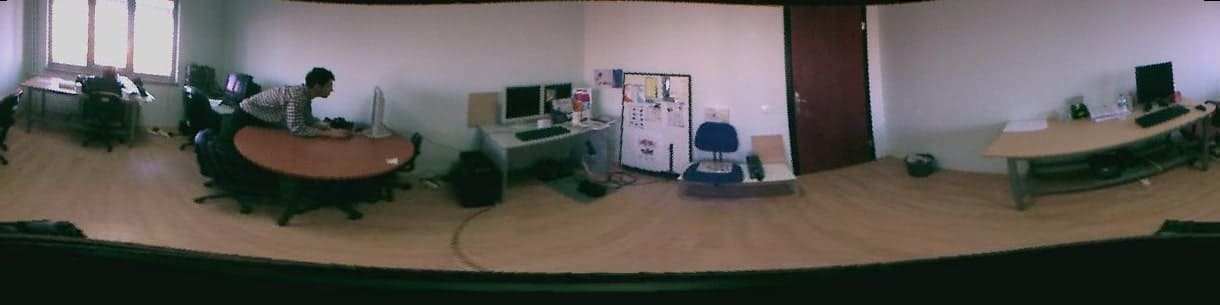}}%
	%\vspace{-5mm}
	\caption{Example of (top-left) a parabolic omnidirectional image, (top-right) fish-eye, and (bottom) panoramic (unwrapped equirectangular) images of the same view. }
	\label{fig:ImageTransform}
\end{figure} 
\fi
\paragraph{\textbf{Baselines} } We group the state-of-the-art methods as: (i) \emph{Feature descriptor based matching for spherical and planar images:} BRISK~\cite{leutenegger2011brisk}, ORB~\cite{rublee2011orb}, SPHORB~\cite{zhao2015sphorb}\footnote{\href{https://github.com/tdsuper/SPHORB}{SPHORB}}.
(ii) \emph{Graph based matching for planar images:} based on employing an \emph{affinity matrix} (FGM)~\cite{ZhouD16,zhou2013deformable}\footnote{\href{http://www.f-zhou.com/gm\_code.html}{FGM}} and \emph{eigenvalues} (EigenAlign)~\cite{feizi2016spectral}\footnote{\href{https://github.com/SoheilFeizi/spectral-graph-alignment}{EigenAlign}}.
(iii) \emph{Higher-order based matching for planar images:} using random clique complex (RCC)~\cite{sharma2018solving} and higher-order matching (Tensor)~\cite{duchenne2011tensor}\footnote{\href{http://www.cs.cmu.edu/~olivierd/}{Tensor}}. 
(iv) Finally, a naive baseline (FGM+geodesic) proposed by us that extends FGM by constructing a graph based on geodesic distances.
%Note that some baselines are only used in our additional experiments that show in our supplementary notes. 
\paragraph{\textbf{Datasets}}

%Description of six datasets are reported in Table~\ref{tab:datasets}. 
We sourced raw spherical image datasets without landmarks and annotated them using 
FAST (Features from Accelerated Segment Test) corner detector~\cite{rosten2006machine} 
%\emph{landmark point detection}\cite{} and 
and generated the \emph{matching ground truth} for all the datasets used in our experiments. 
%We grouped the datasets into two categories. The first category consists of only spherical images (omnidirectional, fish-eye and panoramic) and the second one comprises of both spherical (omnidirectional and fish-eye only) and planar images in them. 
All six datasets are as follows (with their details in supplementary):
$(i)$ \emph{Chessboard} dataset consists of omnidirectional and fish-eye images. 
$(ii)$ \emph{Kamaishi} and $(iii)$ \emph{SUN360} dataset has panoramic images.
$(iv)$ \emph{Desktop} dataset consists of omnidirectional and planar images.
$(v)$ \emph{Parking} dataset also has omnidirectional and planar images. 
$(vi)$ \emph{Table} dataset contains omnidirectional, fish-eye, and planar images. 
Our code\footnote{\href{https://github.com/charusharma1991/PointCorrespondence}{Our Method}} is publicly available.

\begin{figure}
	\centering
	%\vspace{-10mm}
	\makebox[\linewidth]{
	\subfigure[]{%
		\label{fig:Desktop1}%
		\includegraphics[trim={0 0 0 0},clip,width=55mm,height=22mm]{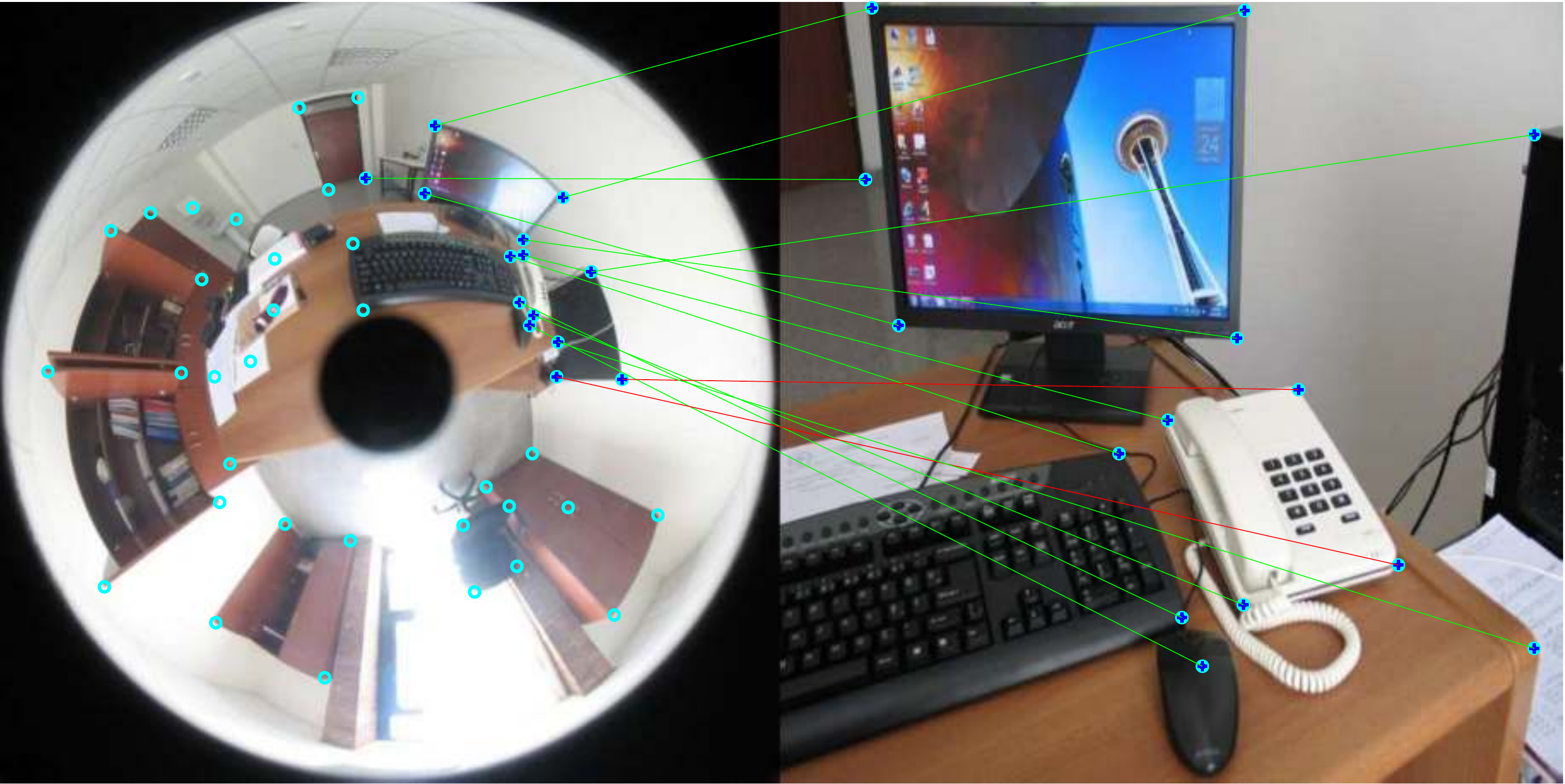}}%
	%\caption{a) Matchings for house frame $1$ (left) with frame $55$ (right), b) Matchings for house frame $1$ (left) with frame $110$ (right), Error: $0.0\%$.}
	%\vspace{-0.2cm}
	\qquad
	%\vspace{-1cm}
	\subfigure[]{%
		\label{fig:Chess1}%
		\includegraphics[trim={0 0 0 0},clip,width=55mm,height=22mm]{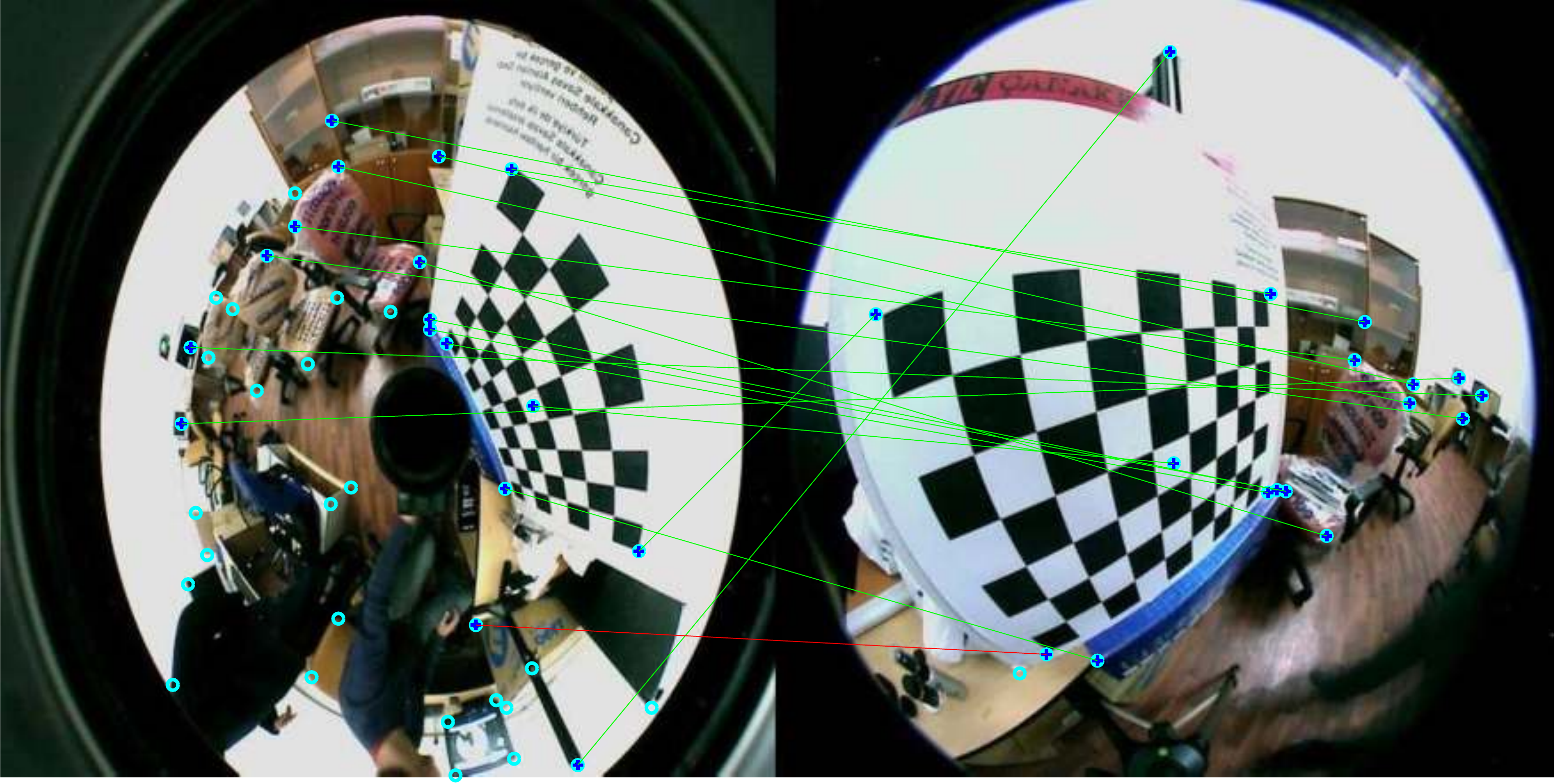}}%
	}
%	\vspace{-5mm}
	%\vspace{-2mm}
	\caption{Instances of matchings between (a) \emph{Desktop} omnidirectional and planar images and (b) \emph{Chessboard} omnidirectional and fish-eye images. Green/red lines show correct/incorrect matches, respectively. Isolated points show no matches.}
	\label{fig:MatchingSS}
\end{figure}

%\paragraph{Multimodal Matching:}
\paragraph{\textbf{Our method vs. planar matching methods with geodesic metric on 3D curved manifolds }}
\begin{table*}
	\centering
	\vspace{-6mm}
	\caption{Error (\%) of pairwise matching between spherical images (omnidirectional, fish-eye and panorama) of five datasets for different methods.}
	\vspace{1mm}
	\footnotesize
	\setlength\tabcolsep{3.5pt}
	\begin{tabular}{llllll}
		\toprule
		\textbf{Algorithms} & \textbf{Kamaishi} & \textbf{Chessboard} & \textbf{Desktop} & \textbf{Parking} & \textbf{Table} \\
		\hline
		
%		\emph{OurCone} & 0.44 $\pm$ 0.0 \%& 3.74 $\pm$ 0.0 \%& 0.21 $\pm$ 0.0 \%& 0.0 $\pm$ 0.0 \%& 0.83 $\pm$ 0.0 \% \\
%		\emph{OurEllip} & 0.31 $\pm$ 0.0 \%& 3.72 $\pm$ 0.0 \%& 0.85 $\pm$ 0.0 \%& 0.0 $\pm$ 0.0 \%& 2.63 $\pm$ 0.0 \% \\
		%BRISKS & & & & & \\ 
		\emph{OurWarped} & \textbf{0.79} $\pm$ \textbf{0.0} \% & \textbf{3.89} $\pm$ \textbf{0.0} \%& \textbf{0.32} $\pm$ \textbf{0.0} \%& \textbf{0.0} $\pm$ \textbf{0.0} \%& \textbf{0.74} $\pm$ \textbf{0.0} \%\\
		\hline
		FGM+geo & 55.6 $\pm$ 0.10 \%& 79.2  $\pm$ 1.21 \%& 23.3 $\pm$ 0.03 \%& 37.5 $\pm$ 0.0 \%& 64.3 $\pm$ 6.58 \%\\
		%(Naive baseline) & & & & & \\
		SPHORB & 90.0 $\pm$ 0.0 \%& 58.5 $\pm$ 0.0 \%& 91.1 $\pm$ 0.0 \%& 95.0 $\pm$ 0.0 \%& 78.5 $\pm$ 0.0 \%\\ 
		%SIFTS & & & & & \\
		BRISK & 85.6 $\pm$ 0.0 \%& 53.6 $\pm$ 0.0 \%& 78.9 $\pm$ 0.0 \%& 81.6 $\pm$ 0.0 \%& 69.2 $\pm$ 0.0 \%\\
		ORB & 90.2 $\pm$ 0.0 \%& 53.8 $\pm$ 0.0 \%& 51.7 $\pm$ 0.0 \%& 71.1 $\pm$ 0.0 \%& 64.4 $\pm$ 0.0 \%\\
		Tensor & 37.7 $\pm$ 0.69 \%& 60.5 $\pm$ 0.41 \%& 23.9 $\pm$ 1.7 \%& 23.7 $\pm$ 7.5 \%& 85.1 $\pm$ 1.05 \% \\
		FGM & 53.3 $\pm$ 0.21 \%& 80.0 $\pm$ 0.11 \%& 31.9 $\pm$ 0.12 \%& 36.0 $\pm$ 1.5 \%& 65.5 $\pm$ 0.01 \%\\
		%SIFT &  & & & \\
		\bottomrule
	\end{tabular}
	%\vspace{-5mm}
	
	\label{tab:SS}
\end{table*}
Here, we match pairwise images directly on the warped images on curved manifolds (as shown in Figures \ref{fig:VaseMatch}, \ref{fig:Retinal}, and~\ref{fig:Chess1}). The comparison between standard higher-order graph matching (Tensor)~\cite{duchenne2011tensor} and our method on manifold is shown in Figures \ref{fig:VaseMatch} and \ref{fig:Retinal} using
Chinese vases\footnote{from Google images} and Fundus images~\cite{kalesnykiene2006diaretdb0}, respectively. 
We observe from Figures~\ref{fig:Vase1} and \ref{fig:Retinal1} that the Tensor based method does not 
perform well on warped images. Although, the matching does improve when images are 
flattened to reduce the effect of curvature in Figure~\ref{fig:Vase2}. Our method outperforms the 
baseline and has a maximum number of correct matches in Figures \ref{fig:Vase3} and~\ref{fig:Retinal2}.

The error percentages of our warped image matching algorithm (OurWarped) are shown in the first row of Table~\ref{tab:SS}. 
%To compare our method against planar matching methods, we conducted two kinds of experiments on one of the popular planar matching methods i.e. 
%FGM. First, we ran FGM directly on spherical images. Second, We modify planar distance to geodesic distance to construct graph of landmark points. Then, this graph is fed to FGM. We consider the modified version of FGM (FGM$+$geodesic) as a naive baseline for our method on smooth surface. 
%Results for both the experiments are mentioned in Table~\ref{tab:SS}.
We observe that our method outperforms all other matching methods, including spherical feature descriptor based ones as well. Additional multimodal warped-planar matching experiments
can be found in our supplementary notes.

For matches between spherical and planar images, we find two variants which match between a spherical and a planar image (Figure \ref{fig:Desktop1}) and matching between different types of spherical images (Figure \ref{fig:Chess1}).
In Table~\ref{tab:SS}, there is a slight increase in error percentages when matching across different types of spherical images, i.e., $3.89 \%$ for Chessboard, as compared to matching similar types, i.e., $0.32 \%$ for Desktop, due to differences in distortion levels. 
In spite of this, we find that our method significantly outperforms naive baseline and other matching methods on spherical images. 
%There are five datasets consist of variants of spherical images like panorama, fish-eye and omnidirectional. Kamaishi dataset has panorama images of 
%a frame sequence of a moving car. It has the wide angle view of the scene which introduces distortion in the images. The other four %datasets includes omnidirectional ($360^{\circ}$) and fish-eye images. 
%We perform pairwise matching of all the spherical images in the datasets. 
%This experiment compares two spherical images where the distortion would almost be same if both the images are taken 
%from same kind of camera. For example, comparing omnidirectional with fish-eye image would be difficult than comparing both the %omnidirectional images. We can find the difference between the two types of images in 
%Figure~\ref{fig:ImageTransform}.
%Our method in Table~\ref{tab:SS} shows better results than the other methods.

\paragraph{\textbf{Our method vs. planar matching methods on 2D-projected curved manifolds}}
\begin{table}
		\centering
		\vspace{-7mm}
		\caption{Error (\%) of pairwise matching between unwrapped equirectangular version of spherical (omnidirectional and fish-eye) images of four datasets for different methods including graph matching methods on flat surfaces.}
		%\vspace{1mm}
		\footnotesize
		\begin{tabular}{lllll}
			\toprule
			\textbf{Algorithms} & \textbf{Chessboard} & \textbf{Desktop} & \textbf{Parking} & \textbf{Table} \\
%			\hline
			\hline
%			\emph{OurCone} & 4.19 $\pm$ 0.0 \%& 0.32 $\pm$ 0.0 \%& 0.0 $\pm$ 0.0 \%& 0.96 $\pm$ 0.0 \%\\
%			\emph{OurEllip} & 4.66 $\pm$ 0.0 \%& 0.64 $\pm$ 0.0 \%& 2.63 $\pm$ 0.0 \%& 3.55 $\pm$ 0.0 \%\\
			\emph{OurWarped} & \textbf{3.64 $\pm$ 0.0} \% & \textbf{1.06 $\pm$ 0.0} \%& \textbf{0.0 $\pm$ 0.0} \%& \textbf{0.57 $\pm$ 0.0} \%\\
			\hline
			RCC & 28.6 $\pm$ 0.94 \%& 11.6 $\pm$ 0.74 \%& 13.2 $\pm$ 11.8 \%& 11.6 $\pm$ 0.57 \%\\ 
			EigenAlign & 98.47 $\pm$ 0.0 \%& 95.24 $\pm$ 0.0 \%& 97.5 $\pm$ 0.0 \%& 97.9 $\pm$ 0.0 \%\\ 
			Tensor & 68.9 $\pm$ 0.16  \%& 26.1 $\pm$ 0.58 \%& 19.0 $\pm$ 3.75 \%& 72.4 $\pm$ 0.67 \%\\
			FGM & 84.0 $\pm$ 0.0 \%& 31.0 $\pm$ 0.0 \%& 38.0 $\pm$ 0.0 \%& 52.0 $\pm$ 0.0 \%\\ 
			SPHORB & 58.6 $\pm$ 0.0 \%& 90.3 $\pm$ 0.0 \%& 97.5 $\pm$ 0.0 \%& 79.2 $\pm$ 0.0 \%\\ 
			%SIFTS &  & & & \\
			BRISK & 54.9 $\pm$ 0.0 \%& 84.9 $\pm$ 0.0 \%& 100.0 $\pm$ 0.0 \%& 74.2 $\pm$ 0.0 \%\\
			ORB & 49.5 $\pm$ 0.0 \%& 78.2 $\pm$ 0.0 \%& 82.5 $\pm$ 0.0 \%& 70.3 $\pm$ 0.0 \%\\
			%SIFT &  & & & \\
			\bottomrule
		\end{tabular}
		%\vspace{-3mm}
		
		\label{tab:P2P2}
\end{table}

Matching between spherical images can also be performed by applying planar graph matching methods on unwrapped equirectangular versions of spherical images. This makes the image flat and standard planar matching algorithms can then be employed. However, any kind of projection (on a flat surface in this case) introduces distortions in the resulting image. We flattened spherical images for four datasets mentioned in Table~\ref{tab:P2P2}. We used two different methods to flatten omnidirectional and fish-eye images. 
The $360^{\circ}$ 
image is unwrapped by dividing it into four parts (quadrants) and concatenated into a single flat image. On the other hand, fish-eye images do not cover the complete view of the scene and add distortion to the image due to curved mirrors and lenses of the cameras. We try to reduce the distortion by removing curves and flattening the image using calibration techniques outlined in~\cite{scaramuzza2006}. 
Since any projection will lead to distortion, we can compare the results from Table~\ref{tab:SS} with Table~\ref{tab:P2P2}. Both the experimental outcomes are based on the same set of spherical images. Our matching algorithm significantly outperforms its competitors on both the spherical images and on curved manifolds.
\paragraph{\textbf{Our method vs. planar matching methods on 2D planar images }}
Our proposed method can also be reduced to a higher-order planar graph matching method. To show the importance of higher-order combinatorial matching not only with geodesic neighborhood, but also with euclidean neighborhood, we run our planar variant (OurPlanar) on popular 2D 
image datasets, competing with standard matching algorithms. We pick four well-known difficult matching datasets (Books, Building, Magazine, and Butterfly) that suffer from heavy occlusions and non-affine transformations~\cite{sharma2018solving}. Results for such an experiment are shown in Table~\ref{tab:Planar}. 
From the results, we observe that our method also serves as a powerful planar matching method and is still competitive using an euclidean neighborhood for our affine weight vectors. It significantly outperforms both the popular planar matching methods.

\begin{table}
	\centering
	%\vspace{-1mm}
	\caption{Error (\%) of pairwise matching between planar images of four datasets for different methods.}
	\vspace{1mm}
	\footnotesize
	%\footnotesize
	\setlength\tabcolsep{3.5pt}
	\begin{tabular}{lllll}
		\toprule
		\textbf{Algorithms} & \textbf{Magazine} & \textbf{Building} & \textbf{Books} & \textbf{Butterfly} \\
		\hline
		%\emph{OurSoR\_Cone} & 14.71 $\pm$ 0.0 \%& 4.41 $\pm$ 0.0 \%& 0.88 $\pm$ 0.0 \%\\
		%\emph{OurSoR\_Ellip} & 13.82 $\pm$ 0.0 \%& 5.94 $\pm$ 0.0 \%& 2.65 $\pm$ 0.0 \%\\
		%\hline
		%BRISKS &  & & \\ 
		\emph{OurPlanar} & \textbf{0.0} $\pm$ \textbf{0.0} \%& \textbf{1.03} $\pm$ \textbf{0.01} \%& \textbf{19.72} $\pm$ \textbf{0.20} \% & \textbf{0.0} $\pm$ \textbf{0.0} \%\\
		FGM & \textbf{0.0} $\pm$ \textbf{0.0} \%& 74.87 $\pm$ 0.07 \%& 97.54 $\pm$ 0.01 \%& 16.12 $\pm$ 0.0 \%\\
		%SPHORB & 41.5 $\pm$ 0.0  \%& 53.85 $\pm$ 0.0 \%& 70.88 $\pm$ 0.0 \%\\ 
		%SIFTS &  & & \\
		%RISK & 37.84 $\pm$ 0.0 \%& 51.05 $\pm$ 0.0 \%& 65.11 $\pm$ 0.0 \%\\
		%ORB & 34.82 $\pm$ 0.0 \%& 48.82 $\pm$ 0.0 \%& 62.78 $\pm$ 0.0 \%\\
		%SIFT &  & & \\
		Tensor & \textbf{0.0} $\pm$ \textbf{0.0} \%& 43.24 $\pm$ 2.98 \%& 32.35 $\pm$ 0.15 \%& 1.07 $\pm$ 0.17\%\\
		%FGM & 98.0 $\pm$ 0.0 \%& 97.50 $\pm$ 0.01 \%& 97.5 $\pm$ 0.01 \%\\
		\bottomrule
	\end{tabular}
	%\vspace{-7mm}
	\label{tab:Planar}
\end{table}

\begin{figure*}[tbp]
	\centering
	%\vspace{-5mm}
	\makebox[\linewidth]{
		\subfigure[]{%
			\label{fig:affine1}%
			\includegraphics[width=30mm,height=27mm]{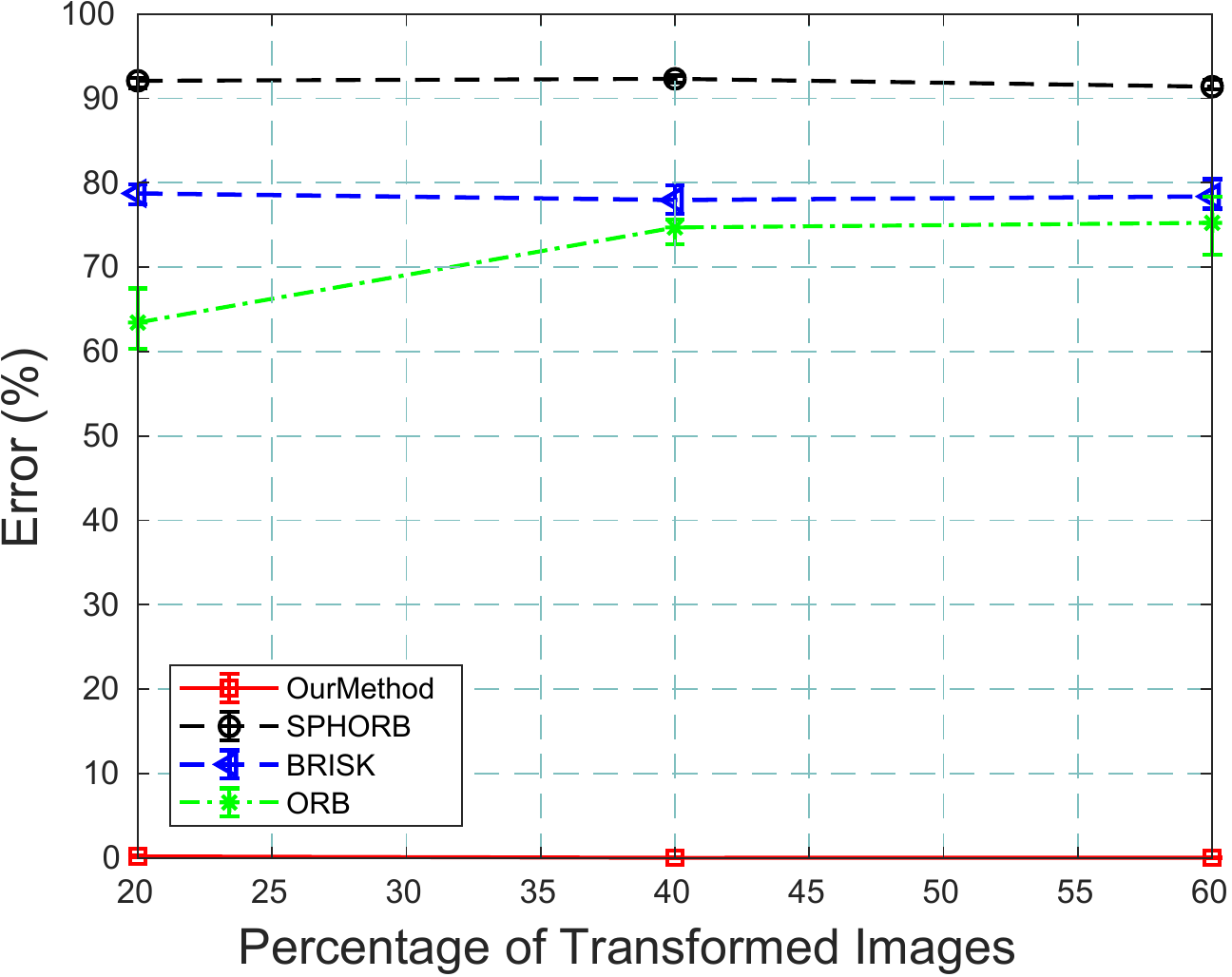}}%
		\hspace{-0.7cm}
		\qquad
		\subfigure[]{%
			\label{fig:affine2}%
			\includegraphics[width=30mm,height=27mm]{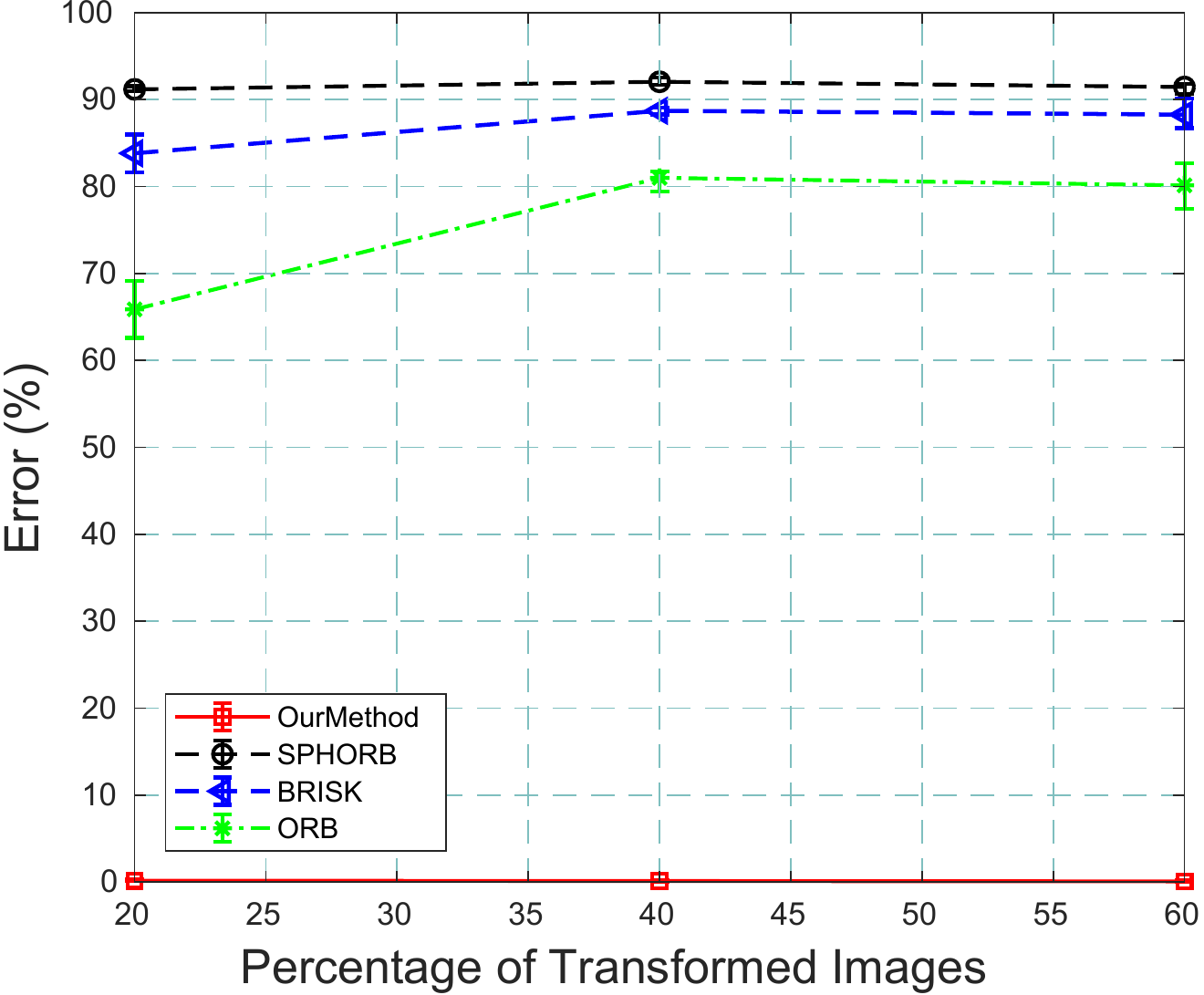}}%
		\hspace{-0.7cm}
		\qquad
		\subfigure[]{%
			\label{fig:affine3}%
			\includegraphics[width=30mm,height=27mm]{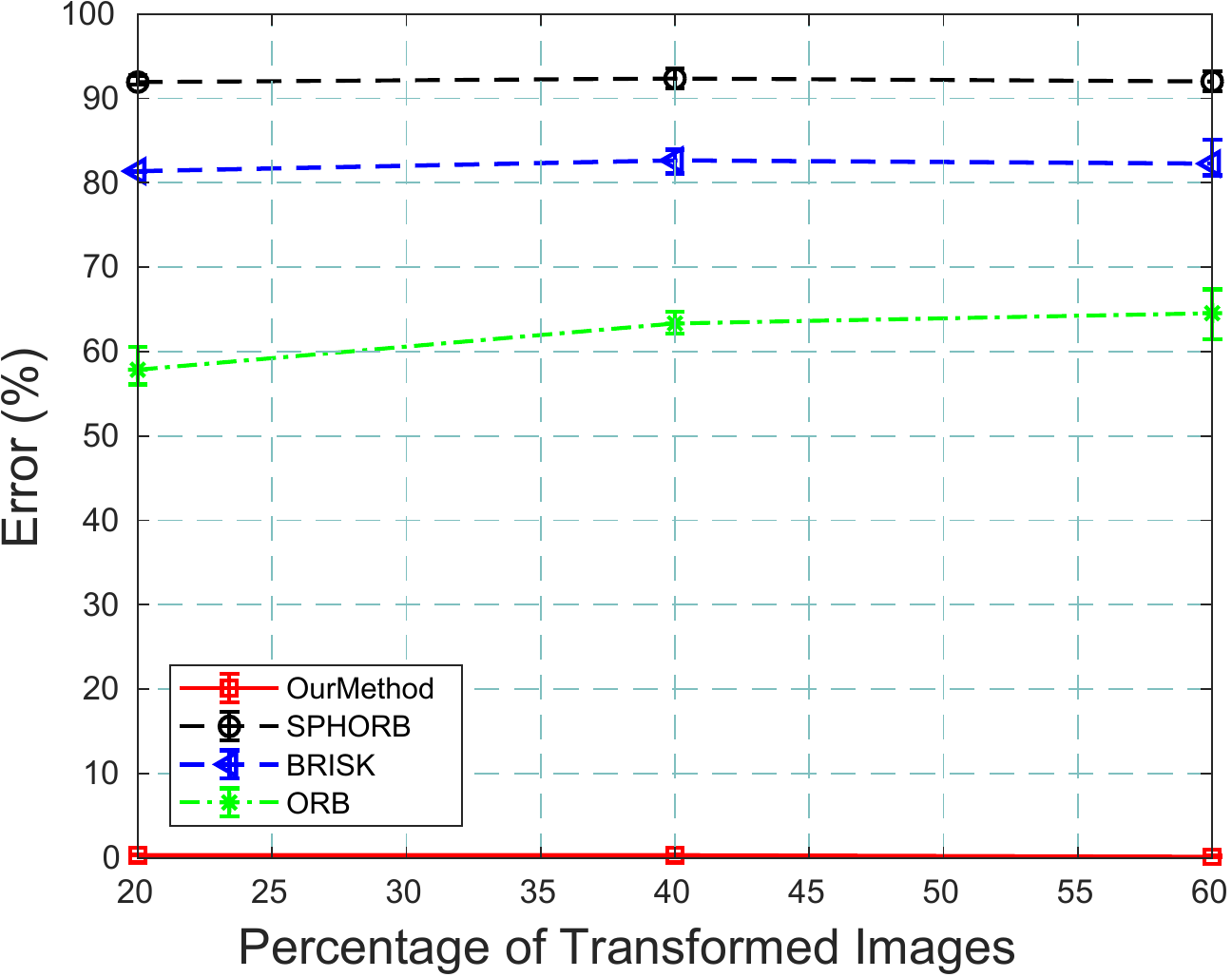}}%
		\hspace{-0.7cm}
		\qquad
		\subfigure[]{%
			\label{fig:affine4}%
			\includegraphics[width=30mm,height=27mm]{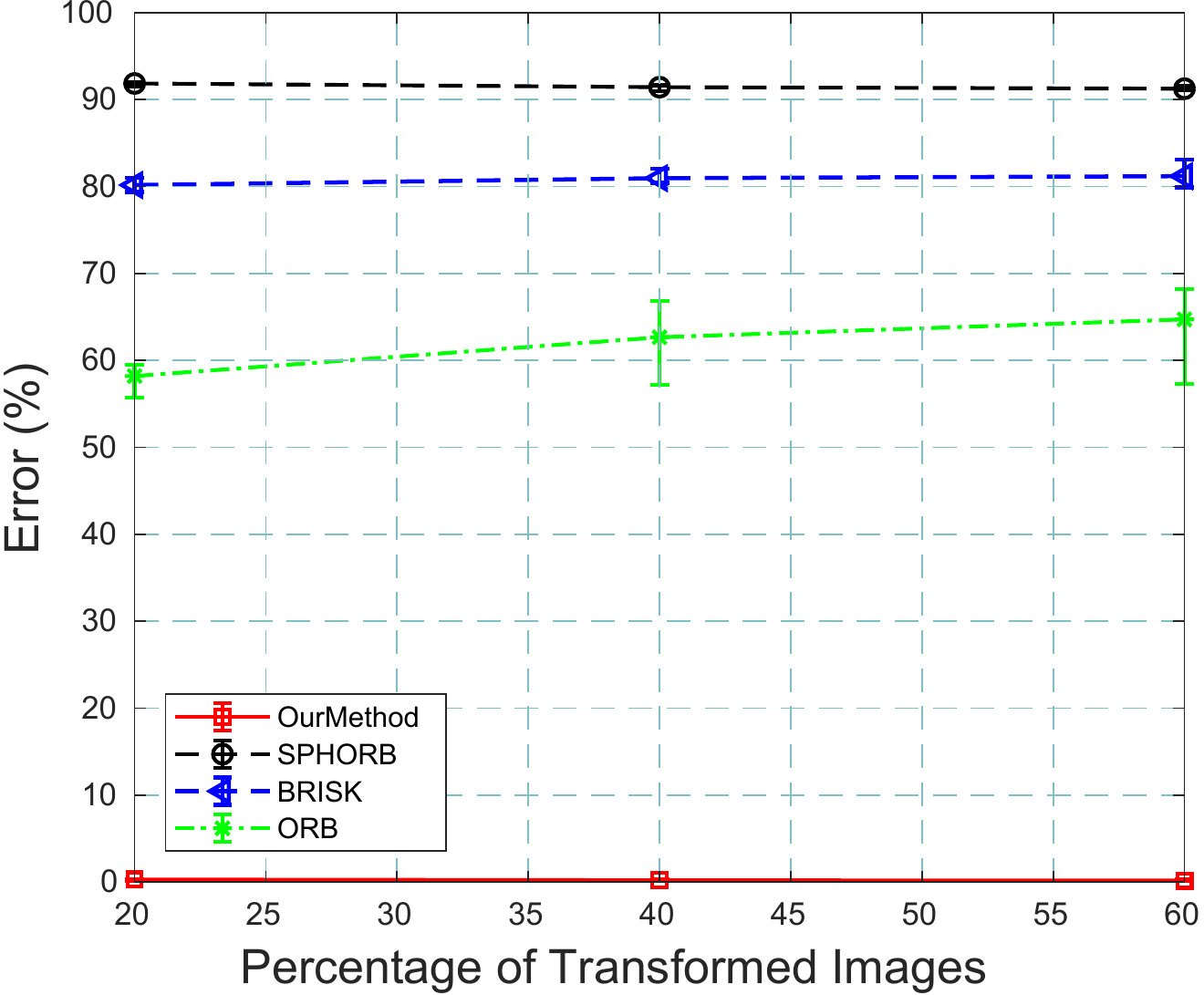}}%
	}
	\vspace{-5mm}
	\caption{Error(\%) in matching when varying the percentage ($20\%$ to $60\%$) of transformed images in the set of spherical images of \emph{Desktop} $(a)$-$(d)$. $(a)$ $40^\circ$ rotation, $(b)$ reflection, $(c)$ scaling and $(d)$ shear.}
	\label{fig:Affine}
\end{figure*}
\vspace{-5mm}
\paragraph{\textbf{RANSAC: Geometric Verification and Rectification}}
We also performed fundamental matrix based geometric verification using RANSAC algorithm~\cite{hartley2003multiple} after descriptor based matching on two datasets for spherical and their planar versions for SPHORB and ORB in Table~\ref{tab:Verify}. 
We observed that the results are improved (but still not better than our proposed method) in some cases but prune a lot of matches. Nearly $40-50\%$ of matches are considered as outliers which makes it difficult to handle the noise. 
On the other hand, our method performs much better in any case while considering outliers.

\begin{table}
	\centering
	%\vspace{-5mm}
	\caption{Error (\%) of pairwise matching between spherical images of Desktop and Parking datasets and on their unwrapped versions for verification.}
	\vspace{1mm}
	\footnotesize
	\setlength\tabcolsep{3.5pt}
	\begin{tabular}{lllll}
		\toprule
		\textbf{Algorithms} & \textbf{Desktop} & \textbf{Desktop\_flat} & \textbf{Parking} & \textbf{Parking\_flat}\\
		\hline
		%SPHORB & 91.1 $\pm$ 0.0 \%& 90.28 $\pm$ 0.0 \%& 95.0 $\pm$ 0.0 \%& 97.5 $\pm$ 0.0 \%\\ 
		%%SIFTS & & & & & \\
		%ORB & 51.7 $\pm$ 0.0 \%& 78.23 $\pm$ 0.0 \%& 71.1 $\pm$ 0.0 \%& 82.5 $\pm$ 0.0 \%\\
		%%BRISK & 78.9 $\pm$ 0.0 \%& 84.91 $\pm$ 0.0 \%& 81.6 $\pm$ 0.0 \%& 100.0 $\pm$ 0.0 \%\\
		OurWarped & \textbf{0.32 $\pm$ 0.0 \%} & \textbf{1.06 $\pm$ 0.0 \%} & \textbf{0.0 $\pm$ 0.0 \%} & \textbf{0.0 $\pm$ 0.0 \%}\\
		%\emph{OurSoR\_Cone} & 0.44 $\pm$ 0.0 \%& 3.74 $\pm$ 0.0 \%& 0.21 $\pm$ 0.0 \%& 0.0 $\pm$ 0.0 \%& 0.83 $\pm$ 0.0 \% \\
		%\emph{OurSoR\_Ellip} & 0.31 $\pm$ 0.0 \%& 3.72 $\pm$ 0.0 \%& 0.85 $\pm$ 0.0 \%& 0.0 $\pm$ 0.0 \%& 2.63 $\pm$ 0.0 \% \\
		%\hline
		%BRISKS & & & & & \\ 
		%\emph{OurSpherical} & \textbf{0.79} $\pm$ \textbf{0.0} \% & \textbf{3.89} $\pm$ \textbf{0.0} \%& \textbf{0.32} $\pm$ \textbf{0.0} \%& \textbf{0.0} $\pm$ \textbf{0.0} \%& \textbf{0.74} $\pm$ \textbf{0.0} \%\\
		SPHORB+RANSAC & 96.1 $\pm$ 0.0 \%& 93.9 $\pm$ 0.0 \%& 95.0 $\pm$ 0.0 \%& 100.0 $\pm$ 0.0 \%\\ 
		%SIFTS & & & & & \\
		ORB+RANSAC & 29.3 $\pm$ 0.0 \%& 70.6 $\pm$ 0.0 \%& 55.0 $\pm$ 0.0 \%& 100.0 $\pm$ 0.0 \%\\
		%BRISK$+$Rectification & 52.11 $\pm$ 0.0 \%& - & - & -\\
		%Tensor & 37.7 $\pm$ 0.69 \%& 60.5 $\pm$ 0.41 \%& 23.9 $\pm$ 1.7 \%& 23.7 $\pm$ 7.5 \%& 85.1 $\pm$ 1.05 \\
		%SIFT &  & & & \\
		\bottomrule
	\end{tabular}
	%\vspace{-3mm}
	
	\label{tab:Verify}
\end{table}

We performed rectification~\cite{bradski2008learning} on spherical images of Desktop dataset followed by BRISK descriptor for matching. The results improved from $78.9\%$ (in Table~\ref{tab:SS}) to $52.11\%$ error. However, we observed that despite these improvements, our method still outperforms them. Also, in most of the cases, the rectification algorithm does not perform well and outputs noisy or distorted images. So, there is no guarantee to find the 
best solution. 

\paragraph{\textbf{Ablative Studies (Effect of Affine Transformation)}}
We \emph{remove} completely at random $40-80\%$ of landmark points 
on the Desktop dataset, and introduce 
affine transformations on these points.
Figure~\ref{fig:Affine} shows the results of affine transformation like \emph{rotation, reflection, scaling,} 
and \emph{shear}.
%Omnidirectional images from Desktop dataset are considered for all the affine transformations. 
We rotated images (clockwise) by $40^\circ$ and performed matching for four algorithms. Then, we generated mirror images along the $x$-axis from the same dataset to introduce reflection. 
We also conducted transformation by scaling and shear of $360^\circ$ images. We resized images in both the directions with scales $0.5$ and $1.5$ randomly. For shearing, we stretched images with $0.5$ factor along $y$-axis. For all types of transformations, we observe that the results shown in Figure~\ref{fig:Affine} clearly indicates that our method is robust to all kinds of affine transformations and easily outperforms other state-of-the-art methods.

\section{Conclusion}
We presented a bijective assignment between sets of landmark points embedded on a pair of images warped onto curved manifolds by the following steps. 
First, we built a \emph{graph induced simplicial complex} on the warped images. 
%This was done to encode higher-order structures. 
Second, we proposed a constrained QAP that matches corresponding co-dimensional simplices between two simplicial complexes along with an efficient algorithm to solve the constrained QAP.
Finally, we conducted extensive experiments, broadly grouped as \emph{comparative matching} and \emph{ablative studies}, in order to gain insight into the accuracy and robustness of our method.
%, given the novel challenges that are encountered with images on smooth curved surfaces. 
%Our empirical studies confirm the accuracy and robustness of our approach and we clearly outperform other state-of-the-art spherical matching methods on a diverse range of datasets under various conditions (like affine transformations, missing points by application of random noise models, etc.)
We are currently exploring the possibility of integrating such high-dimensional combinatorial structures into \emph{Spherical CNNs}~\cite{taco2018} to capture higher-order and latent structure. 
%Given the prevalence of omnidirectional images captured from drones, robots, and autonomous vehicles, proposing a new CNN architecture that can handle matching on any smooth surface is also a promising future research direction. 

\clearpage
\bibliographystyle{splncs04}
\bibliography{egbib}

\clearpage
\appendix

\section{Proof of Lemma 1} 
\begin{figure}
	%\makebox[\linewidth]{
		%\centering
	\vspace{-15mm}
	\makebox[\linewidth]{
		\subfigure[]{%
			\label{fig:cylinder}%
			\includegraphics[width=55mm,height=50mm]{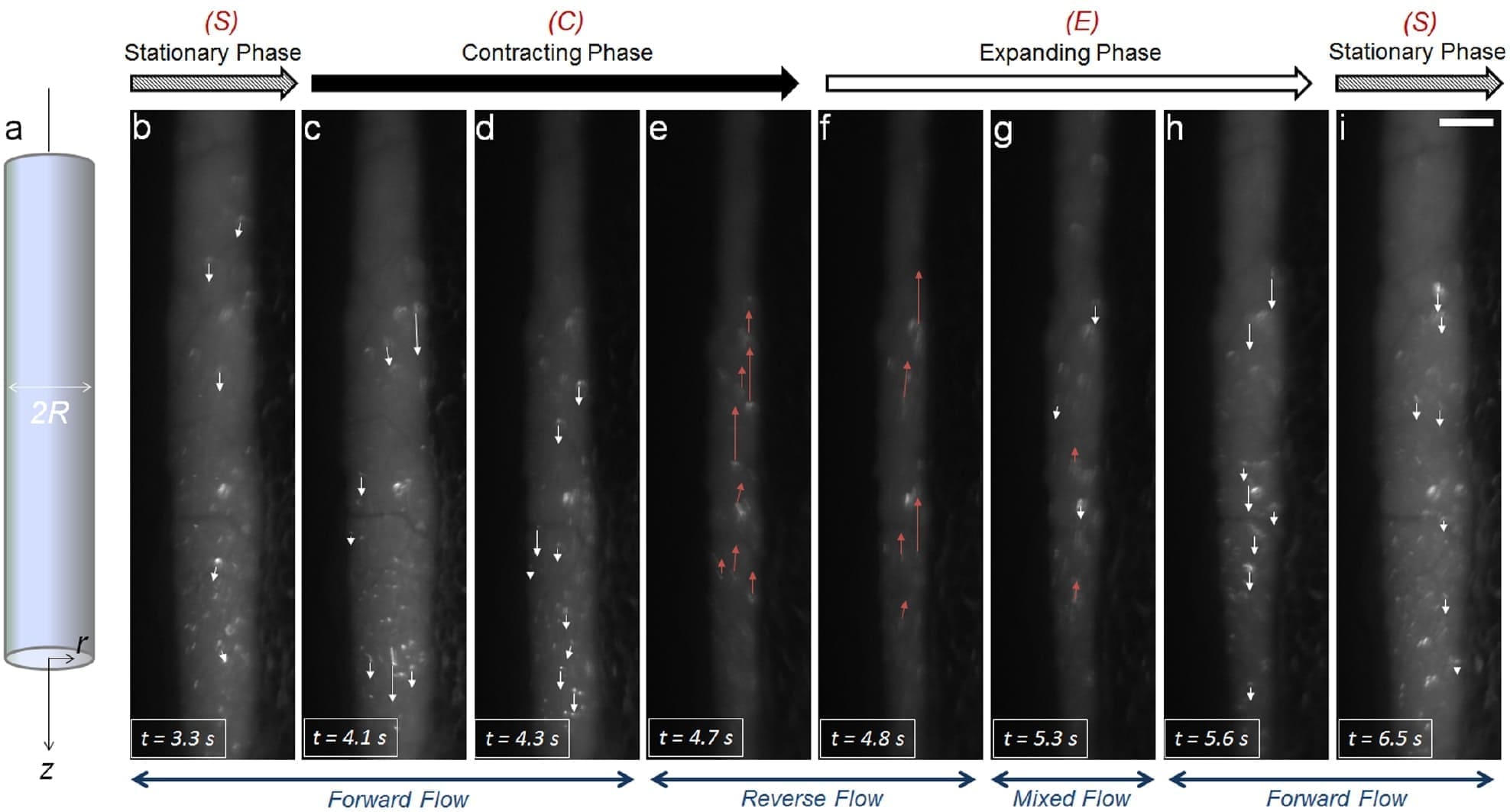}}%
		\qquad
		\subfigure[]{%
			\label{fig:fundus}%
			\includegraphics[width=55mm,height=55mm]{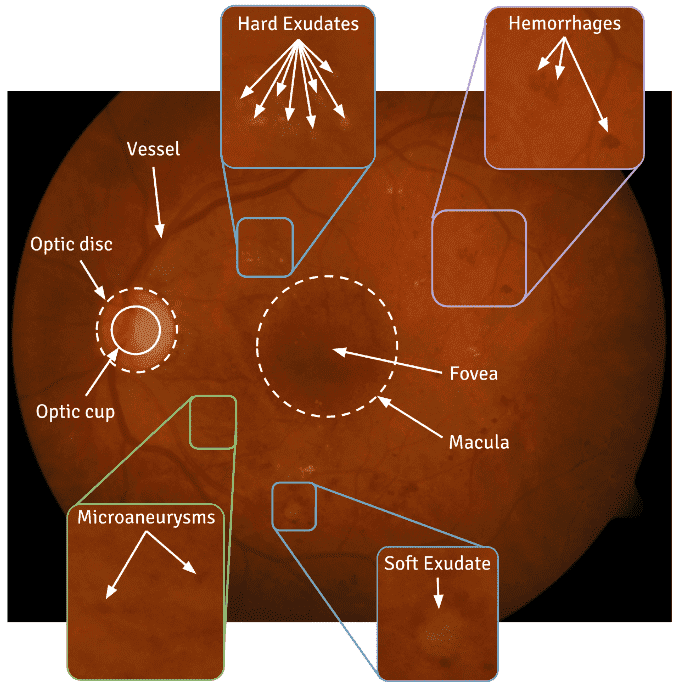}}%
	%}
	%\vspace{-0.5cm}
	%\centering
	%\vspace{-0.2cm}
	}
	\vspace{-5mm}
	\caption{Example of (a) a nanofluid microflow pattern monitored by tracking the colloidal microparticles~\cite{Sung2015}, (b) a fundus image with retinal morphologies and pathologies~\cite{sengupta2018ophthalmic}. }
	\label{fig:fundus}
\end{figure}
We begin this proof by focusing on the case where $k \geq 2$. We know that each $k$-clique 
contains $k \choose 2$ edges in it. 

Now, we study how many $k$-cliques an arbitrary edge $(u,v)$ can belong to. 
Notice that if edge $(u,v)$ belongs to a $k$-clique $K$, then all other vertices in the $K$
must also be adjacent to $u$ and vertex $u$ can have at most $\delta -1$ neighbors that 
are not $v$. Therefore, if $K$ contains both vertices $u$ and $v$, then $K$ contains 
$k-2$ other vertices and each of them must also be neighbors of $u$.
So, combining all these observations, if a $k$-clique $K$ must contain both $u$ and $v$
and $k-2$ of the other $\delta -1$ maximum allowed neighbors of $u$, then there must be
at most $\delta-1 \choose k-2$ such $k$-cliques. 

Thus, $G$ has at most $\frac{m {\delta-1 \choose k-2} }{ {k \choose 2}}$ $k$-cliques.
Then, the total number of $k$-cliques in $G$ for $k \geq 2$ is
\begin{align}
\label{eq:1}
\sum_{k=2}^{h} \frac{m {\delta-1 \choose k-2} }{ {k \choose 2}} 
&= m \sum_{k=2}^{h} \frac{2}{k(k-1)} \frac{ (\delta-1) ! }{(k-2) ! (\delta + 1 - k) !} \\
\label{eq:2}
&= m \sum_{k=2}^{h} \frac{2(\delta-1) !}{ k ! (\delta + 1 - k) ! }
\end{align} 
We need a $(\delta+1) !$ term in the numerator of Equation QAP, since we know that $h \leq (\delta+1)$. So,
\begin{align}
&m \sum_{k=2}^{h} 2 \left\lbrace \frac{  (\delta+1) ! }{ (\delta+1)(\delta) } \right\rbrace 
\frac{1}{ k ! (\delta+1-k) !  } \\
&= \frac{2m}{\delta(\delta+1)} \sum_{k=2}^{h} \frac{ (\delta+1) !  }{   k ! (\delta+1-k) !  }\\
&= \frac{2m}{\delta(\delta+1)} \left( \sum_{k=0}^{h} \frac{ (\delta+1) !  }{   k ! (\delta+1-k) !  } 
- \frac{(\delta+1) !}{1 ! \delta !} 
- \frac{(\delta+1) !}{0 ! (\delta +1) !}
\right) \\
\label{eq:3}
&= \frac{2m}{\delta(\delta+1)} \left(  \left\lbrace \sum_{k=0}^{h} { \delta+1 \choose k }        \right\rbrace  - \delta - 2         \right) \\
&\leq \frac{2m}{\delta(\delta+1)} \left[    \min \left\lbrace  (\delta+1)^h +1, \left(  \frac{e(\delta+1)}{h}         \right)^h  \right\rbrace   - \delta -2   \right]	
\end{align}

In Equation~\ref{eq:3}, the term in braces \{\}, is the partial sum of the first $h$ binomial coefficients.
Note that for $h = \delta+1$, the term would reduce to $2^{\delta+1}$. Finally, we must also account for the $n$ 1-cliques (vertices) in $G$.
This completes the proof.\qed
%The proof of Lemma~\ref{lemma:clique_count} can be found in our supplementary notes.

\section{Additional Experiments}

\subsection{Datasets}
Details of all six datasets used for warped image matching are as follows: 
$(i)$ \emph{Chessboard} dataset has $28$ spherical images in which $16$ are omnidirectional and $12$ are fish-eye images. 
The images contain a view of a room with a chessboard, chair, table, people, and computer systems. 
$(ii)$ \emph{Kamaishi} dataset has $15$ panoramic images of video frames from a moving car. 
$(iii)$ We picked $10$ panoramic images covering various scenes from \emph{SUN360} dataset.
$(iv)$ \emph{Desktop} dataset consists of two kinds of images ($7$ omnidirectional and $9$ planar). Images represent a room in which a desktop is positioned on a table. 
$(v)$ \emph{Parking} dataset also has 
omnidirectional and planar images with a view outside a house near a parking area. 
$(vi)$ \emph{Table} dataset contains omnidirectional, fish-eye, and planar images. This dataset has few tables in the 
images with computer systems, boards, chairs, etc. placed in a room. All our annotated datasets will be made publicly available.

\begin{figure}
	\centering
	\includegraphics[width=.25\textwidth]{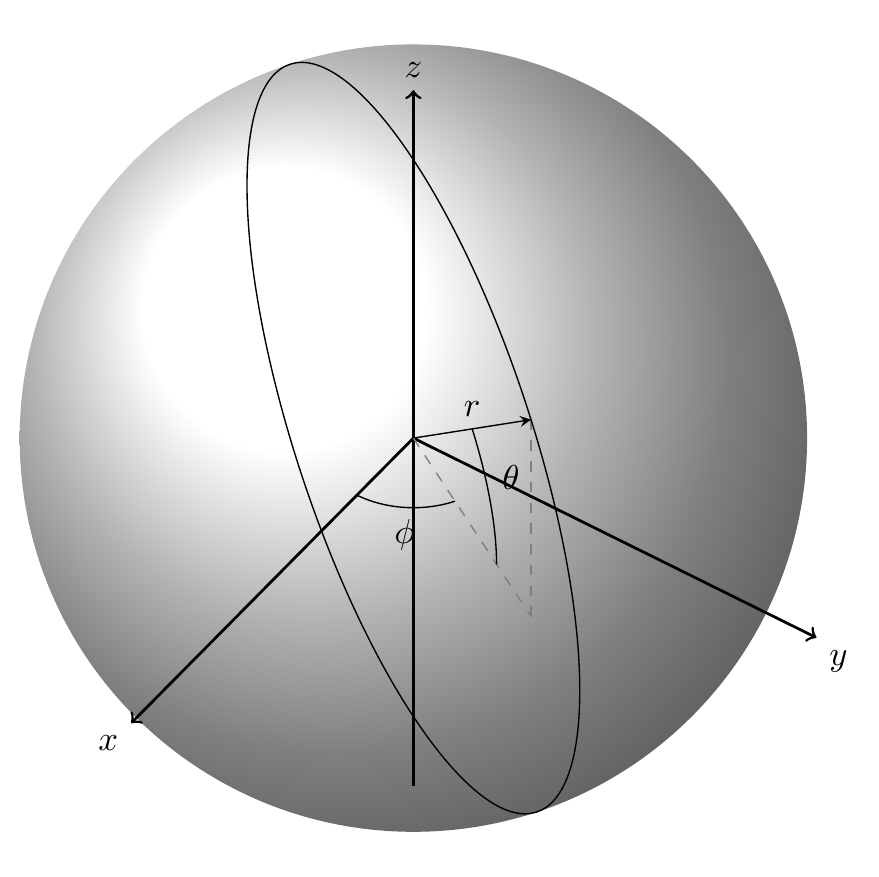}
	\caption{Parameterized $\mS^2$ (for Example 2 in main draft) }
\end{figure}

\begin{figure*}
	\makebox[\linewidth]{
	\centering
	\vspace{-10mm}
	  \subfigure[]{%
	  \label{fig:KamaishiMatch}%
	    \includegraphics[trim={0 0 0 0},clip,width=55mm,height=30mm]{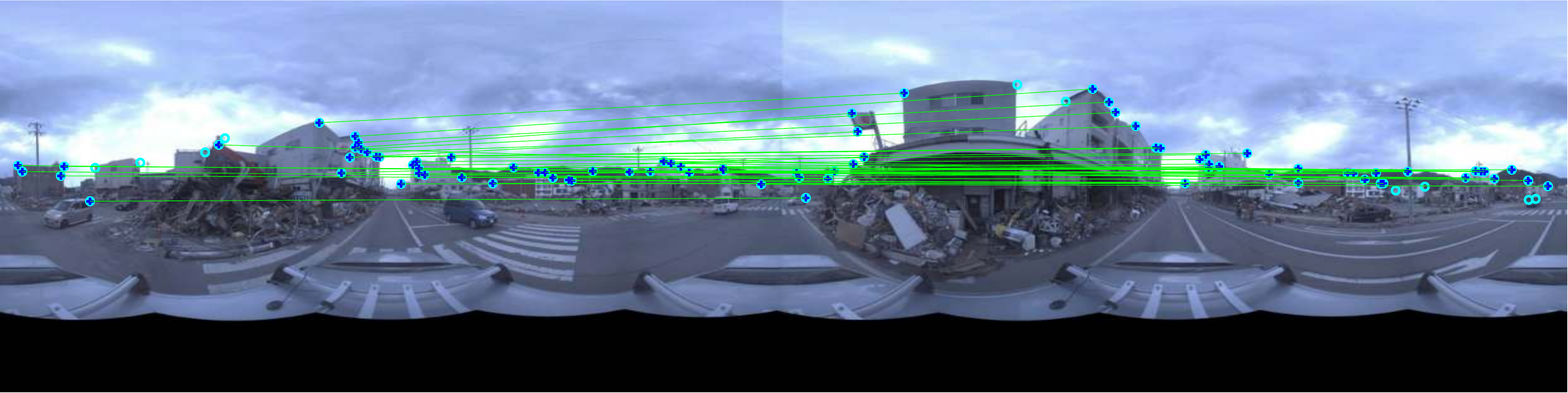}}%
	  %\caption{a) Matchings for house frame $1$ (left) with frame $55$ (right), b) Matchings for house frame $1$ (left) with frame $110$ (right), Error: $0.0\%$.}
	  \qquad
	  %\hspace{0.5mm}
	  \subfigure[]{%
	  \label{fig:KamaishiMatch2}%
	    \includegraphics[trim={0 0 0 0},clip,width=55mm,height=30mm]{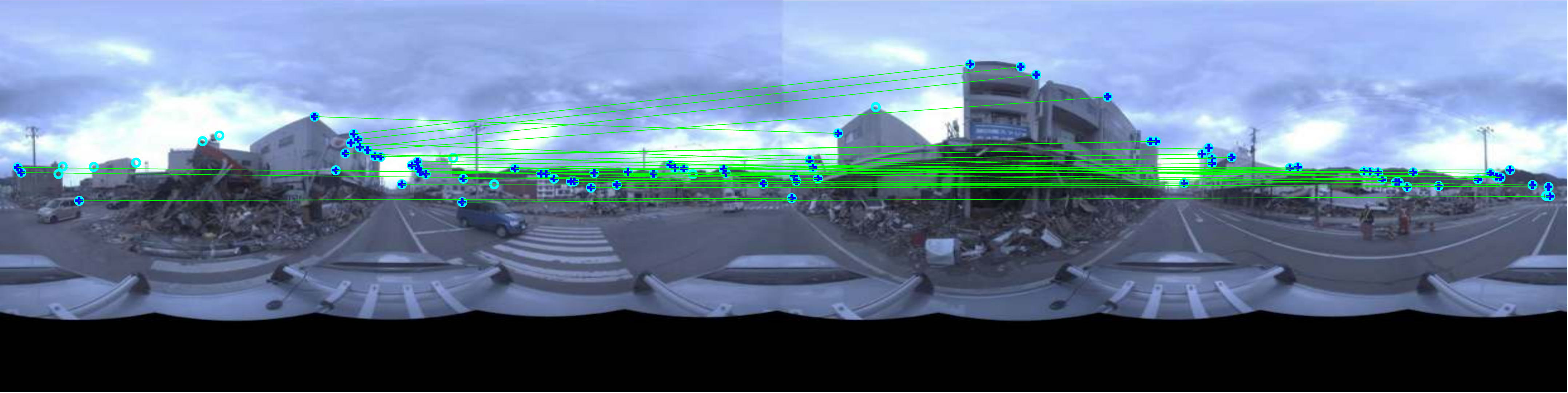}}%
	    %\vspace{-5cm}
	  }
	  \vspace{-2mm}
	  \qquad
	  \makebox[\linewidth]{
	  \centering
	  \hspace{-17mm}
	  \subfigure[]{%
	  \label{fig:HorseRotate}%
	    \includegraphics[trim={0 0 0 0},clip,width=55mm,height=28mm]{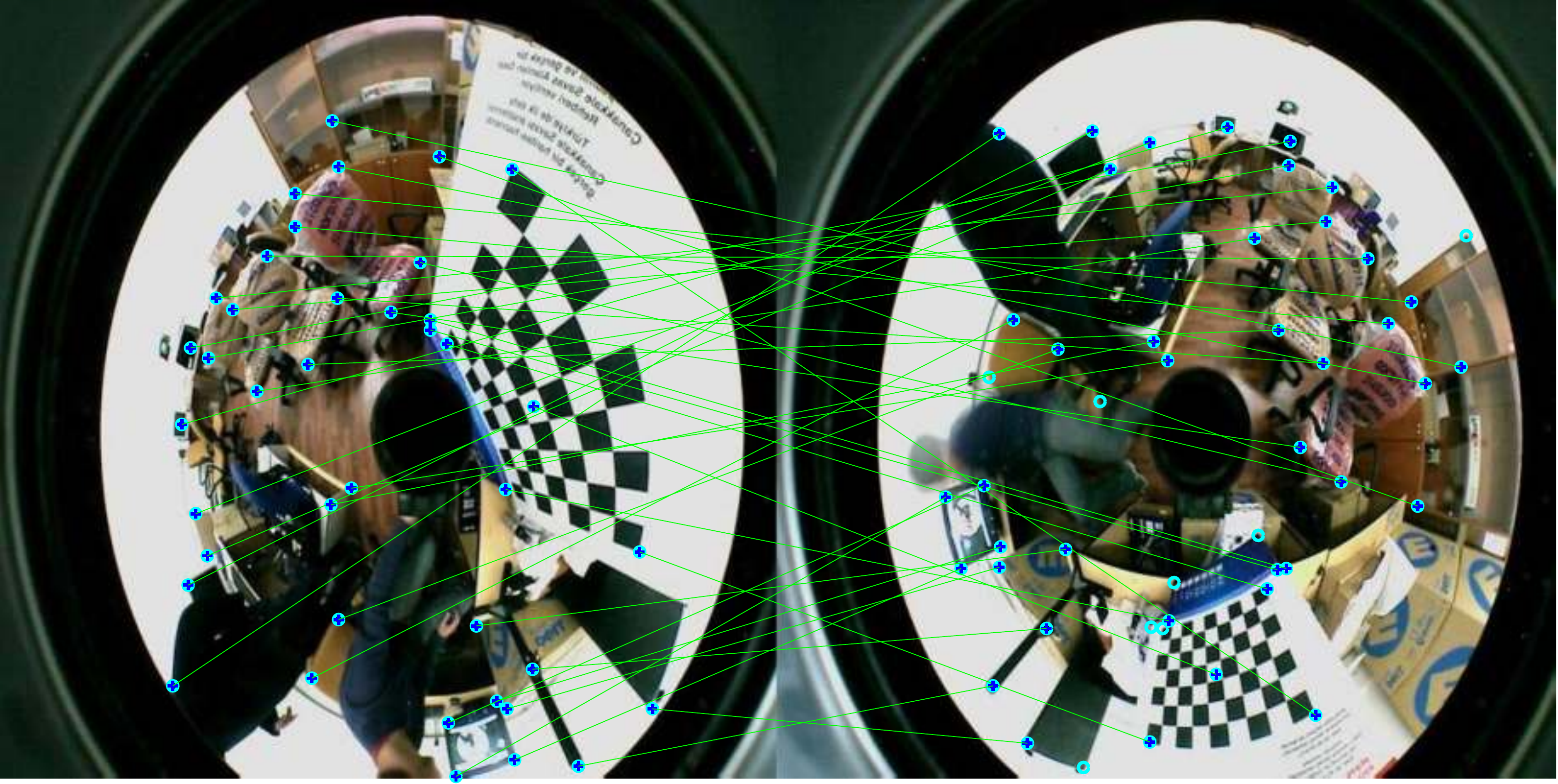}}%
	  \qquad
	  \subfigure[]{%
	  \label{fig:HorseShear}%
	    \includegraphics[trim={0 0 0 0},clip,width=55mm,height=28mm]{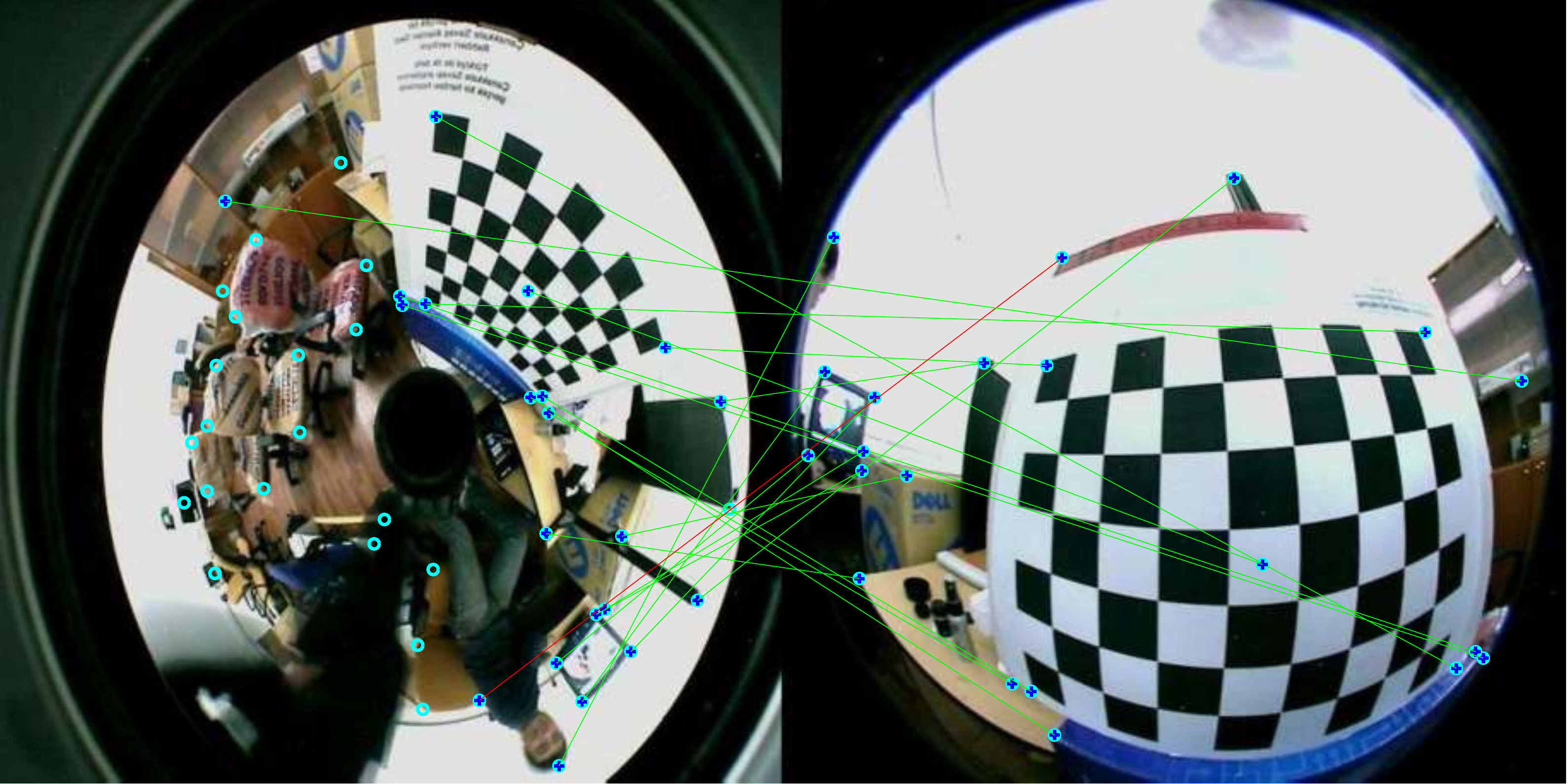}}%
	  }
	  \vspace{-2mm}
	  \qquad
	  \makebox[\linewidth]{
	  \centering
	  \hspace{-17mm}
	  \subfigure[]{%
	  \label{fig:Car}%
	    \includegraphics[width=55mm,height=28mm]{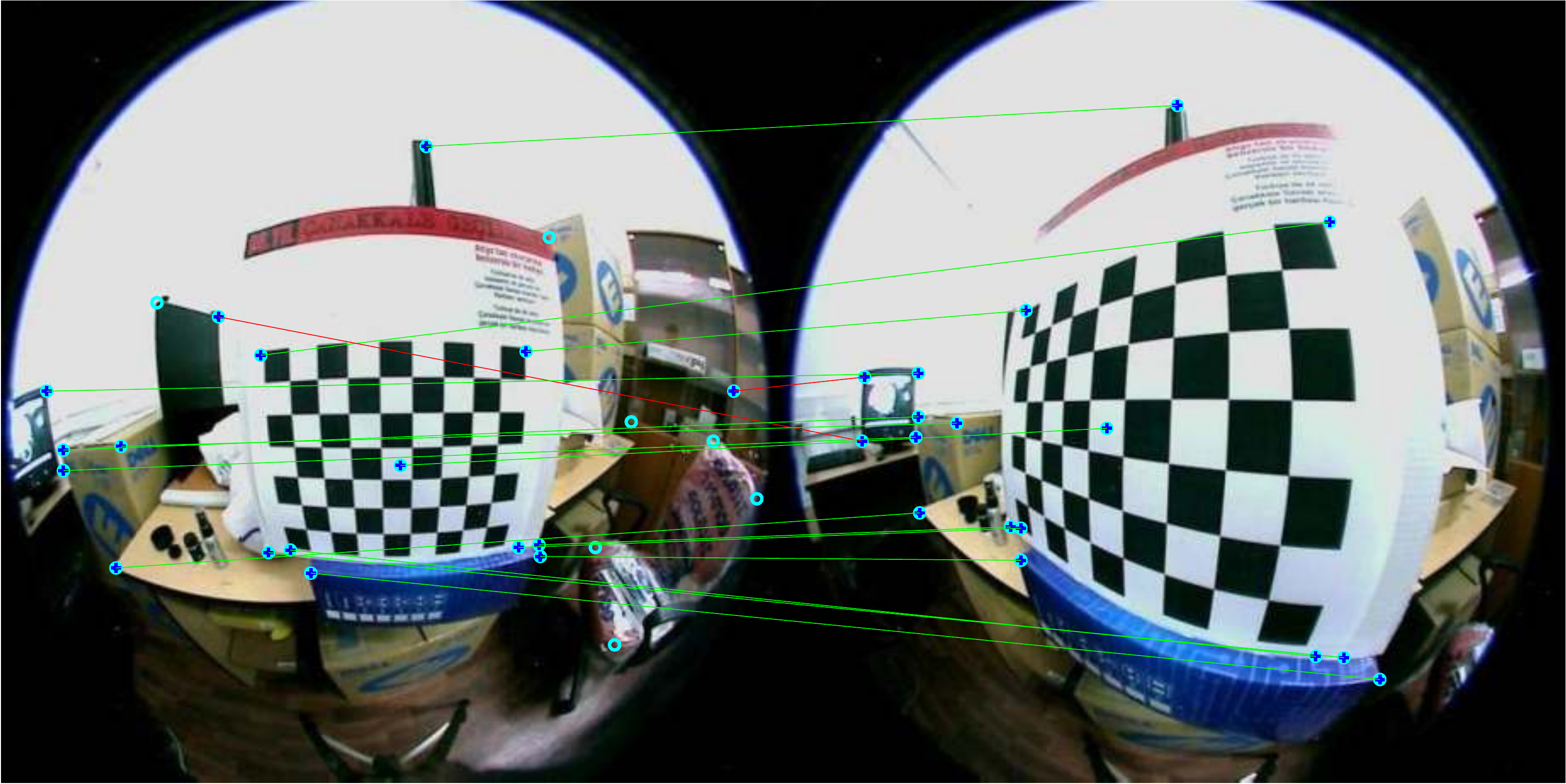}}%
	  \qquad
	  \subfigure[]{%
	  \label{fig:Bike}%
	    \includegraphics[width=55mm,height=28mm]{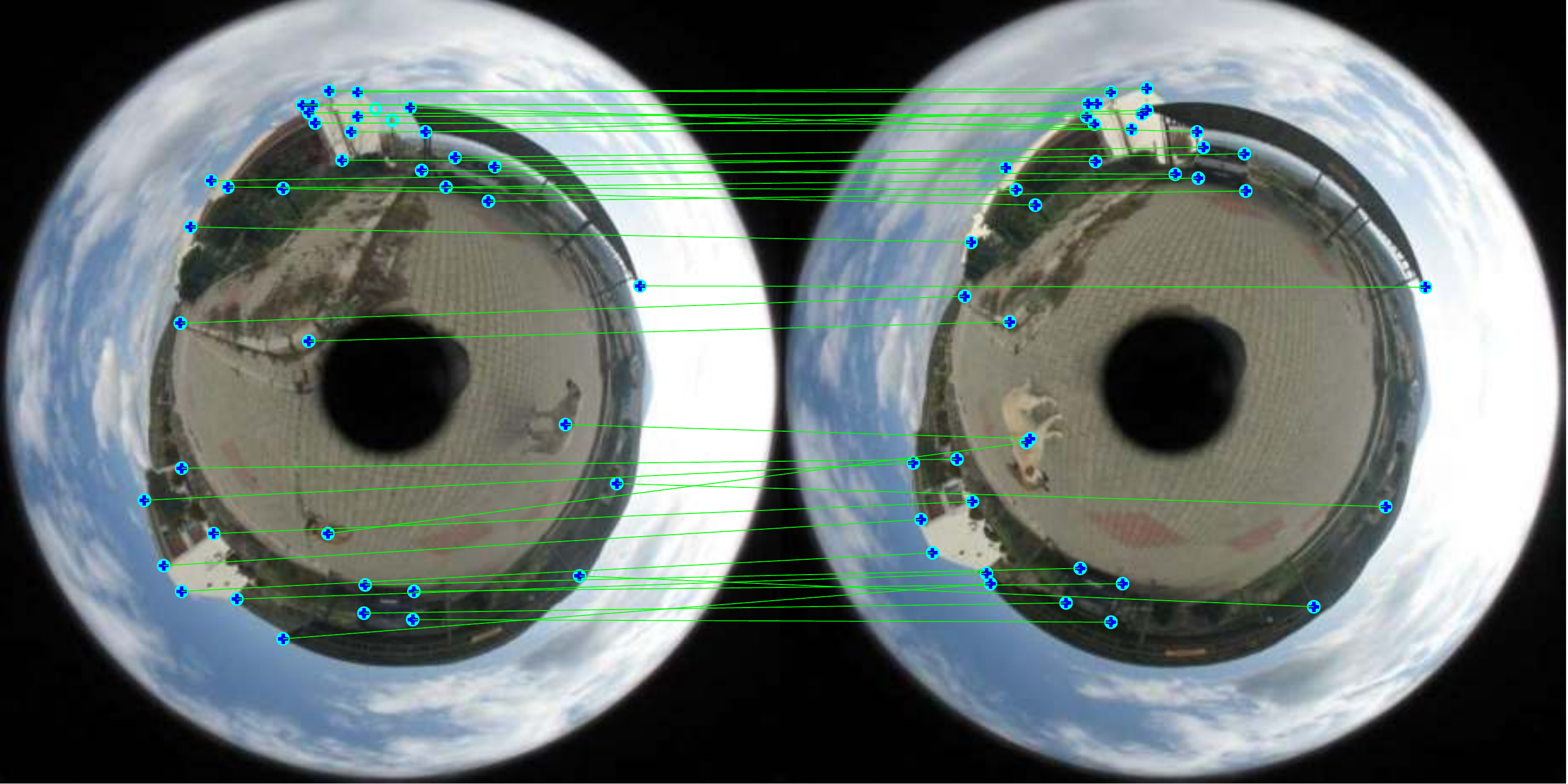}}%
	  }
	  \vspace{-2mm}
	  \qquad
	  \makebox[\linewidth]{
	  \centering
	  \hspace{-17mm}
	  \subfigure[]{%
	  \label{fig:Butterfly}%
	    \includegraphics[width=55mm,height=28mm]{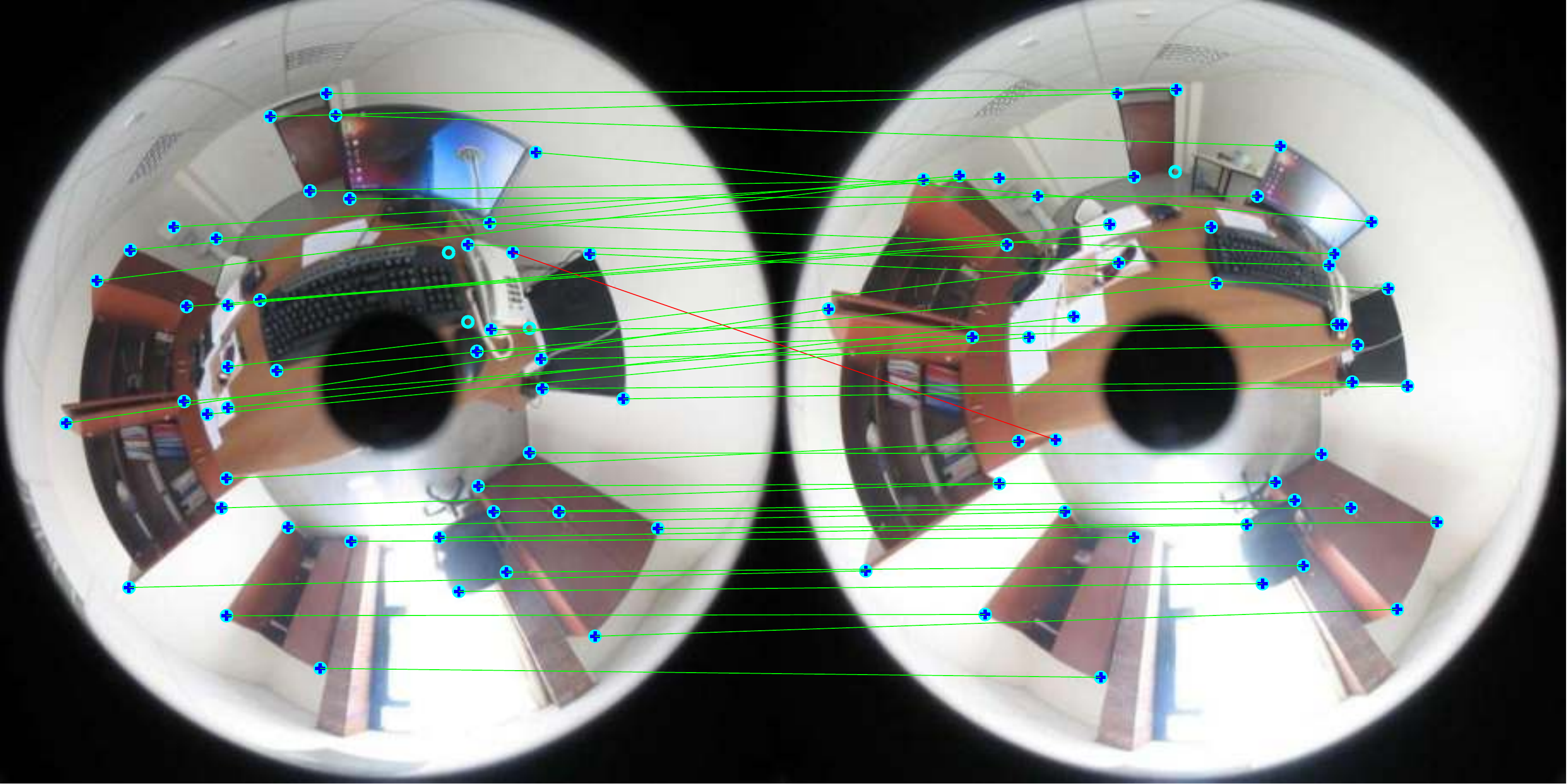}}%
	  \qquad
	  \subfigure[]{%
	  \label{fig:Spectrum}%
	    \includegraphics[width=55mm,height=28mm]{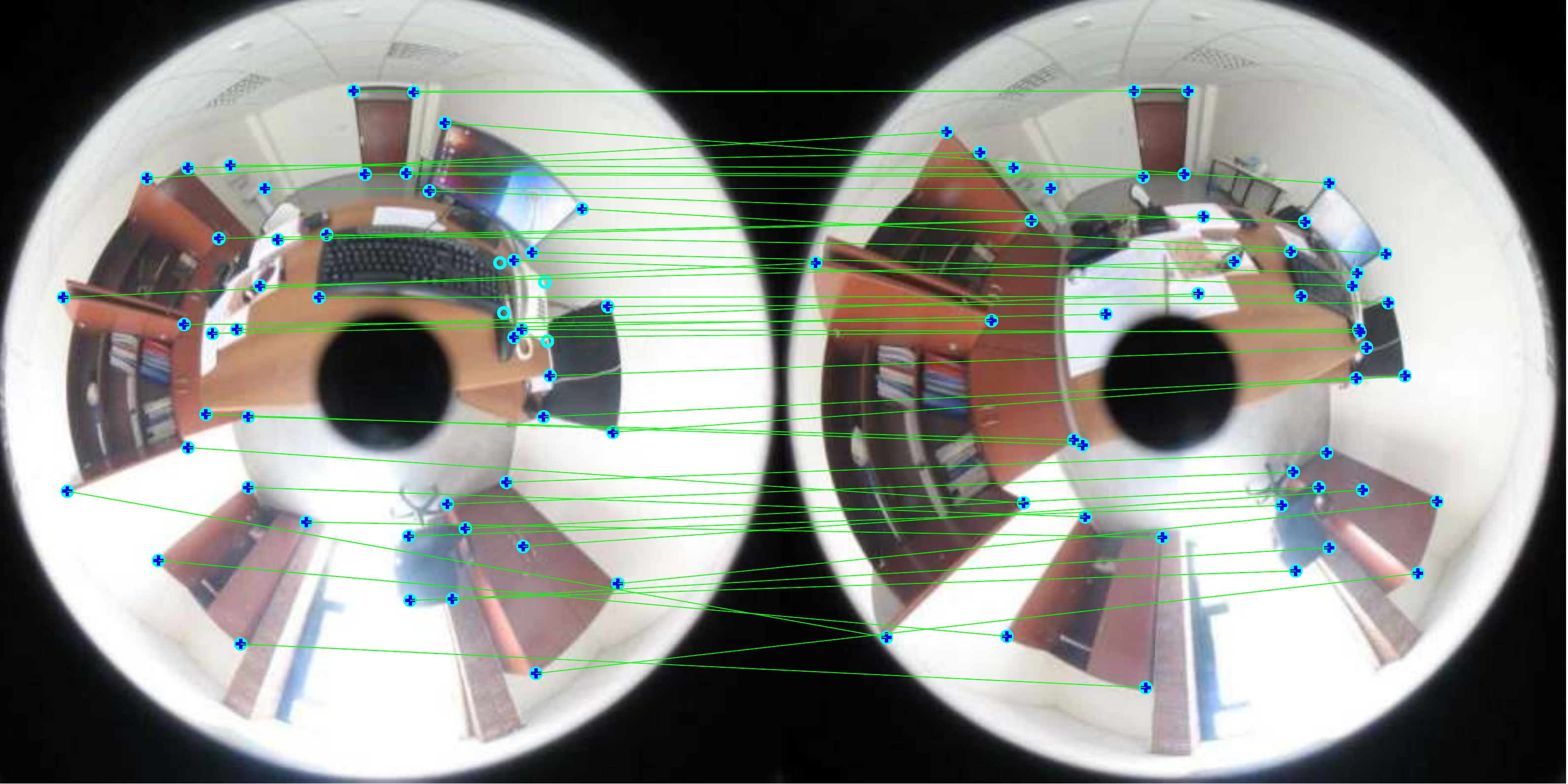}}%
	  }
	  \vspace{-2mm}
	  \qquad
	  \hspace{-9mm}
	  \makebox[\linewidth]{
	  \centering
	  \subfigure[]{%
	  \label{fig:Bldg}%
	    \includegraphics[trim={0 0 0 0},clip,width=55mm,height=28mm]{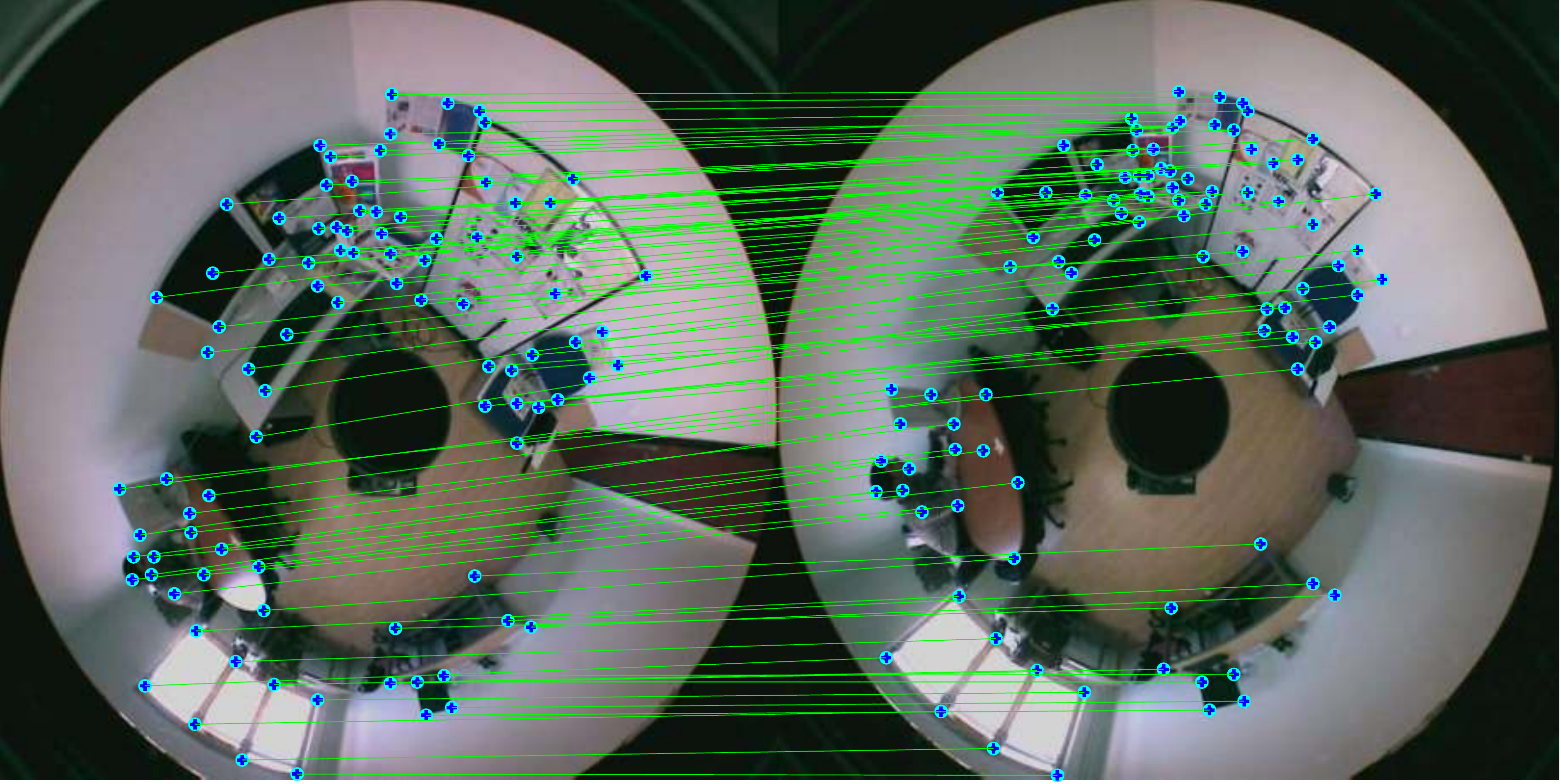}}%
	  \qquad
	  \subfigure[]{%
	  \label{fig:Book}%
	    \includegraphics[trim={0 0 0 0},clip,width=55mm,height=28mm]{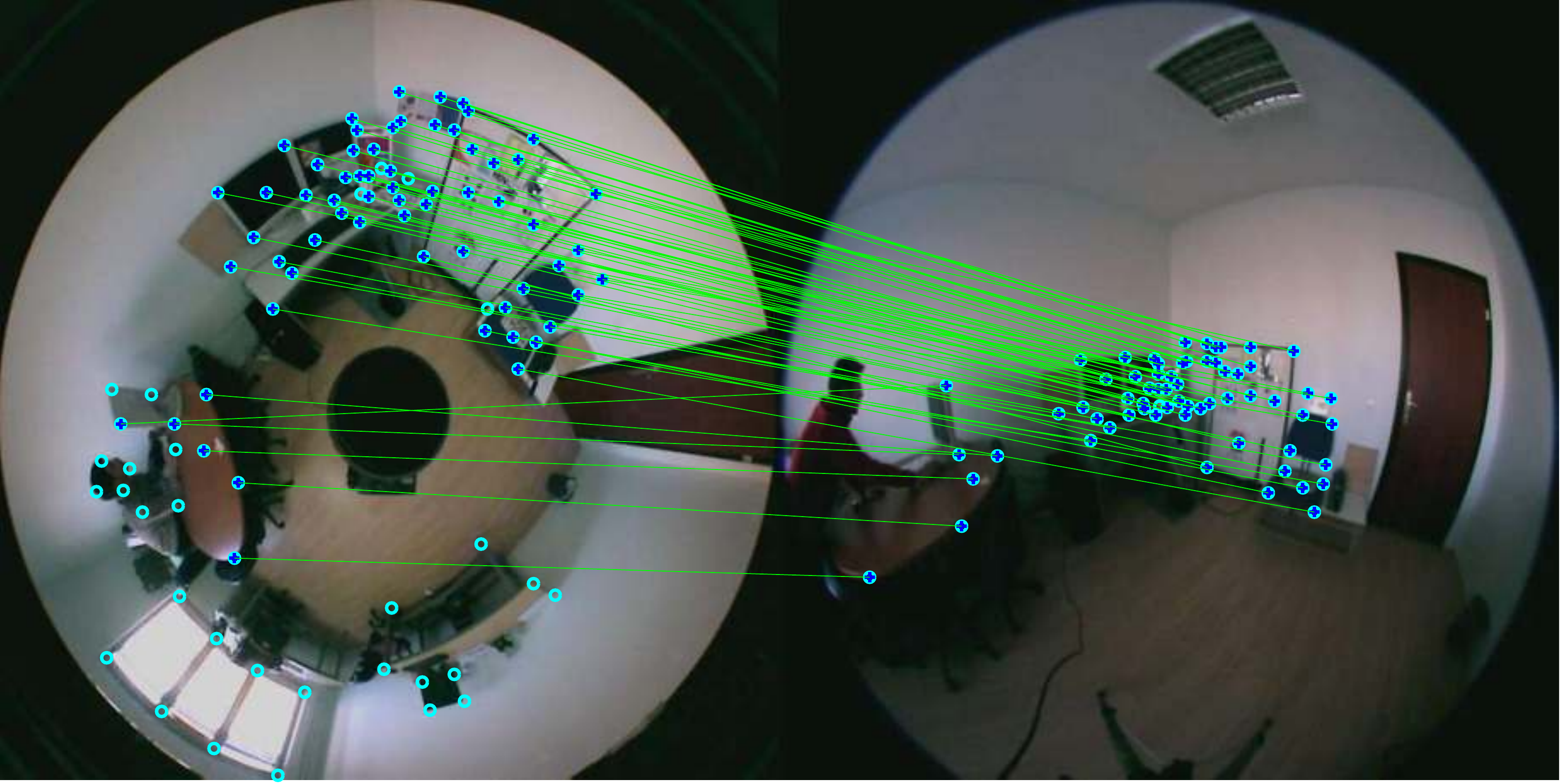}}%
	  }
	  
	  %\vspace{-2mm}
	  \caption{Instances of matchings between spherical images of $(a)-(b)$ Kamaishi, $(c)-(e)$ Chessboard, $(f)$ Parking, $(g)-(h)$ Desktop and $(i)-(j)$ Table datasets. Green/red lines show correct/incorrect matches respectively. Isolated points show no matches.}
	  \label{fig:Matching}
\end{figure*}

In addition to experiments in our main paper, we conducted additional experiments for different settings. In \emph{multi-modal matching}, we performed matching on \emph{unwrapped-unwrapped spherical/warped} images and 
\emph{unwrapped spherical/warped vs planar} images. In \emph{ablative studies}, we perfomed experiments considering two noise models and geometric verification and rectification. Additional experiments on rotation are performed for SUN360 dataset and Desktop dataset. We also analyzed the effect of missing 
points on four datasets (Chessboard, Desktop, Kamaishi and Table). Additionally, we also studied effects of varying $k$ values in a neighborhood of landmark points, varying the radius of the sphere and manifold, and varying the number of landmark points for matching. Matching images for different cases are shown in Figures~\ref{fig:Matching} and~\ref{fig:MatchingSP}. We also created a video of frames of Kamaishi dataset where we match points of 
frames in a sequence which can be useful for tracking purposes. The video titled \emph{Kamaishi\_Panorama\_Matching\_Video.avi} is located in the folder of our supplementary material.
\begin{table}
	\centering
	%\vspace{-1mm}
	\caption{Error (\%) of pairwise matching between spherical (omnidirectional and fish-eye) and planar images of three datasets for different methods.}
	\vspace{1mm}
	%\tiny
	\footnotesize
	\setlength\tabcolsep{3.5pt}
	\begin{tabular}{llll}
		\toprule
		\textbf{Algorithms} & \textbf{Desktop} & \textbf{Parking} & \textbf{Table} \\
		\hline
		\emph{OurWarped\_Cone} & 14.71 $\pm$ 0.0 \%& 4.41 $\pm$ 0.0 \%& 0.88 $\pm$ 0.0 \%\\
		\emph{OurWarped\_Ellip} & 13.82 $\pm$ 0.0 \%& 5.94 $\pm$ 0.0 \%& 2.65 $\pm$ 0.0 \%\\
		\hline
		%BRISKS &  & & \\ 
		\emph{OurSpherical} & \textbf{15.16} $\pm$ \textbf{0.0} \%& \textbf{5.53} $\pm$ \textbf{0.0} \%& \textbf{0.78} $\pm$ \textbf{0.0} \%\\
		FGM + geodesic & 97.94 $\pm$ 0.24 \%& 97.53 $\pm$ 0.57 \%& 96.98 $\pm$ 2.51 \%\\
		SPHORB & 41.5 $\pm$ 0.0  \%& 53.85 $\pm$ 0.0 \%& 70.88 $\pm$ 0.0 \%\\ 
		%SIFTS &  & & \\
		BRISK & 37.84 $\pm$ 0.0 \%& 51.05 $\pm$ 0.0 \%& 65.11 $\pm$ 0.0 \%\\
		ORB & 34.82 $\pm$ 0.0 \%& 48.82 $\pm$ 0.0 \%& 62.78 $\pm$ 0.0 \%\\
		%SIFT &  & & \\
		Tensor & 93.85 $\pm$ 0.0 \%& 94.43 $\pm$ 0.22 \%& 56.94 $\pm$ 1.02 \%\\
		FGM & 98.0 $\pm$ 0.0 \%& 97.50 $\pm$ 0.01 \%& 97.5 $\pm$ 0.01 \%\\
		\bottomrule
	\end{tabular}
	%\vspace{-7mm}
	\label{tab:PS}
\end{table}
\subsection{Matching on Spherical/Warped vs. Planar }
To understand the effects of matching spherical / warped images versus planar images,  
we conducted matching experiments between planar and warped images, whose results 
are shown in the first two rows of Table~\ref{tab:PS}. We find that the matching error
is higher for warped-planar matching compared to warped-warped matching shown in Table in main paper
because of the difference in geodesic distances between landmarks embedded on the warped versus
the Euclidean distance between landmarks on planar images.

%due to fewer points on planar images, which does not allow for the formation of higher-dimensional cliques.
Next, planar images are compared to both fish-eye and omnidirectional images. 
The results are shown in Table~\ref{tab:PS} in which we clearly observe an increase in error for the
Desktop dataset when compared to the omni-omni matching results in Table in main paper. As planar images cover
a much smaller portion of the scenes as compared to omnidirectional images, they also contain much fewer 
landmark points with many occluded regions, which in turn affects our method due to the lack of higher-dimensional cliques. However, our method still beats other state-of-the-art methods. 
%We also conducted additional experiments: (1) comparing against baselines using \emph{stereo rectification} and \emph{geometric verification} and (2) the effect of various \emph{noise models}
%and found our method still outperforming others. For brevity, we placed the results of these experiments in our 
%supplementary notes.
%We also conducted additional comparisons against methods using \emph{stereo rectification} and \emph{geometric verification}  (results in supplementary notes), but found our method significantly outperforming them.
%Desktop dataset has the highest error among the three datasets. This is due to less number of points present in planar images since planar 
%images cover only a small part of the whole scene whereas the omnidirectional image covers the complete ($360^{\circ}$) view of the scene. Since occlusion has already been a challenge in state-of-the-art 
%of matching problem. In addition to the distortion from omnidirectional camera, matching 
%would also be affected by severe occlusion and missing points between the spherical and planar images. 

\subsection{Matching on Planar vs. Unwrapped Spherical/Warped}
We observe that matching between spherical and planar is a challenge due to the difference in euclidean and geodesic distances in planar and spherical images, respectively, as is evidenced in Table $3$ of our main paper. Therefore, to 
compare these kinds of images, we can perform matching between unwrapped equirectangular spherical/warped and planar images. In this way, we can also apply state-of-the-art graph matching methods which 
perform well on planar images. Thus, we consider three datasets with both spherical (omnidirectional and fish-eye) and planar images and conducted the experiment. Table~\ref{tab:PP2} shows the results. We can compare the results with 
Table $3$ in the main paper, we can clearly see that the results are almost same even after flattening of the image. Thus, comparing flattened and planar images does not reduce the error of matchings 
between spherical and planar images. However, our method outperforms the existing methods.

\begin{table}
		\centering
		\vspace{-4mm}
		\caption{Error (\%) of pairwise matching between unwrapped equirectangular version of spherical (omnidirectional and fish-eye) images and planar images of three datasets for different methods including graph matching methods on flat surfaces.}
		%\vspace{1mm}
		\footnotesize
		\begin{tabular}{llll}
			\toprule
			\textbf{Algorithms} & \textbf{Desktop} & \textbf{Parking} & \textbf{Table} \\
			\hline
			\hline
			\emph{OurWarped\_Cone} & 14.26 $\pm$ 0.0 \%& 5.23 $\pm$ 0.0 \%& 1.09 $\pm$ 0.0 \%\\
			\emph{OurWarped\_Ellip} & 15.16 $\pm$ 0.0 \%& 3.59 $\pm$ 0.0 \%& 2.96 $\pm$ 0.0 \%\\
			\hline
			\emph{OurSpherical} & \textbf{13.08 $\pm$ 0.0} \%& \textbf{4.10 $\pm$ 0.0} \%& \textbf{0.93 $\pm$ 0.0} \%\\
			
			RCC & 30.85 $\pm$ 1.78 \%& 12.38 $\pm$ 1.53 \%& 11.69 $\pm$ 1.58 \%\\ 
			EigenAlign & 99.94 $\pm$ 0.0 \%& 99.22 $\pm$ 0.0 \%& 99.25 $\pm$ 0.0 \%\\ 
			Tensor & 97.88 $\pm$ 0.11 \%& 95.59 $\pm$ 0.35 \%& 81.04 $\pm$ 1.43 \%\\
			FGM & 99.0 $\pm$ 0.0 \%& 100.0 $\pm$ 0.0 \%& 97.0 $\pm$ 0.0 \%\\ 
			SPHORB & 41.33 $\pm$ 0.0 \%& 59.9 $\pm$ 0.0 \%& 71.15 $\pm$ 0.0 \%\\ 
			%SIFTS &  & & \\
			BRISK & 37.67 $\pm$ 0.0 \%& 55.98 $\pm$ 0.0 \%& 67.8 $\pm$ 0.0 \%\\
			ORB & 34.43 $\pm$ 0.0 \%& 45.03 $\pm$ 0.0 \%& 64.84 $\pm$ 0.0 \%\\
			%SIFT &  & & \\

			\bottomrule
		\end{tabular}
		%\vspace{-5mm}
		
		\label{tab:PP2}
\end{table}

\begin{figure}
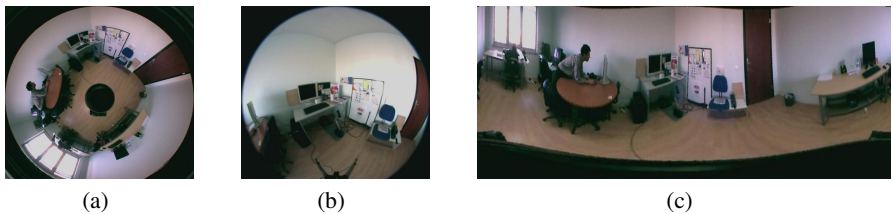

	%\makebox[\linewidth]{
		%\centering
	\makebox[\linewidth]{
		\subfigure[]{%
			\label{fig:zero11}%
			\includegraphics[width=25mm,height=23mm]{Experiments/Images/3.jpg}}%
		\qquad
		\subfigure[]{%
			\label{fig:first11}%
			\includegraphics[width=25mm,height=23mm]{Experiments/Images/7.jpg}}%
	%}
	%\vspace{-0.5cm}
	%\centering
	%\vspace{-0.2cm}
	\qquad
	\subfigure[]{%
		\label{fig:second11}%
		\includegraphics[width=55mm,height=23mm]{Experiments/Images/3_flat.jpg}}%
	}
	\vspace{-5mm}
	\caption{Example of (a) a parabolic omnidirectional image, (b) fish-eye, and (c) panoramic (unwrapped equirectangular) images of the same view. }
	\label{fig:ImageTransform}
\end{figure}
\begin{figure}
	\centering
	%\hspace{-17mm}
	\makebox[\linewidth]{
	\subfigure[]{%
		\label{fig:Desktop2}%
		\includegraphics[trim={0 0 0 0},clip,width=55mm,height=20mm]{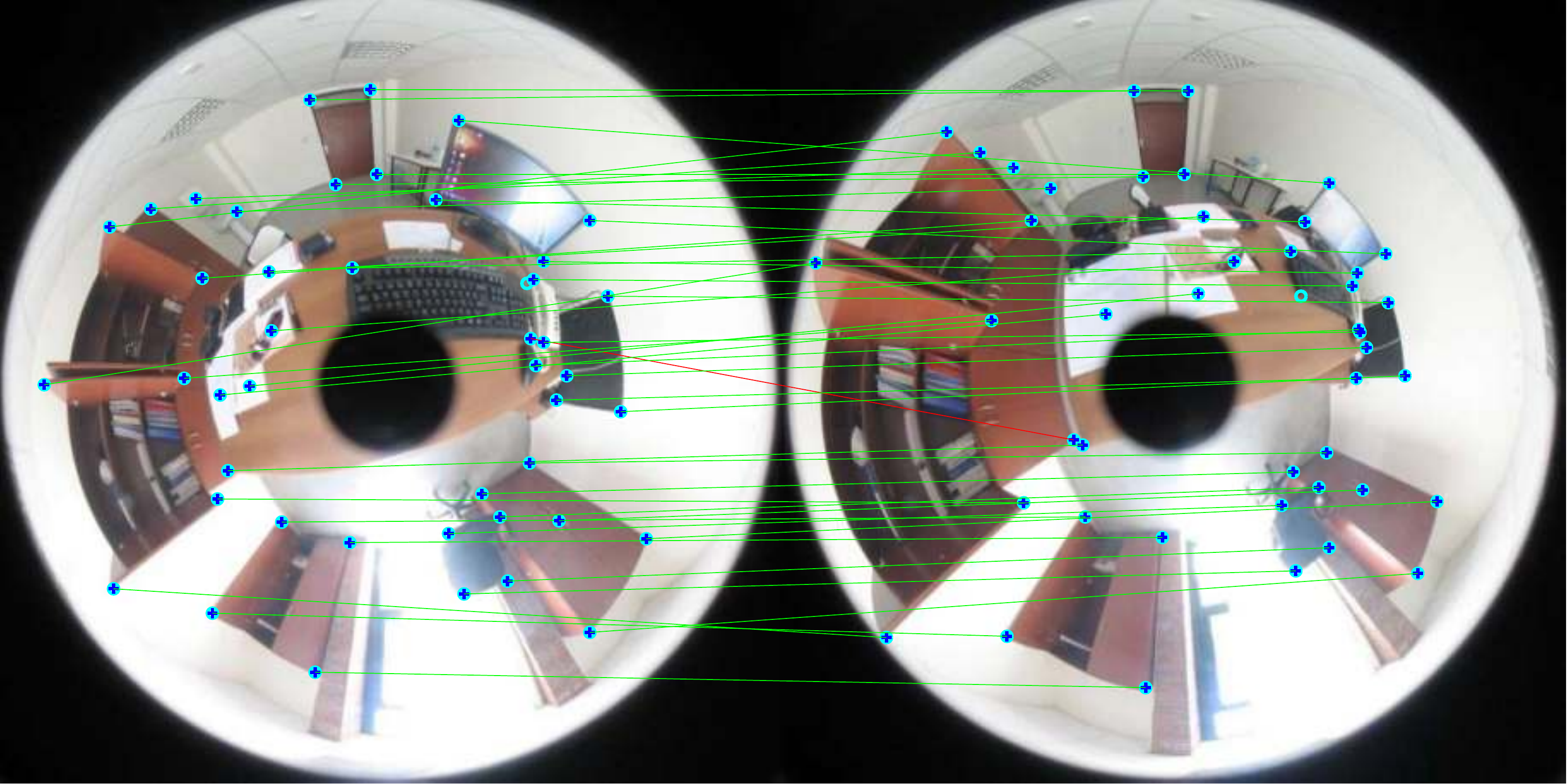}}%
	%\vspace{-0.2cm}
	\qquad
	\subfigure[]{%
		\label{fig:Table1}%
		\includegraphics[trim={0 0 0 0},clip,width=55mm,height=20mm]{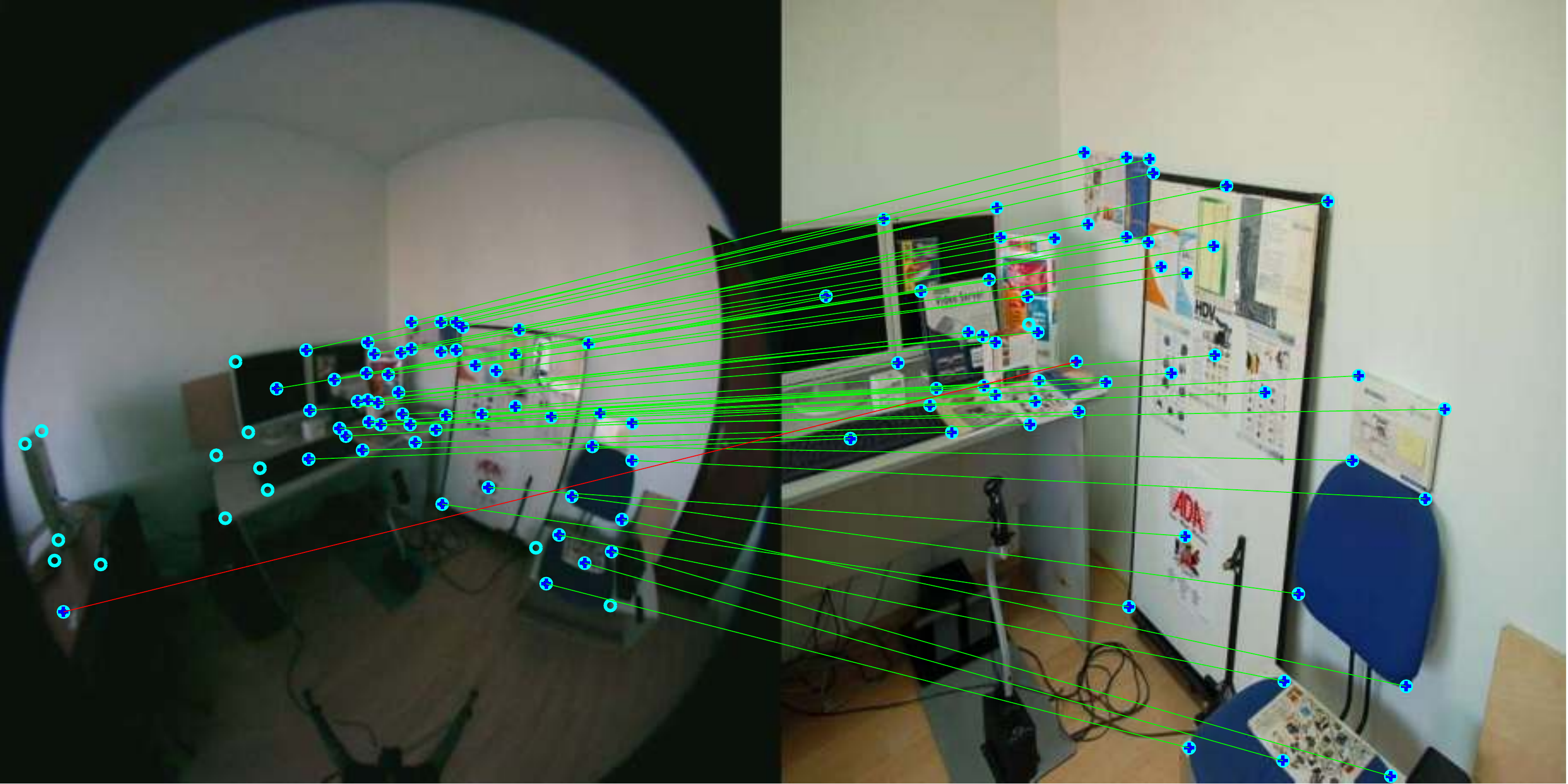}}%
	}
	\vspace{-5mm}
	\caption{Instances of matchings between (a) \emph{Desktop} omnidirectional images and (b) 
		\emph{Table} fish-eye and planar images. Green/red lines show correct/incorrect matches respectively. Isolated points show no matches.}
	\label{fig:MatchingSP}
\end{figure}

\subsection{Effect of Noise Models }

%\noindent\textbf{Effect of Noise Models: }\label{mathrefs}
Our proposed matching method is subjected to two random noise models proposed by Feizi et. al.~\cite{feizi2016spectral}.
While matching two 
simplicial complexes $\mM$ and $\mM'$, noise is introduced in the first simplicial complex $\mM$. 
We denote the \emph{noisy simplicial complex} as $\tilde{\mM}$.
Next, we attempt a matching between 
$\tilde{\mM}$ and $\mM'$. 
The noisy complex $\tilde{\mM}$ is generated as follows:
%\textbf{Noise Model I:}
\begin{equation} \label{eq:n1} \nonumber
\textbf{Noise Model I: }\hspace{0.7mm} \tilde{\mM} = \mM \odot (1-P) + (1-\mM) \odot P
%\label{eq:n1}
\end{equation}  
%\textbf{Noise Model II:}
\begin{equation} \label{eq:n2} \nonumber
\textbf{Noise Model II: }\hspace{0.5mm} \tilde{\mM} = \mM \odot (1-P) + (1-\mM) \odot Q
\end{equation}
\begin{table}
	\centering
	%\vspace{-5mm}
	\caption{Error (\%) of pairwise matching between unwrapped spherical (omnidirectional and fish-eye) images of four datasets for different graph matching methods with two noise models.}
	%\vspace{1mm}
	\footnotesize
	\setlength\tabcolsep{2.5pt}
	\begin{tabular}{lllll}
		\toprule
		\textbf{Algorithms} & \textbf{Chessboard} & \textbf{Desktop} & \textbf{Parking} & \textbf{Table} \\
		\hline
		\hline
		%\vspace{1mm}
		\multicolumn{1}{l}{} & \multicolumn{4}{G}{\textbf{Noise Model I}}\\
		\hline 
		\emph{OurMethod} & \textbf{4.17} $\pm$ \textbf{0.0} \%& \textbf{0.85} $\pm$ \textbf{0.0} \%& \textbf{0.0} $\pm$ \textbf{0.0} \%& \textbf{0.31} $\pm$ \textbf{0.0} \%\\
		RCC & 29.7 $\pm$ 0.68 \%& 17.1 $\pm$ 1.17 \%& 15.8 $\pm$ 7.89 \%& 16.2 $\pm$ 0.48 \%\\ 
		EigenAlign & 99.5 $\pm$ 0.0 \%& 96.5 $\pm$ 0.0 \%& 97.7 $\pm$ 0.0 \%& 98.2 $\pm$ 0.0 \%\\ 
		%Tensor & 68.98 $\pm$ 0.16  & 26.08 $\pm$ 0.58 & 19.0 $\pm$ 3.75 & 72.35 $\pm$ 0.67\\
		FGM & 93.0 $\pm$ 0.0 \%& 72.0 $\pm$ 0.0 \%& 65.0 $\pm$ 0.0 \%& 92.0 $\pm$ 0.0 \%\\ 
		%SPHORB & 58.61 $\pm$ 0.0 & 90.28 $\pm$ 0.0 & 97.5 $\pm$ 0.0 & 79.18 $\pm$ 0.0\\ 
		%SIFTS &  & & & \\
		%BRISK & 54.91 $\pm$ 0.0 & 84.91 $\pm$ 0.0 & 100.0 $\pm$ 0.0 & 74.16 $\pm$ 0.0 \\
		%ORB & 49.48 $\pm$ 0.0 & 78.23 $\pm$ 0.0 & 82.5 $\pm$ 0.0 & 70.25 $\pm$ 0.0 \\
		%SIFT &  & & & \\
		\hline
		\multicolumn{1}{l}{} & \multicolumn{4}{G}{\textbf{Noise Model II}}\\
		\hline
		\emph{OurMethod} & \textbf{4.36} $\pm$ \textbf{0.0} \%& \textbf{0.64} $\pm$ \textbf{0.0} \%& \textbf{0.0} $\pm$ \textbf{0.0} \%& \textbf{0.44} $\pm$ \textbf{0.0} \%\\
		RCC & 29.9 $\pm$ 0.52 \%& 17.0 $\pm$ 1.01 \%& 14.2 $\pm$ 5.26 \%& 16.1 $\pm$ 0.42 \%\\ 
		EigenAlign & 98.9 $\pm$ 0.0 \%& 97.2 $\pm$ 0.0 \%& 98.5 $\pm$ 0.0 \%& 99.1 $\pm$ 0.0 \%\\ 
		%Tensor & 68.98 $\pm$ 0.16  & 26.08 $\pm$ 0.58 & 19.0 $\pm$ 3.75 & 72.35 $\pm$ 0.67\\
		FGM & 94.0 $\pm$ 0.0 \%& 71.0 $\pm$ 0.0 \%& 60.0 $\pm$ 0.0 \%& 92.5 $\pm$ 0.0 \%\\ 
		\bottomrule
	\end{tabular}
	%\vspace{-5mm}
	\label{tab:Noise}
\end{table}
where, $\odot$ represents the \emph{element-wise multiplication} with matrices $P$ / $Q$ which are \emph{binary random symmetric matrices} drawn from a Bernoulli distribution with $v$ vertices and $p$ / $q$ probabilities, respectively. Here, $P$ \emph{flips} the \emph{vertex-vertex adjacency} with probability $p$ and $Q$ \emph{adds new edges} between non-connected vertices with probability $q$ to the underlying geometric graph $G=(V,E,g)$ of the simplicial complex $\mM$. Experiments are performed between pairwise flattened (unwrapped) spherical images of four datasets against graph matching methods. Table~\ref{tab:Noise} shows matching results after the application of both noise models and it 
shows that our method is the most robust.

%\newpage
\begin{figure*}[h]
    \centering
    %\vspace{-5mm}
    \makebox[\linewidth]{
    \subfigure[]{%
    \label{fig:affine1}%
    \includegraphics[width=35mm,height=30mm]{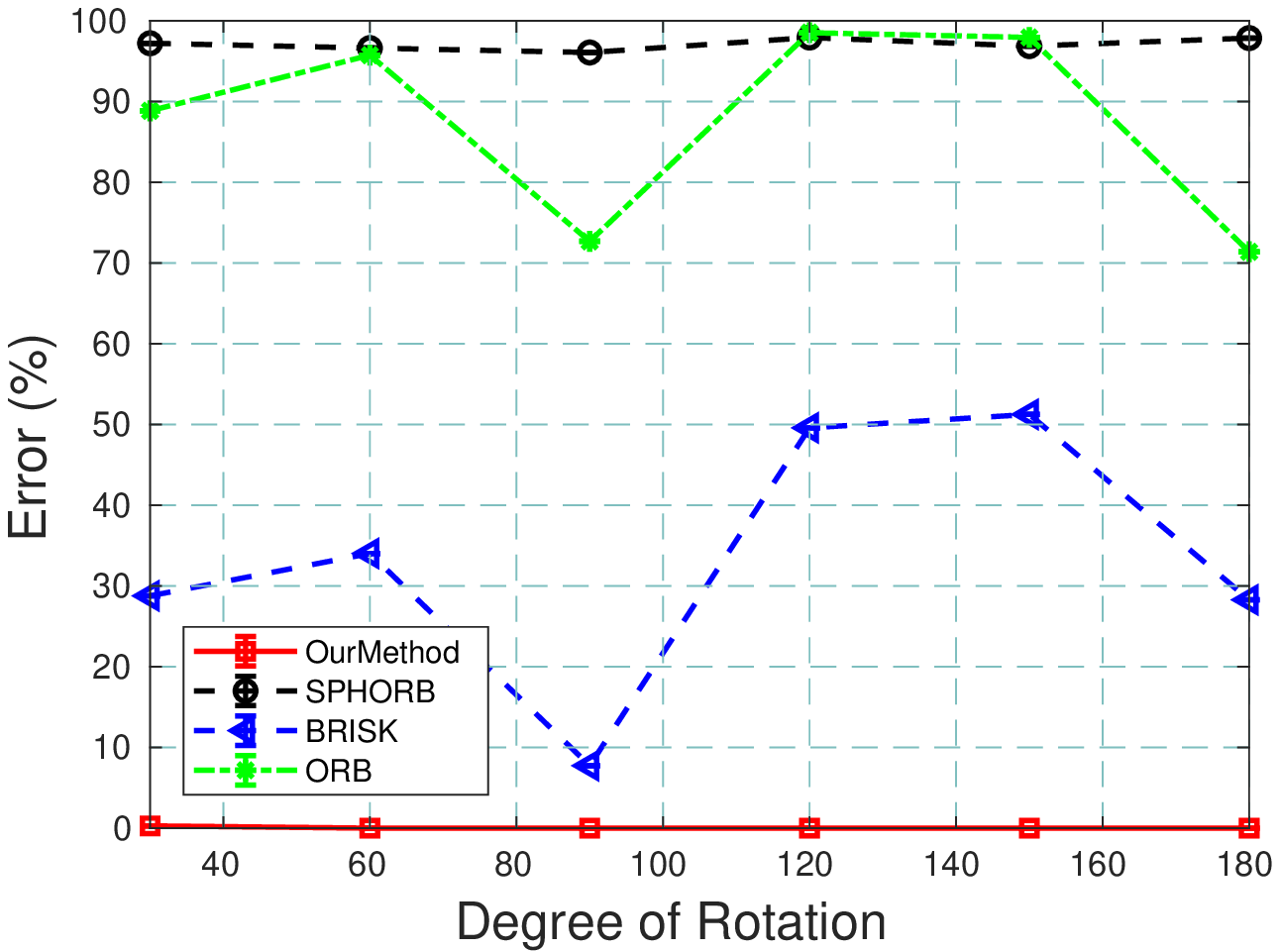}}%
    \hspace{-3mm}
    \qquad
    \subfigure[]{%
    \label{fig:affine2}%
    \includegraphics[width=35mm,height=30mm]{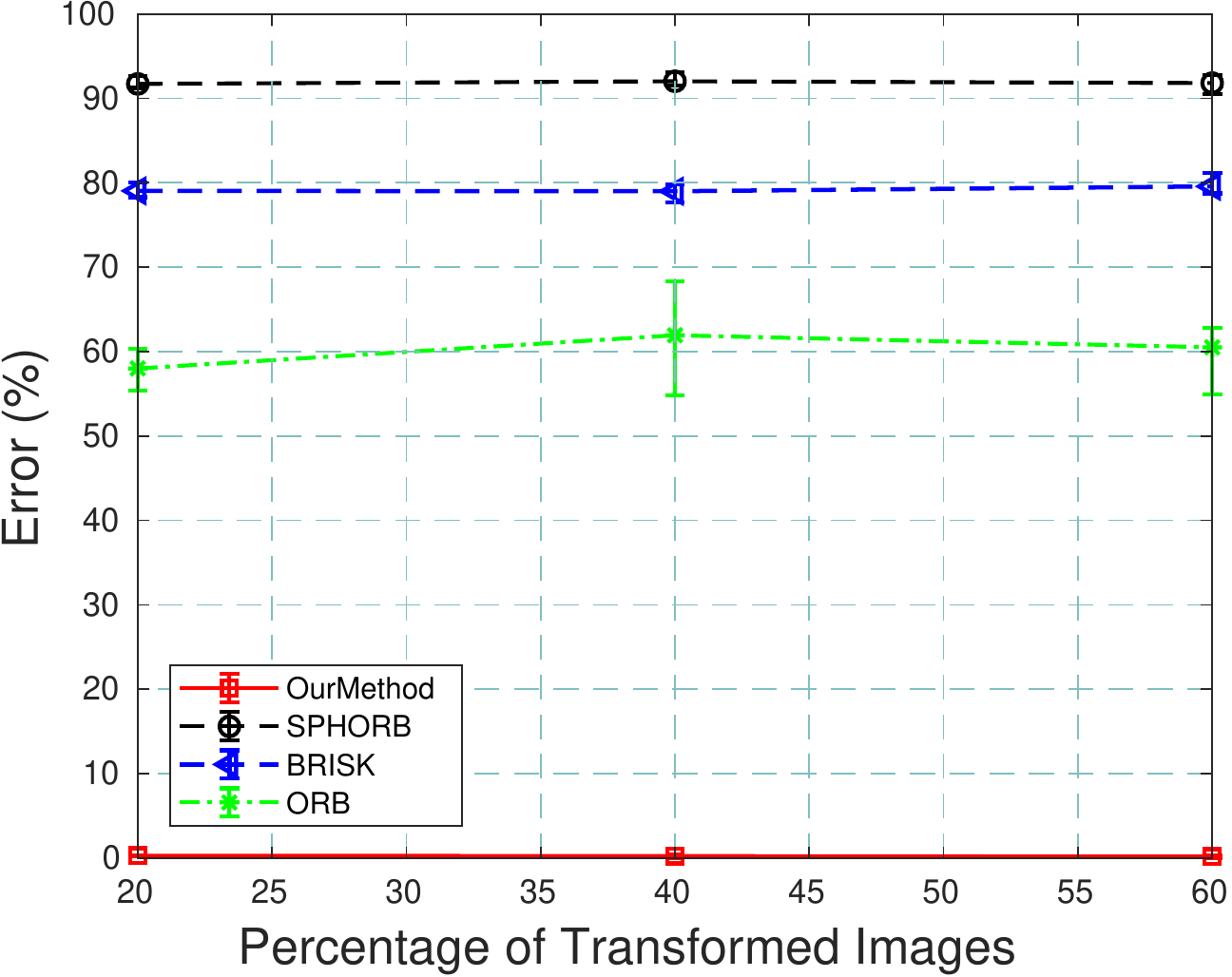}}%
    \hspace{-3mm}
        \qquad
    \subfigure[]{%
    \label{fig:affine3}%
    \includegraphics[width=35mm,height=30mm]{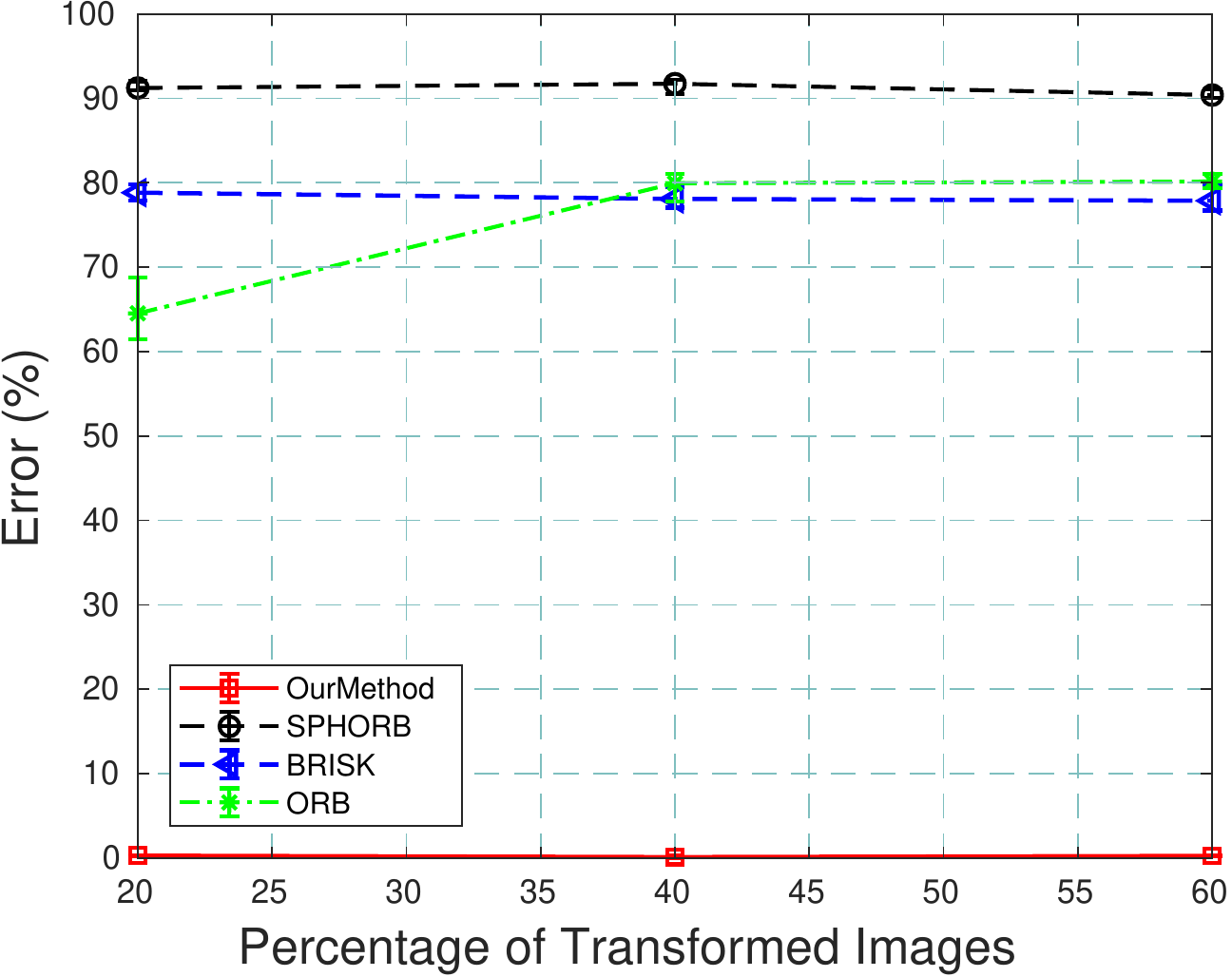}}%
    }
    \vspace{-5mm}
    \caption{Error(\%) of matching $(a)$ on SUN 360 dataset. We selected $10$ spherical images and created their rotated versions. Matching pairs of images for different angles from $30^{\circ}$ to $180^{\circ}$ rotation. $(b)-(c)$ when varying the percentage ($20\%$ to $60\%$) of transformed images in the set of spherical images of \emph{Desktop} for $20^\circ$ and $60^\circ$ rotation respectively.}
    \label{fig:Affine2}
\end{figure*}
\subsection{Effect of Rotation}

\begin{figure*}[h]
	%\centering
	\subfigure[]{%
		\label{fig:zero11}%
		\includegraphics[width=28mm,height=20mm]{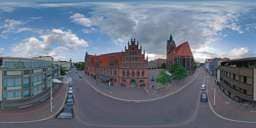}}%
	\qquad
	\hspace{-5mm}
	\subfigure[]{%
		\label{fig:first11}%
		\includegraphics[width=28mm,height=20mm]{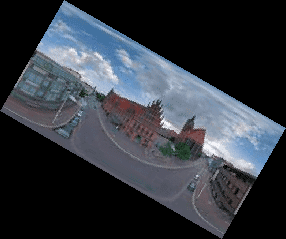}}%
	\qquad
	\hspace{-5mm}
	%\subfigure[]{%
	%	\label{fig:second11}%
	%	\includegraphics[width=27mm]{Experiments/Images/pano_7_90.png}}%
	%\qquad
	\subfigure[]{%
		\label{fig:third11}%
		\includegraphics[width=28mm,height=20mm]{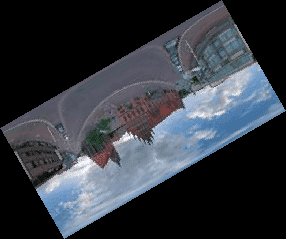}}%
	\qquad
        \hspace{-5mm}
	\subfigure[]{%
		\label{fig:fourth11}%
		\includegraphics[width=28mm,height=20mm]{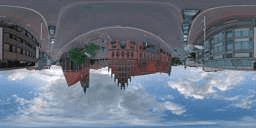}}%
	\vspace{-5mm}
	\caption{$(a)$ Original image from SUN360 dataset, rotated images after $(b)$ $30^{\circ}$ rotation, $(c)$ $150^{\circ}$ rotation, and $(d)$ $180^{\circ}$ rotation. Matching results are shown in Figure~\ref{fig:affine1} with the original image $(a)$.}
	\label{fig:Rotation}
\end{figure*}

We study the effect of rotation (affine transformation) on a simulated dataset. We collected $10$ different images from SUN360 dataset and annotated them. Therefore, the number of landmark 
points vary in all the images. The images are panoramic spherical images. We perform rotation on the images from $30^{\circ}$ to $180^{\circ}$ and compare the original image against the rotated 
images. Figure~\ref{fig:first11} to ~\ref{fig:fourth11} are the rotated versions of the original image in Figure~\ref{fig:zero11}. Matching results are presented in Figure~\ref{fig:affine1} against 
three feature descriptor methods. We note that, there is an increase in the error percentage for other methods as the degree of rotation increases from $30^{\circ}$ to $180^{\circ}$, except in
the $90^{\circ}$ rotation case, as $90^{\circ}$ rotation does not introduce distortions in the image and the pairwise-distances between points can be preserved. Whereas, our method is quite stable to 
transformation by rotation. We also rotated images (clockwise) by $20^\circ$ and $60^\circ$ for omnidirectional images of Desktop dataset and randomly transformed $20\%$ to $60\%$ of images 
in a set of spherical images. Results for both the rotations are shown in Figures~\ref{fig:affine2} and~\ref{fig:affine3}.
\begin{figure*}[h]
	\vspace{-5mm}
	\centering
	\makebox[\linewidth]{
	\subfigure[]{%
		\includegraphics[width=0.4\textwidth]{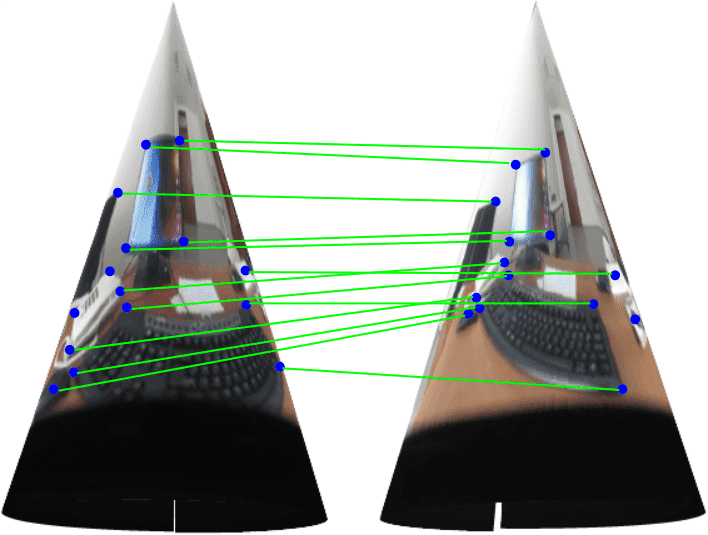}}%
	\qquad
	\subfigure[]{%
		%\hspace{0.6cm}
		\includegraphics[width=0.5\textwidth,height=3.0cm]{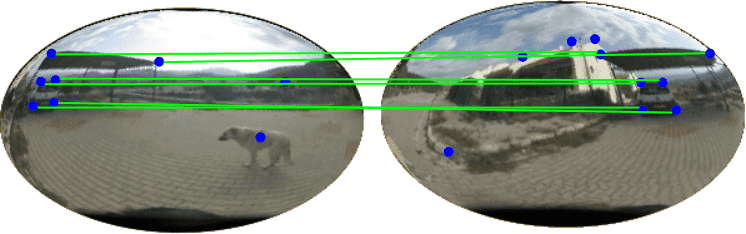}}%	
	}
	\vspace{-5mm}
	\caption{Matching on warped images of (left) Cone and (right) Ellipsoid. Green lines show correct matches and isolated points show no matches.}
	\label{fig:SOR}
\end{figure*}

\subsection{Effect of varying $k$-NN, Radius of surface and Number of points}
\begin{figure*}[h]
    \centering
    %\vspace{-5mm}
    \makebox[\linewidth]{
    \subfigure[]{%
    \label{fig:vary1}%
    \includegraphics[width=35mm,height=30mm]{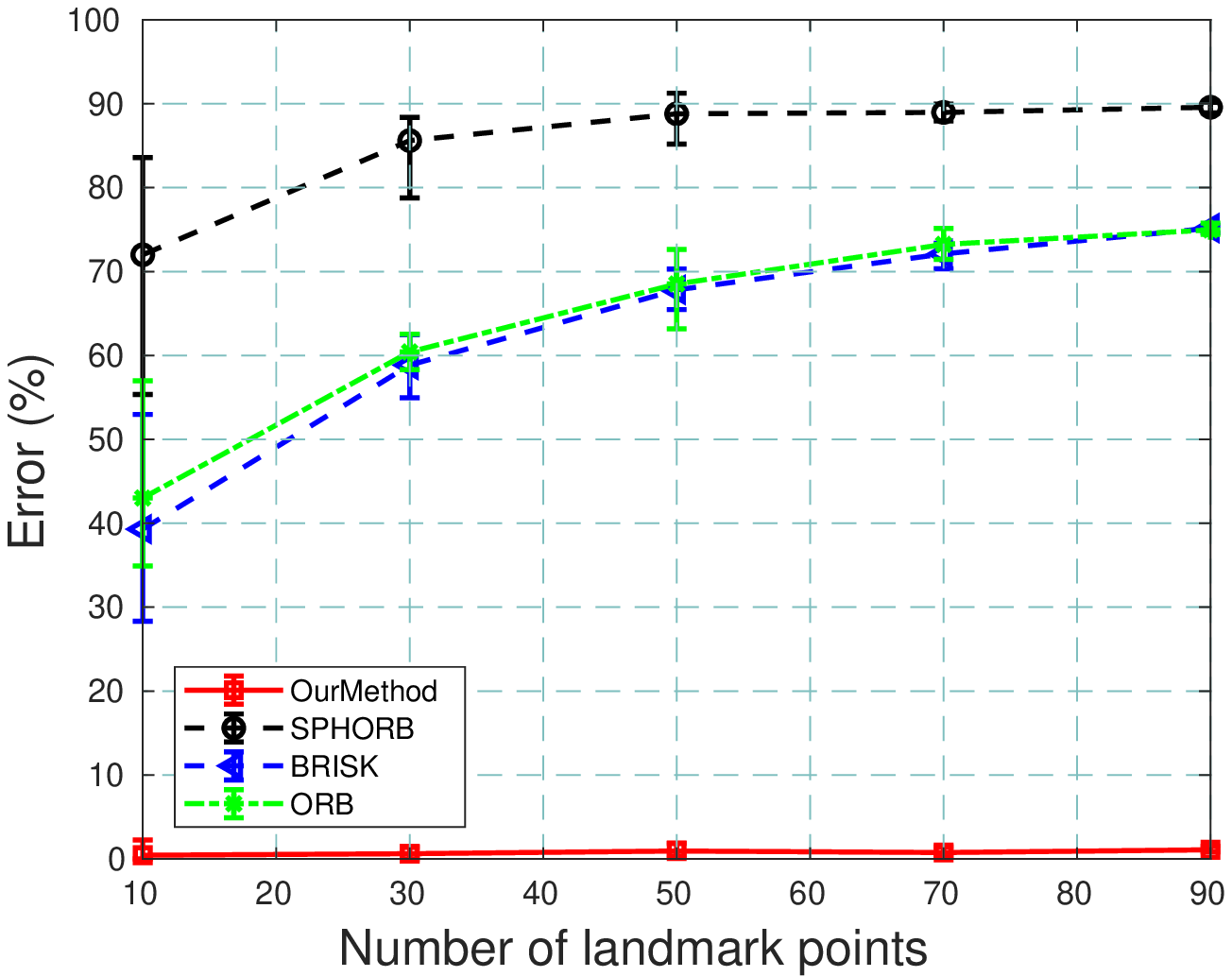}}%
    \hspace{-3mm}
    \qquad
    \subfigure[]{%
    \label{fig:vary2}%
    \includegraphics[width=35mm,height=30mm]{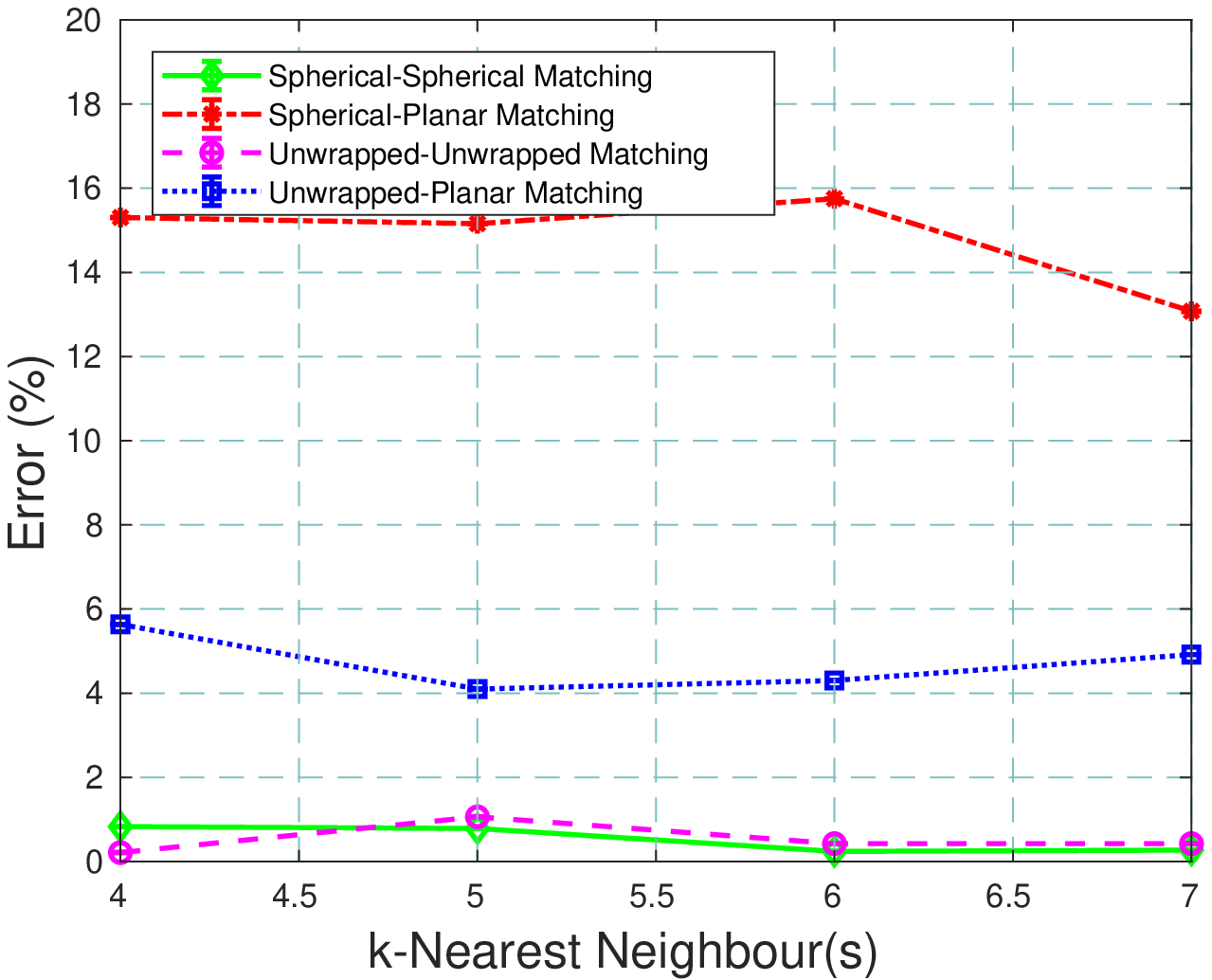}}%
    \hspace{-3mm}
        \qquad
    \subfigure[]{%
    \label{fig:vary3}%
    \includegraphics[width=35mm,height=30mm]{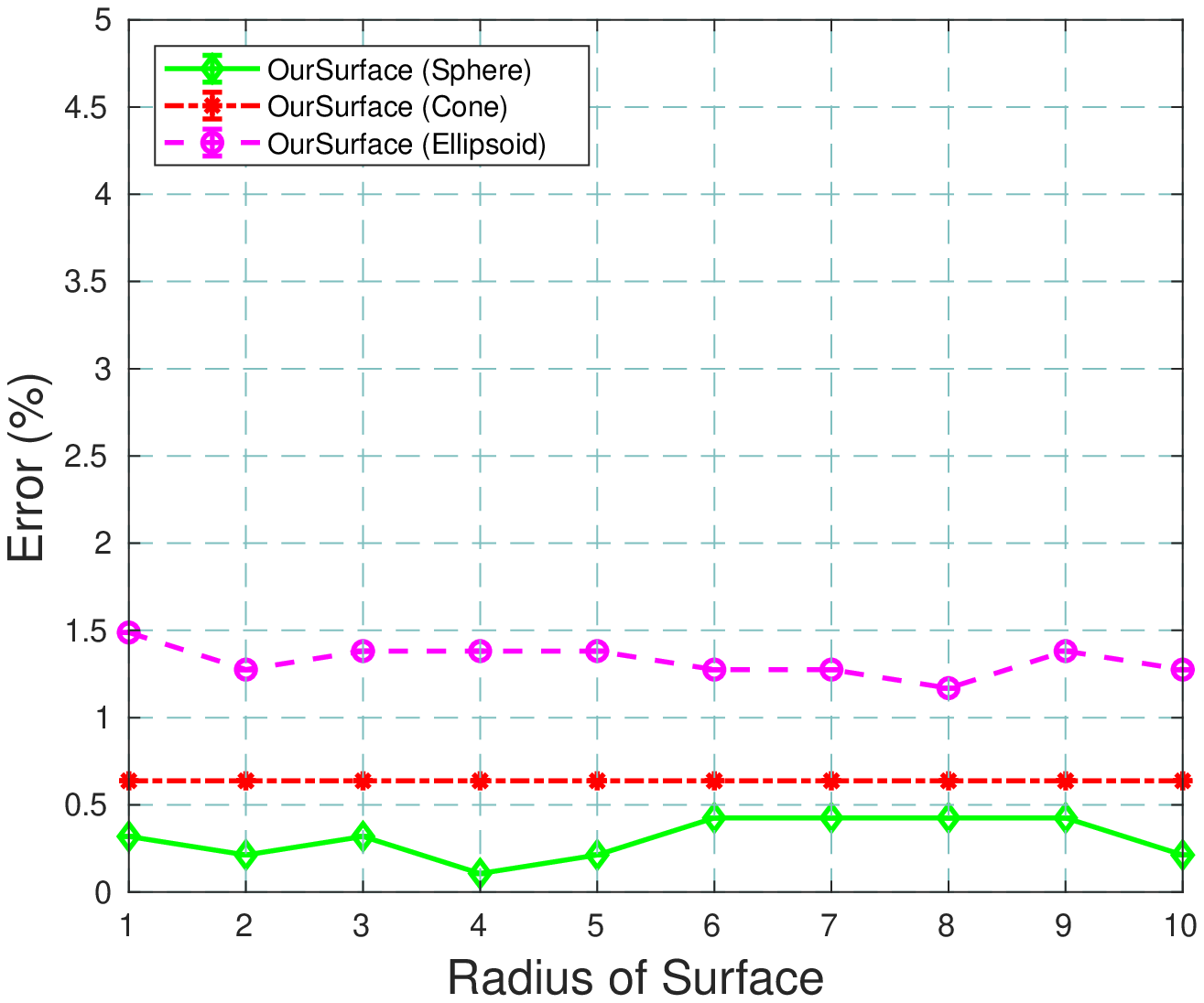}}%

    }
    \vspace{-5mm}
    \caption{Error(\%) of matching by $(a)$ varying the number of landmark points in the images of Table dataset(Sparse to Dense), $(b)$ varying the value of $k$ (from $k = 4$ to $7$) in $k$-NN ball of a landmark point, and $(c)$ varying the radius of surface (from $r = 1$ to $10$) of a sphere, cone and an ellipsoid for Desktop dataset.}
    \label{fig:VaryKRP}
\end{figure*}

We study the effect of parameters ($k$, $R$) and the effect of sparse / dense set of landmark points by performing few experiments for our method (spherical) and our warped on cone and ellipsoid methods 
and the existing feature descriptor based methods, respectively.\\ 
\textbf{Vary $k$-Nearest Neighbour:} We performed few experiments of matching between spherical vs. spherical (for Kamaishi dataset), spherical vs. planar (Desktop), unwrapped spherical vs. unwrapped 
spherical (Desktop) and unwrapped spherical vs. planar (Parking) images by varying $k$ parameter of $k$-NN while generating a simplicial complex. 
Figure~\ref{fig:vary2} shows the results for four different datasets with $k$ varying from range $4$ to $7$ on $x$-axis. Although, the value of $k$ changes the 
matching results, the error doesn't change significantly. 
Selection of $k$ is important since it defines the local connectivity of a 
vertex. So, larger the value of $k$, larger are the adjacent vertices of a vertex in question. However, a large value of $k$ can slow down the algorithm and hence this trade-off must be taken into consideration while choosing $k$.
\\
\textbf{Vary Radius of Surface:} Projecting an image onto a sphere or manifold depends on the radius of the underlying manifold. Change in radius affects the rate of distortion of projection onto the manifold.
Thus, we conducted an experiment on spherical (omnidirectional) images of Desktop dataset. Results for varying $R$ ranging from $1$ to $10$ are shown in Figure~\ref{fig:vary3}. Matching is performed on our three 
methods including spherical (radius of $2$-sphere), cone (radius of cone) and ellipsoid (radius of $2$-sphere projected to an ellipse with a fixed major and minor axis). Results report that the change 
in radius doesn't affect the matching accuracy. \\
\textbf{Sparse to Dense set of landmark points:} We considered spherical (omnidirectional and fish-eye) images of Table dataset which consists of $93$ landmark points originally. We randomly selected $10$ to $90$ landmark points 
on $x$-axis in Figure~\ref{fig:vary1}. This experiment analyzes if the matching algorithm depends on sparsity / density of landmark points in an image. From the results, our method doesn't seem to be 
affected by \emph{removal} or \emph{addition} of points whereas there is a substantial increase in error for other methods.
\\

\subsection{Effect of Missing Points}
\begin{figure*}[t]
    \centering
    %\vspace{-5mm}
    \makebox[\linewidth]{
    \subfigure[]{%
    \label{fig:trans1}%
    \includegraphics[width=35mm,height=30mm]{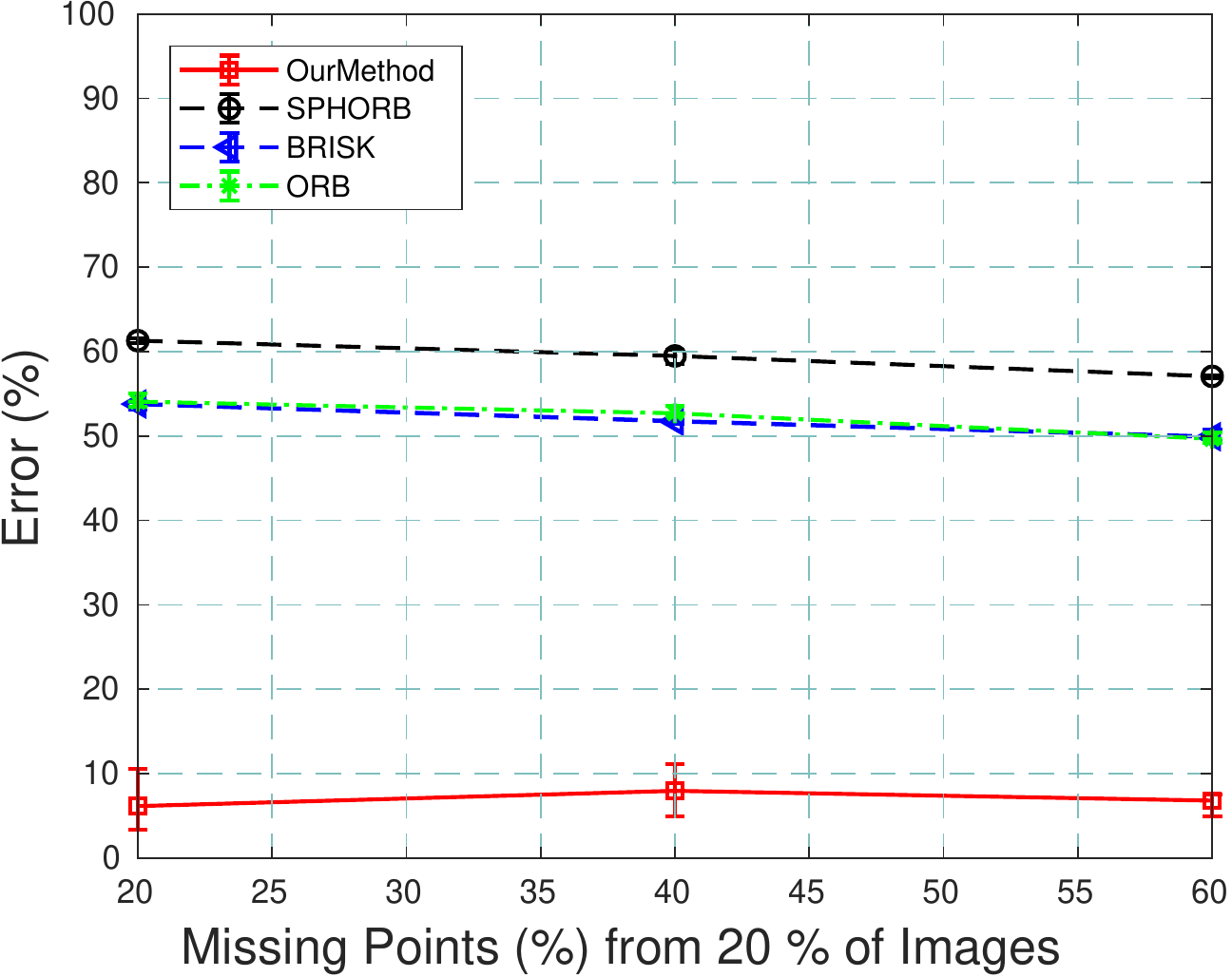}}%

    \qquad
    \subfigure[]{%
    \label{fig:trans2}%
    \includegraphics[width=35mm,height=30mm]{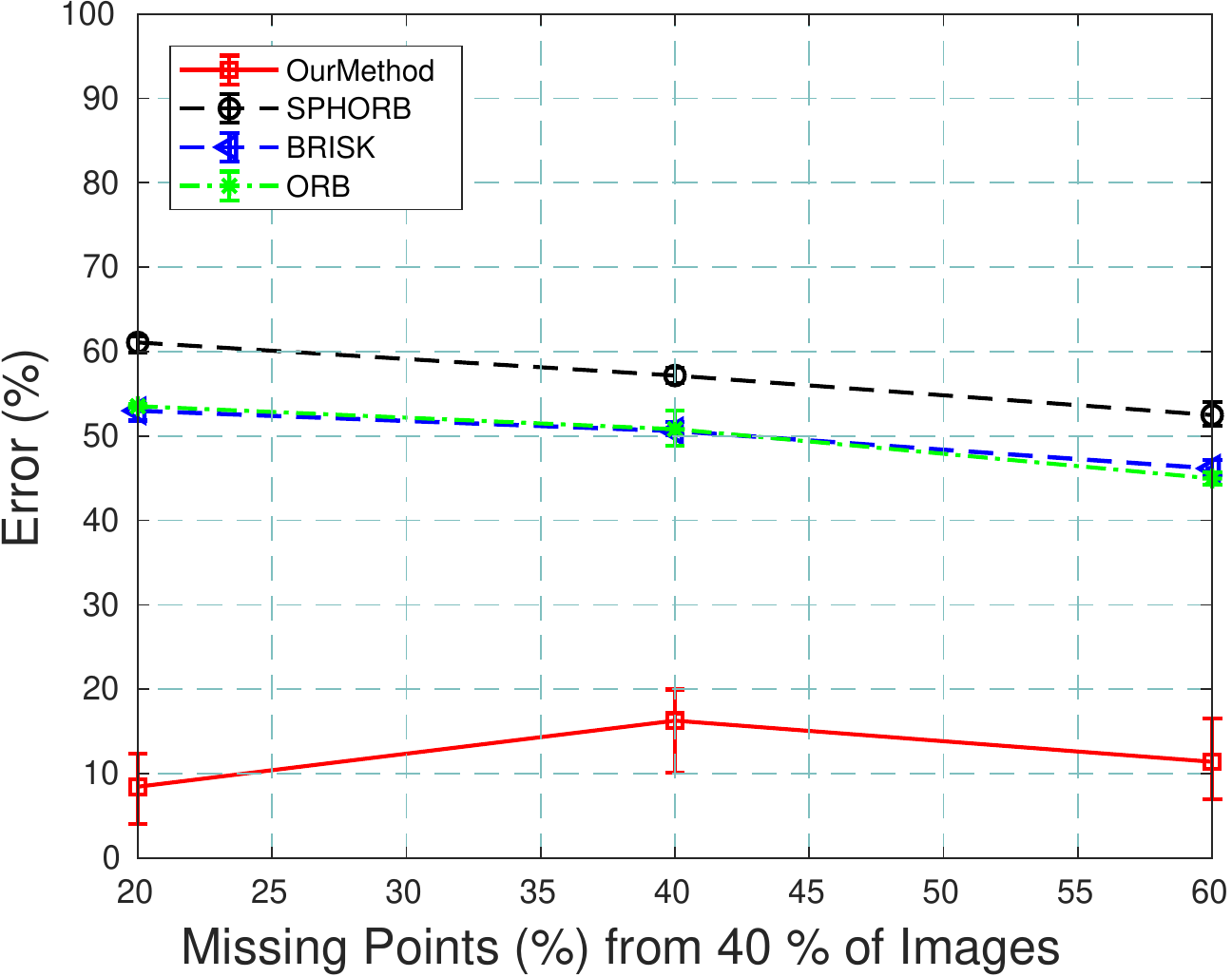}}%
        \qquad
    \subfigure[]{%
    \label{fig:trans3}%
    \includegraphics[width=35mm,height=30mm]{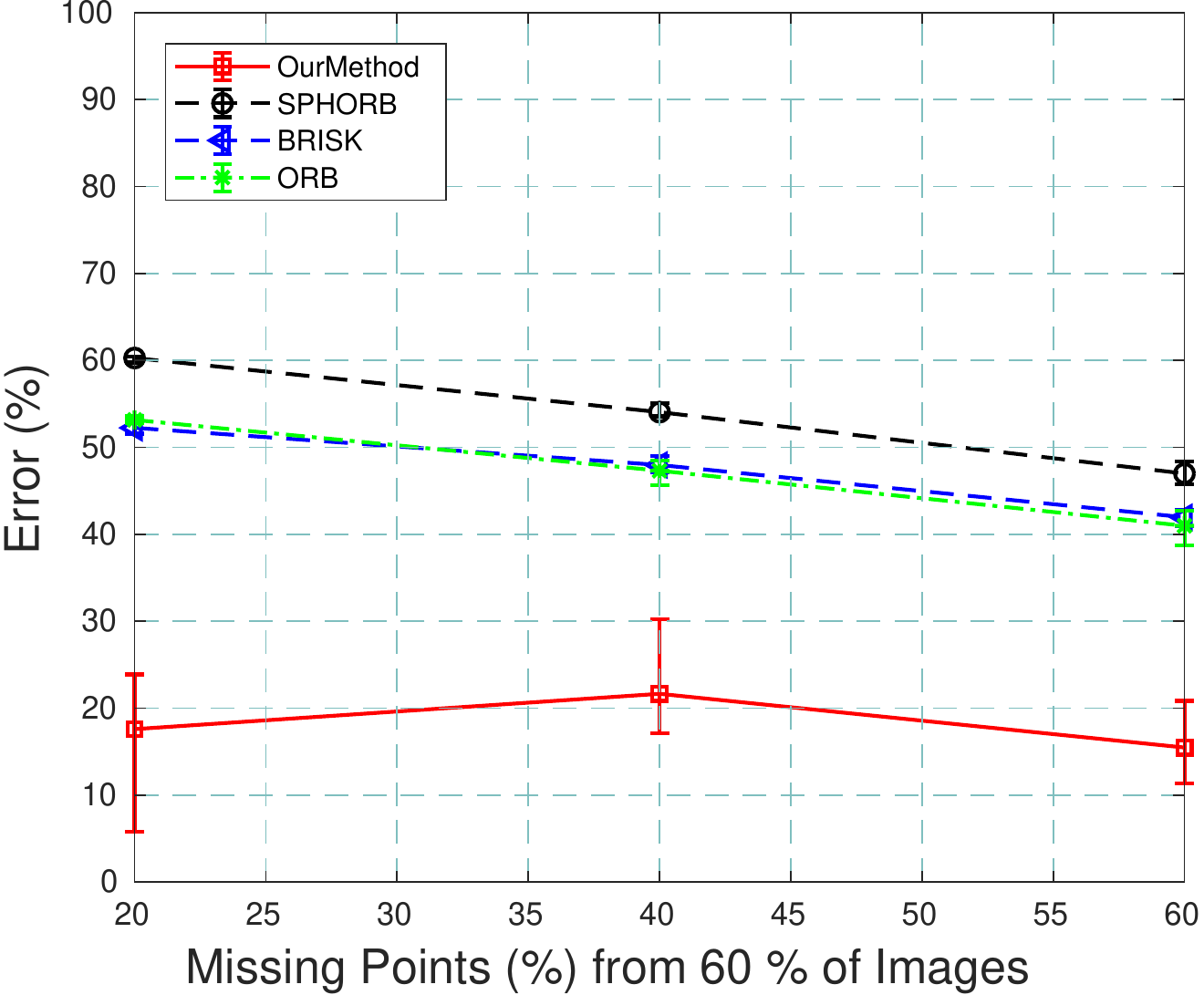}}%
    }
    \vspace{-2mm}
    \makebox[\linewidth]{
    \subfigure[]{%
    \label{fig:trans4}%
    \includegraphics[width=35mm,height=30mm]{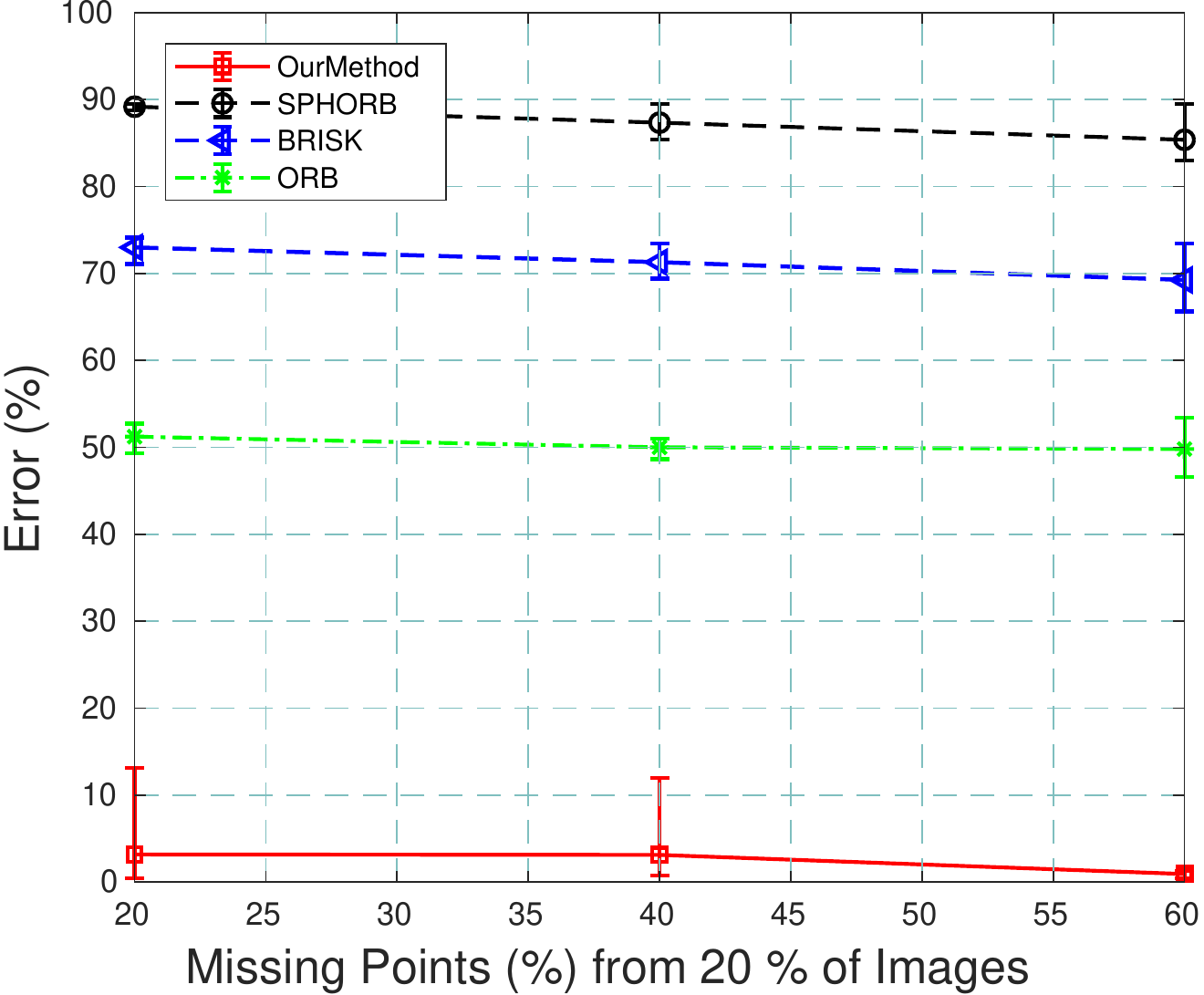}}%

        \qquad
    \subfigure[]{%
    \label{fig:trans5}%
    \includegraphics[width=35mm,height=30mm]{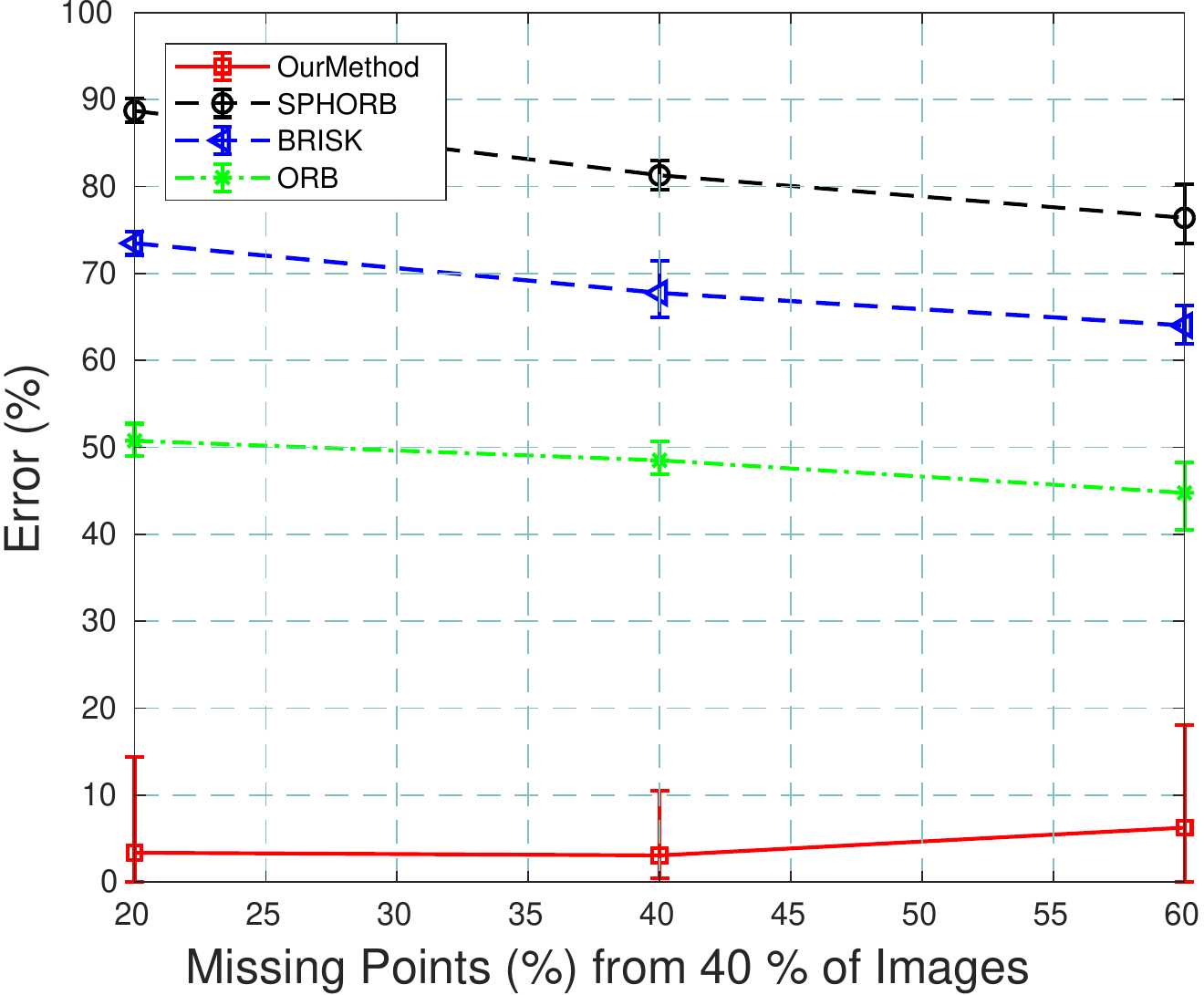}}%
            \qquad
    \subfigure[]{%
    \label{fig:trans6}%
    \includegraphics[width=35mm,height=30mm]{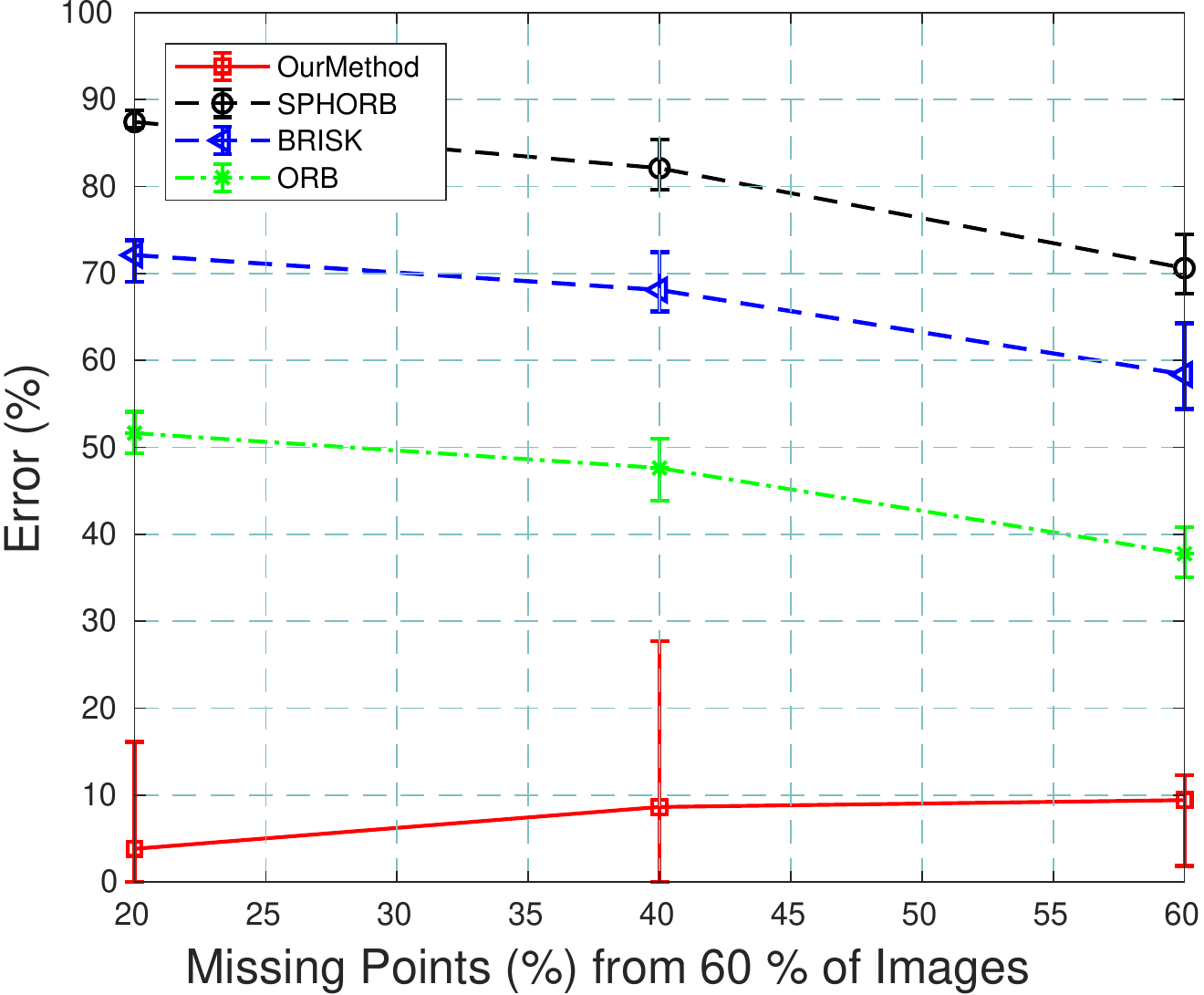}}%
    }
        \vspace{-2mm}
    \makebox[\linewidth]{
    \subfigure[]{%
    \label{fig:trans7}%
    \includegraphics[width=35mm,height=30mm]{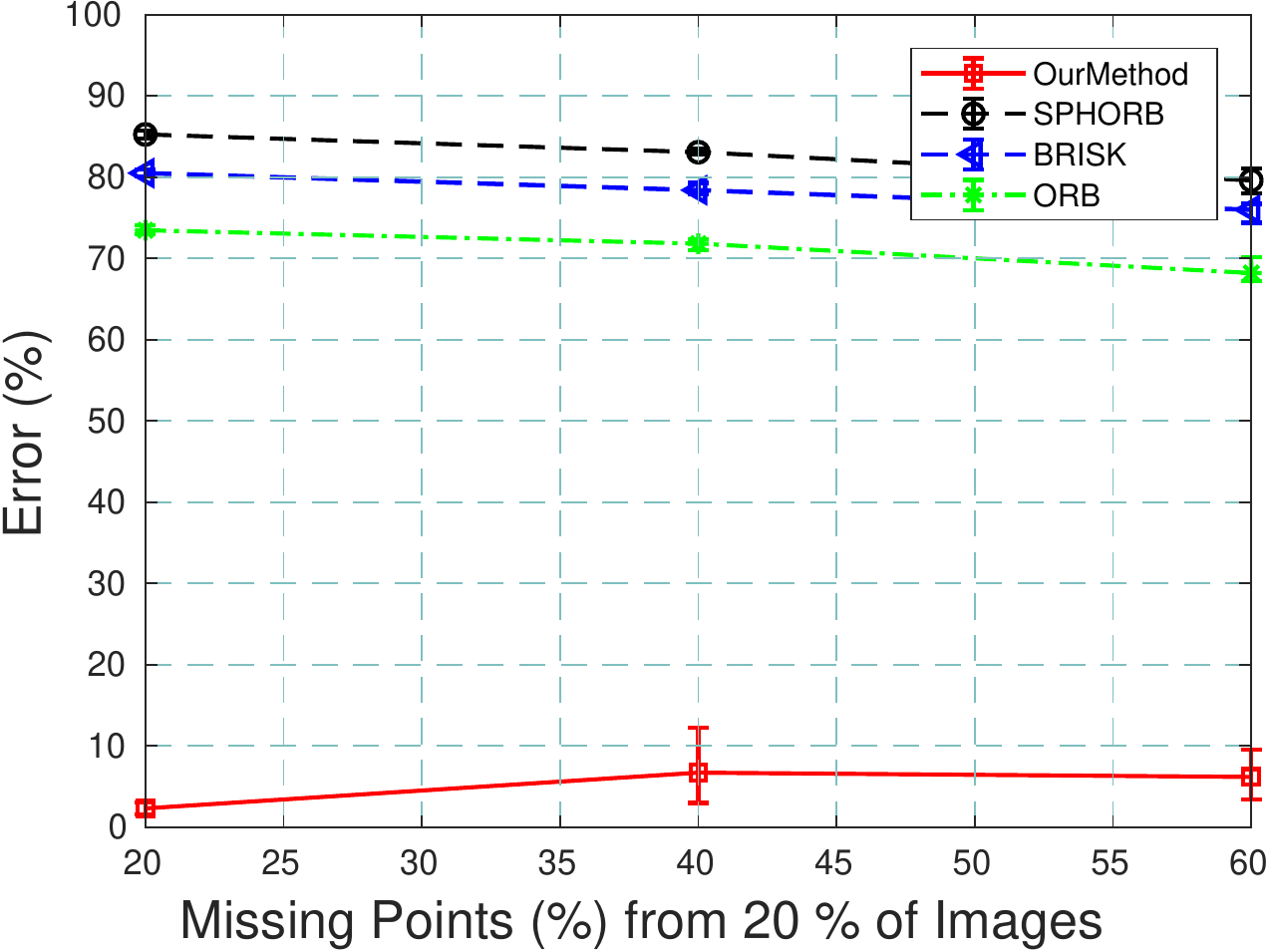}}%
        \qquad
    \subfigure[]{%
    \label{fig:trans8}%
    \includegraphics[width=35mm,height=30mm]{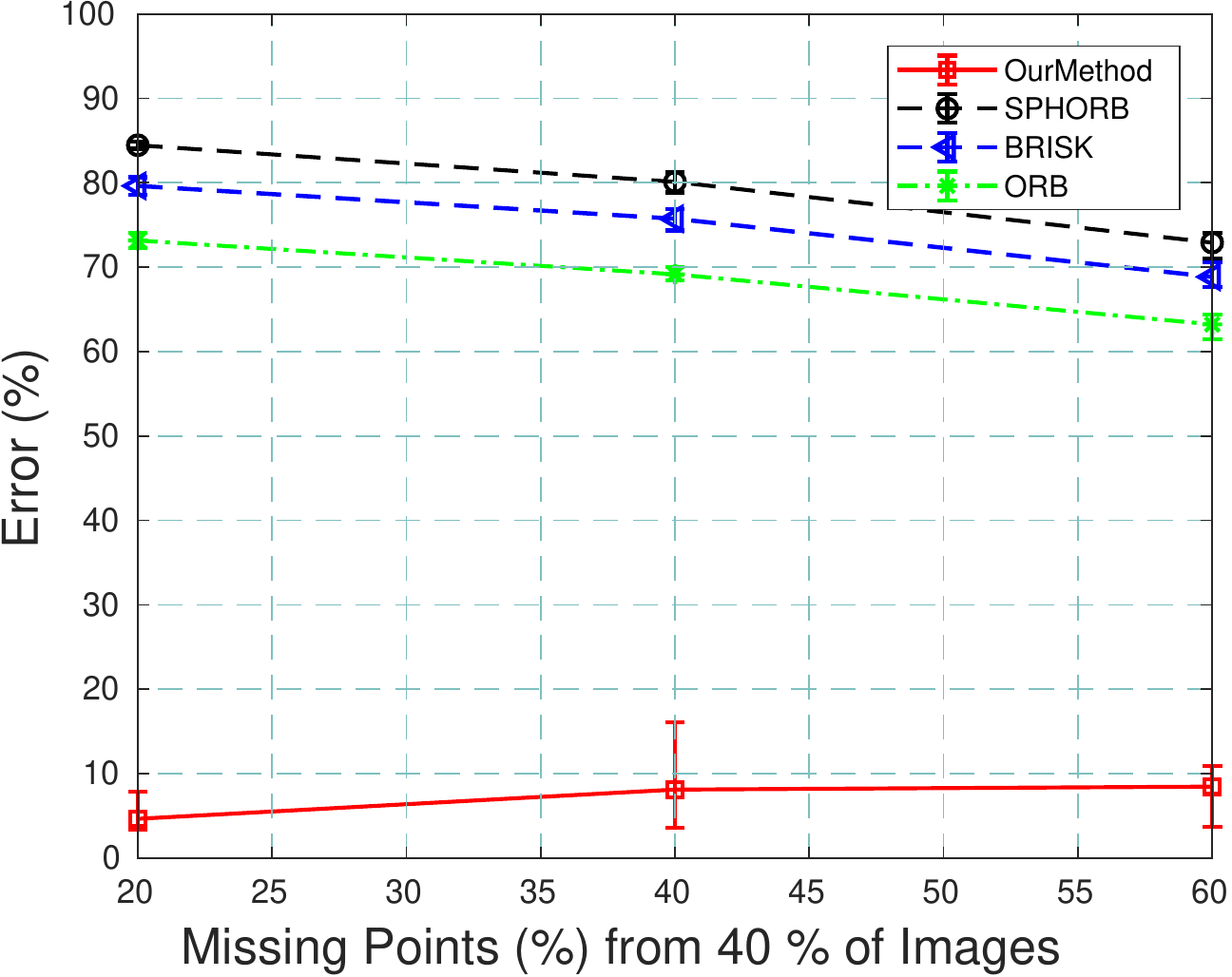}}%
            \qquad
    \subfigure[]{%
    \label{fig:trans8}%
    \includegraphics[width=35mm,height=30mm]{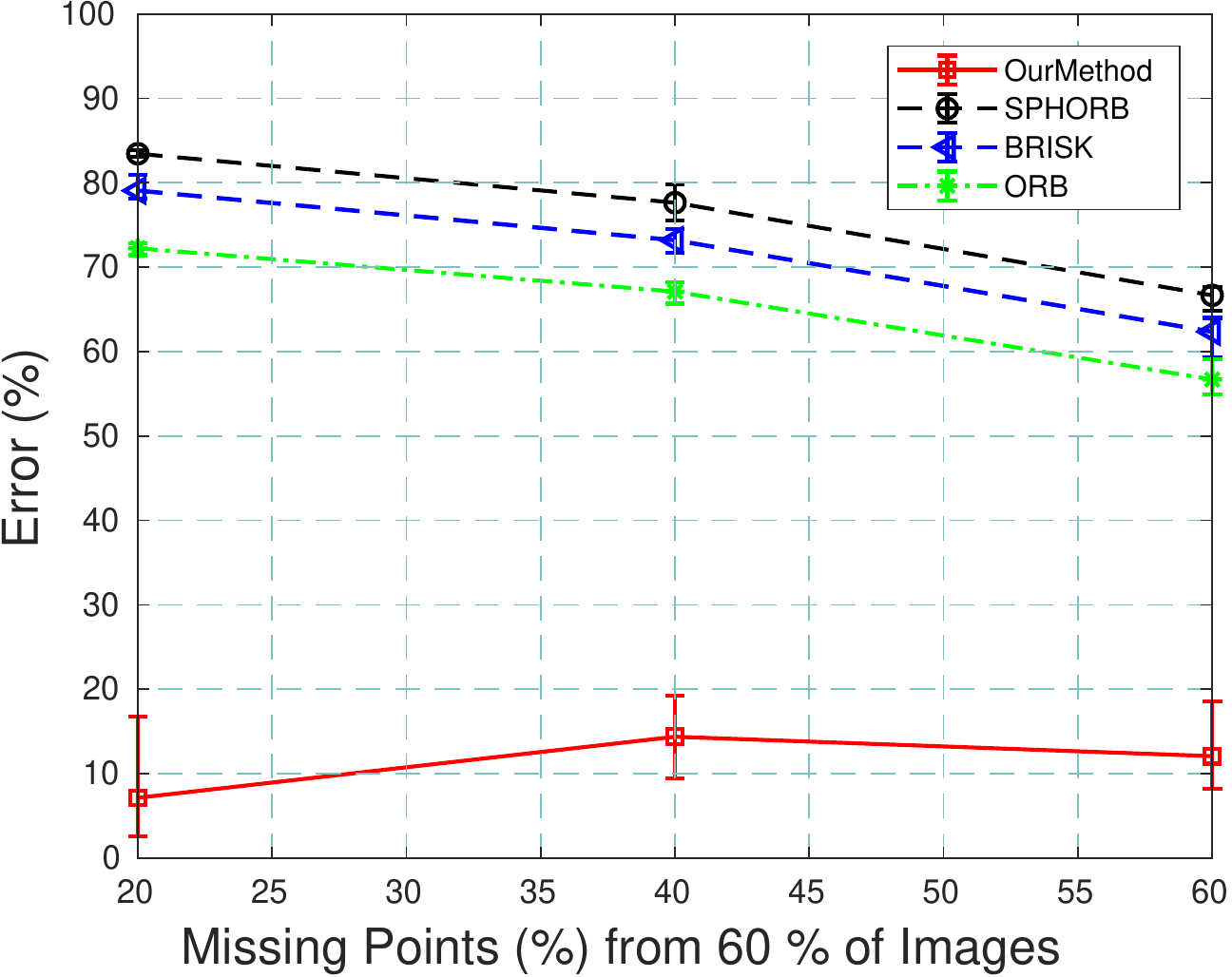}}%
    }
    \vspace{-2mm}
    \makebox[\linewidth]{
    \subfigure[]{%
    \label{fig:trans7}%
    \includegraphics[width=35mm,height=30mm]{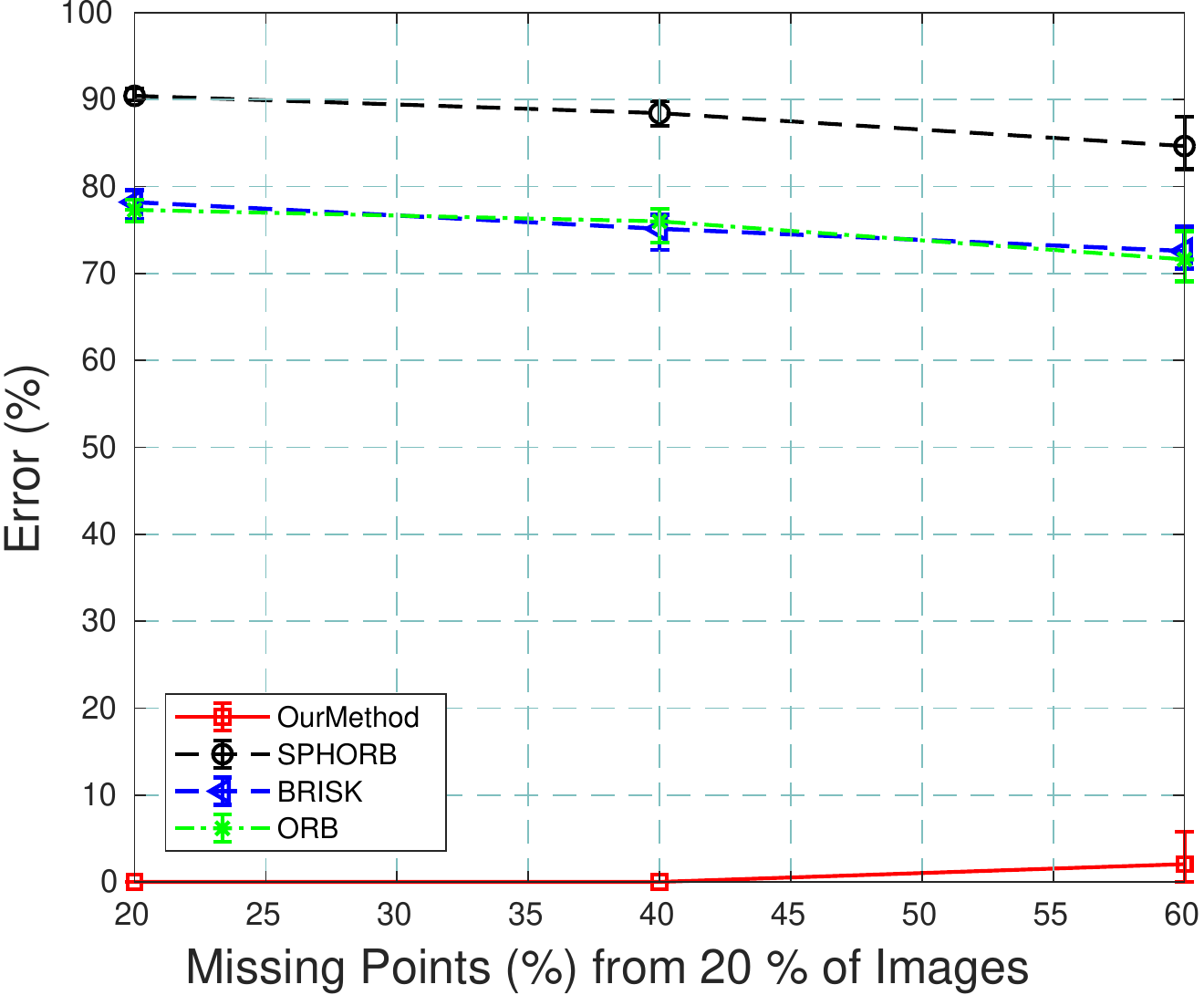}}%
        \qquad
    \subfigure[]{%
    \label{fig:trans8}%
    \includegraphics[width=35mm,height=30mm]{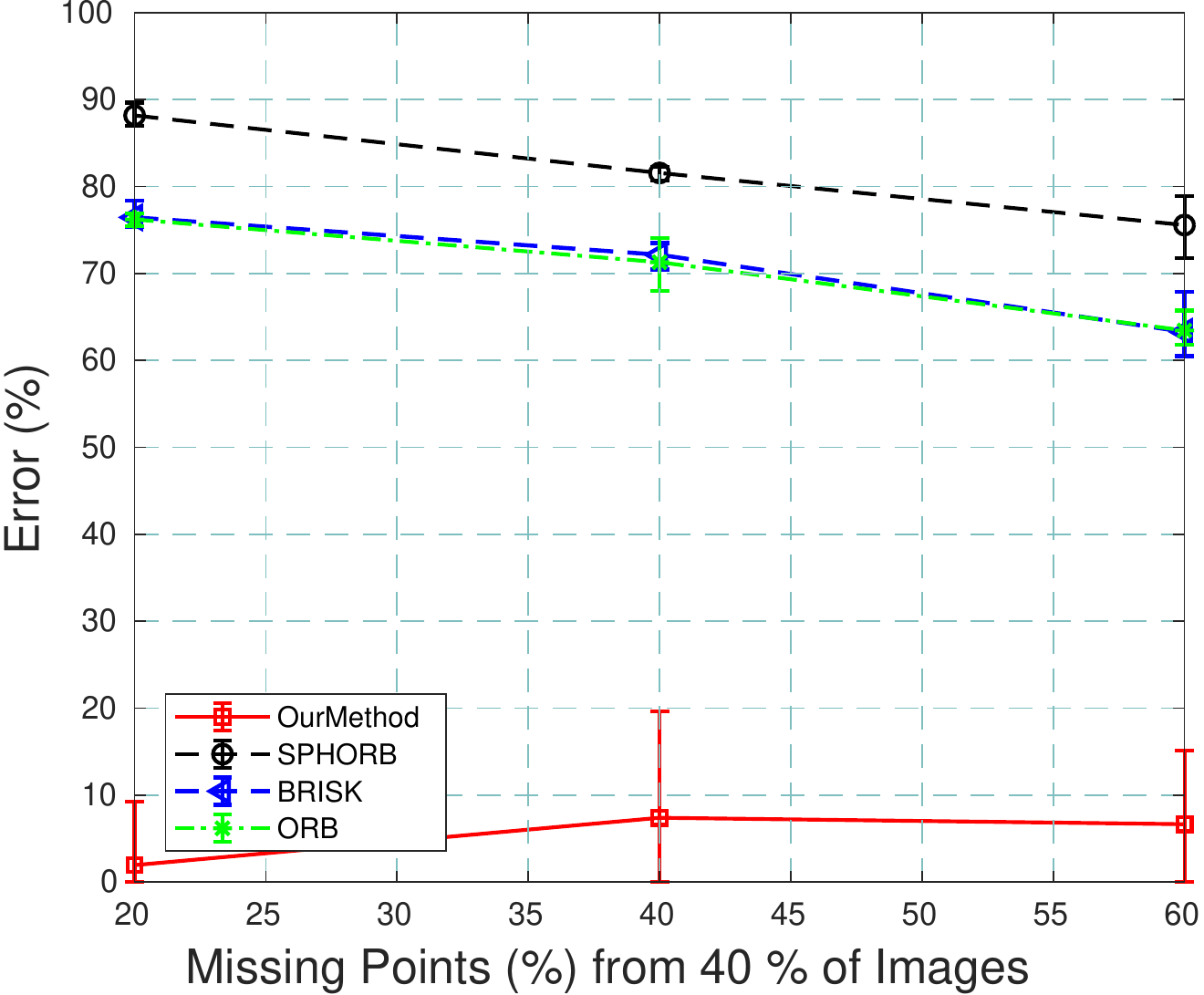}}%
            \qquad
    \subfigure[]{%
    \label{fig:trans8}%
    \includegraphics[width=35mm,height=30mm]{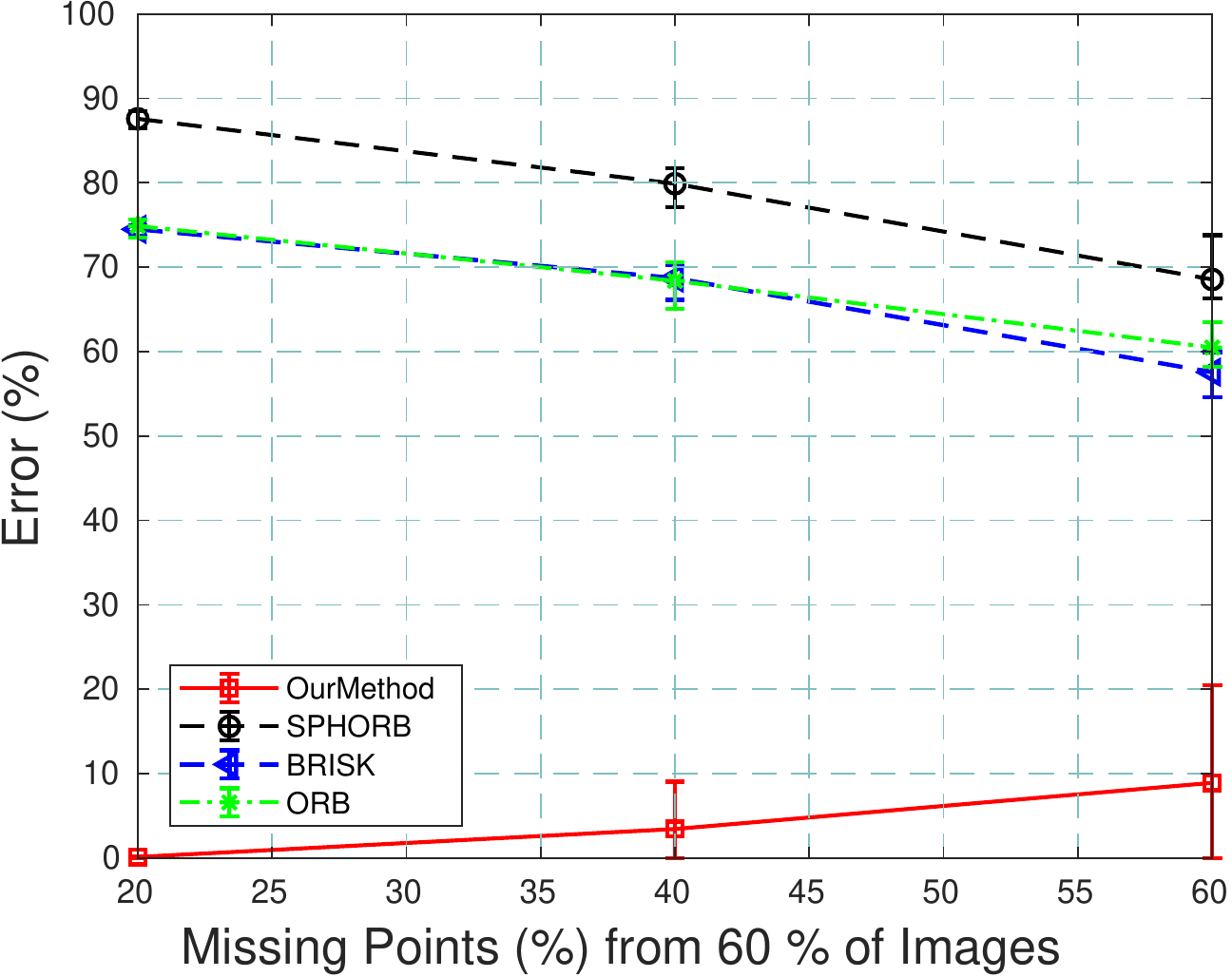}}%
    }
    %\vspace{-5mm}
    \caption{Error (\%) of matching between spherical images by randomly removing points ($20\%-60\%$) from $20\%$, $40\%$ and $60\%$ of the images. Matching is computed for four datasets ($(a)$-$(c)$ Chessboard , $(d)$-$(f)$ Desktop, $(g)$-$(i)$ Kamaishi, $(j)$-$(l)$ Table) from $1^{st}$ frame to the other $N-1$ frames.}
    \label{fig:Missing}
\end{figure*}

In order to compare the robustness of our method versus the state-of-the-art methods under the effect of landmark points \emph{missing completely at random (MCAR)}, 
we randomly remove $20\%$ to $60\%$ of the landmark points from $20\%$ to $60\%$ of images in a set of spherical images. 
We picked an arbitrary image as the original image and match it against the rest of the spherical images, affected by missing points, in the set. We conducted this 
experiment on four datasets mentioned in Figure~\ref{fig:Missing}. We have shown results for removal of points from $20\%$, $40\%$ and $60\%$ of images in Figure~\ref{fig:Missing}. 
We observe that there is no large increase in error when $20\%$ to $60\%$ of the landmark points are removed randomly for our method. Also, results for other methods 
are not affected by varying percentage of missing points and there is no increase in the error percentage since removal of points reduces the number of points in matching, which in turn also reduces error 
for feature descriptor methods because of decrease in probability to mismatch.
%\newpage
\begin{figure*}
	\makebox[\linewidth]{
	\centering
	\vspace{-10mm}
	  \subfigure[]{%
	  \label{fig:SOR1}%
	    \includegraphics[trim={0 0 0 0},clip,width=55mm,height=45mm]{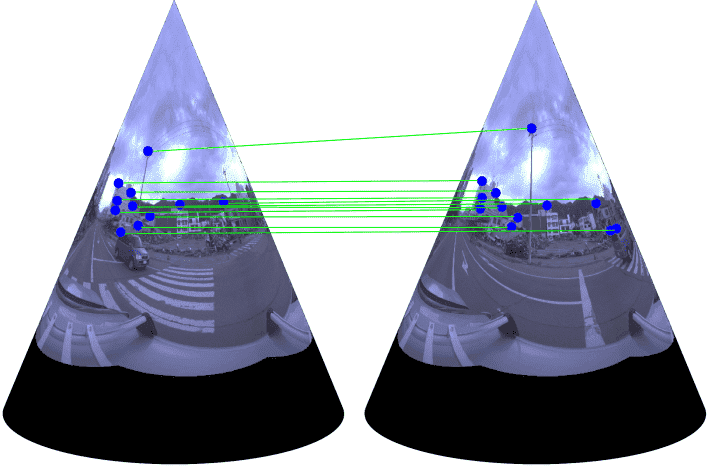}}%
	  %\caption{a) Matchings for house frame $1$ (left) with frame $55$ (right), b) Matchings for house frame $1$ (left) with frame $110$ (right), Error: $0.0\%$.}
	  \qquad
	  %\hspace{0.5mm}
	  \subfigure[]{%
	  \label{fig:SOR2}%
	    \includegraphics[trim={0 0 0 0},clip,width=65mm,height=28mm]{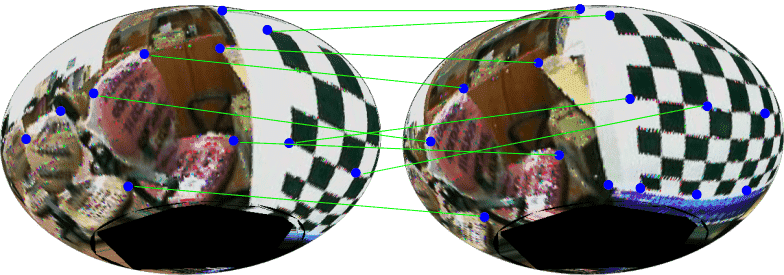}}%
	    %\vspace{-5cm}
	  }
	  \vspace{-1mm}
	  \qquad
	  \makebox[\linewidth]{
	  \centering
	  \hspace{-17mm}
	  \subfigure[]{%
	  \label{fig:SOR3}%
	    \includegraphics[trim={0 0 0 0},clip,width=55mm,height=45mm]{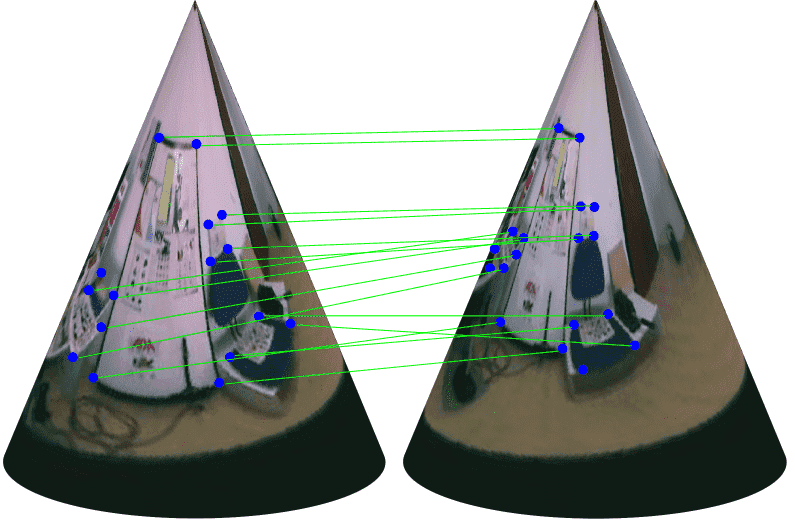}}%
	  \qquad
	  \subfigure[]{%
	  \label{fig:SOR4}%
	    \includegraphics[trim={0 0 0 0},clip,width=65mm,height=28mm]{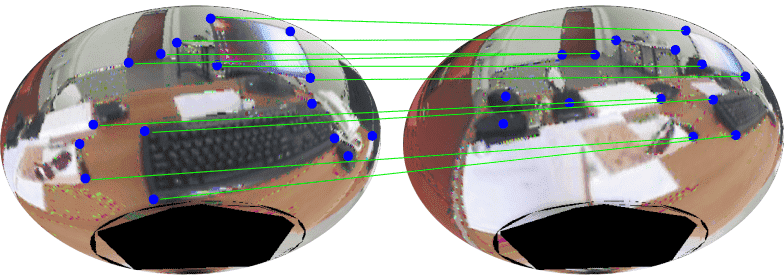}}%
	  }
	  	  \vspace{-5mm}
	  \qquad
	  \makebox[\linewidth]{
	  \centering
	  \hspace{-17mm}
	  \subfigure[]{%
	  \label{fig:SOR3}%
	    \includegraphics[trim={0 0 0 0},clip,width=55mm,height=45mm]{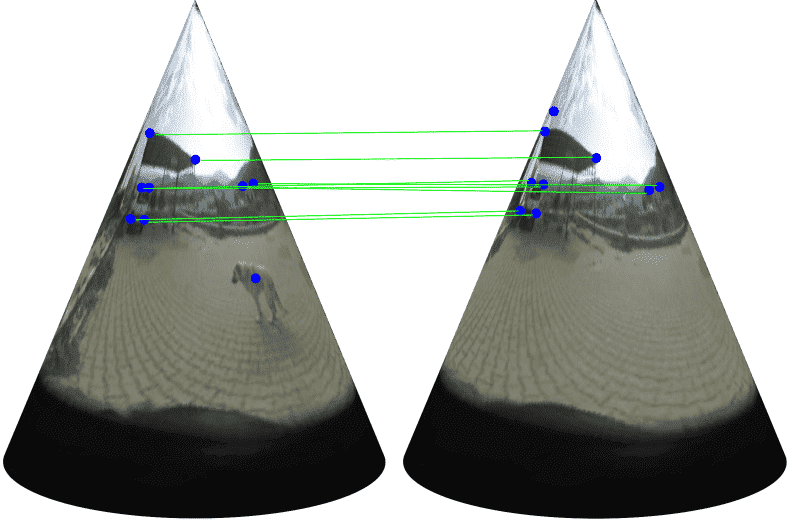}}%
	  \qquad
	  \subfigure[]{%
	  \label{fig:SOR4}%
	    \includegraphics[trim={0 0 0 0},clip,width=65mm,height=28mm]{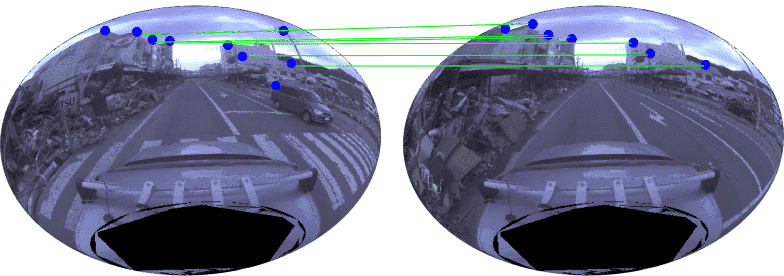}}%
	  }
	  \vspace{-1mm}
	  \caption{Instances of a matching of unwrapped images of different datasets on curved manifold for $(a)$ Kamaishi, $(b)$ Chessboard, $(c)$ Table, $(d)$ Desktop, $(e)$ Parking and $(f)$ Kamaishi. Green lines show correct matches respectively. Isolated points show no matches.}
	  \label{fig:SoR}
\end{figure*}

\begin{figure*}
	\makebox[\linewidth]{
	\centering
	\vspace{-10mm}
	  \subfigure[]{%
	  \label{fig:KamaishiMatch}%
	    \includegraphics[trim={0 0 0 0},clip,width=55mm,height=28mm]{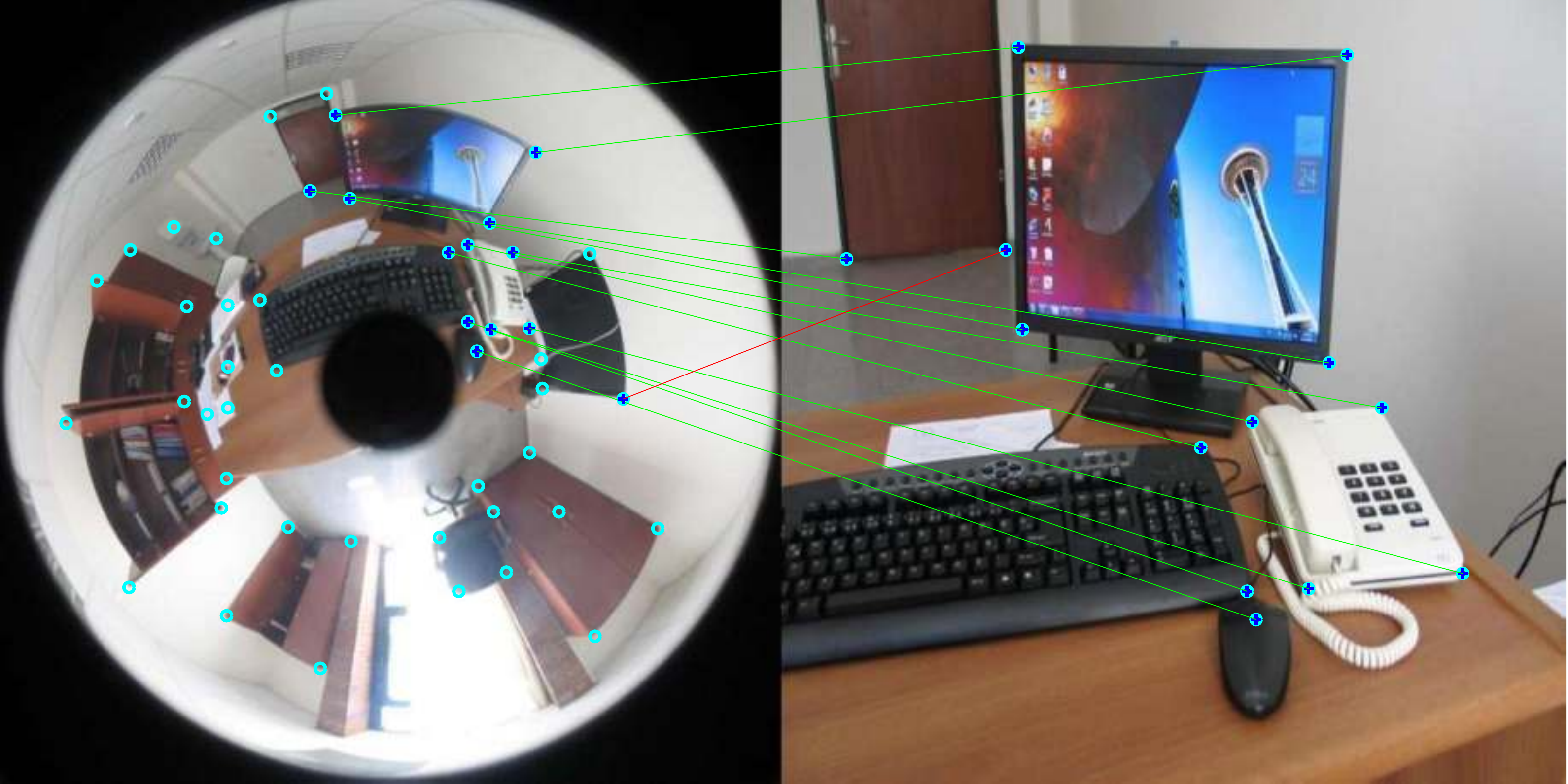}}%
	  %\caption{a) Matchings for house frame $1$ (left) with frame $55$ (right), b) Matchings for house frame $1$ (left) with frame $110$ (right), Error: $0.0\%$.}
	  \qquad
	  %\hspace{0.5mm}
	  \subfigure[]{%
	  \label{fig:KamaishiMatch2}%
	    \includegraphics[trim={0 0 0 0},clip,width=55mm,height=28mm]{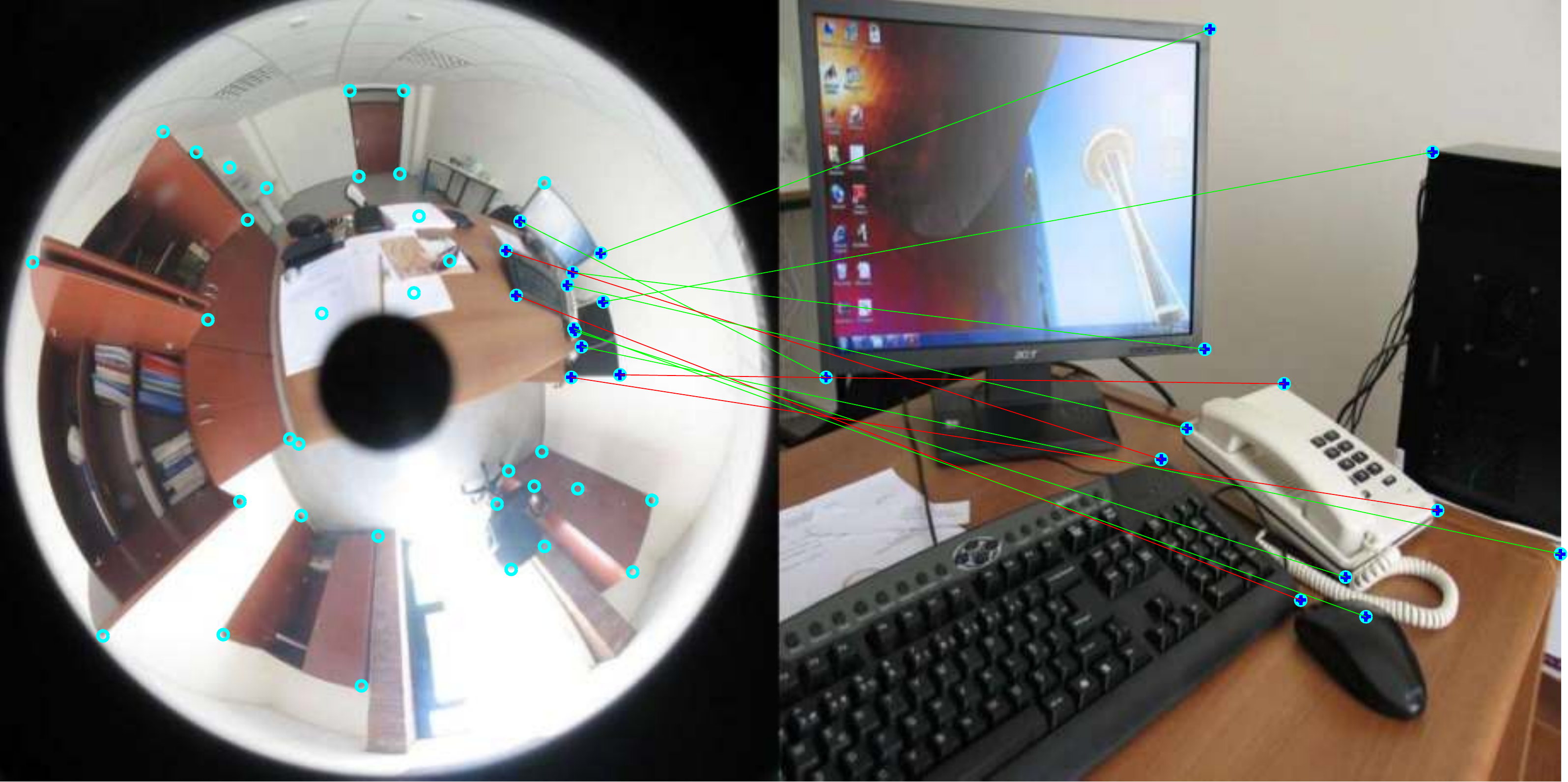}}%
	    %\vspace{-5cm}
	  }
	  \vspace{-2mm}
	  \qquad
	  \makebox[\linewidth]{
	  \centering
	  \hspace{-17mm}
	  \subfigure[]{%
	  \label{fig:HorseRotate}%
	    \includegraphics[trim={0 0 0 0},clip,width=55mm,height=28mm]{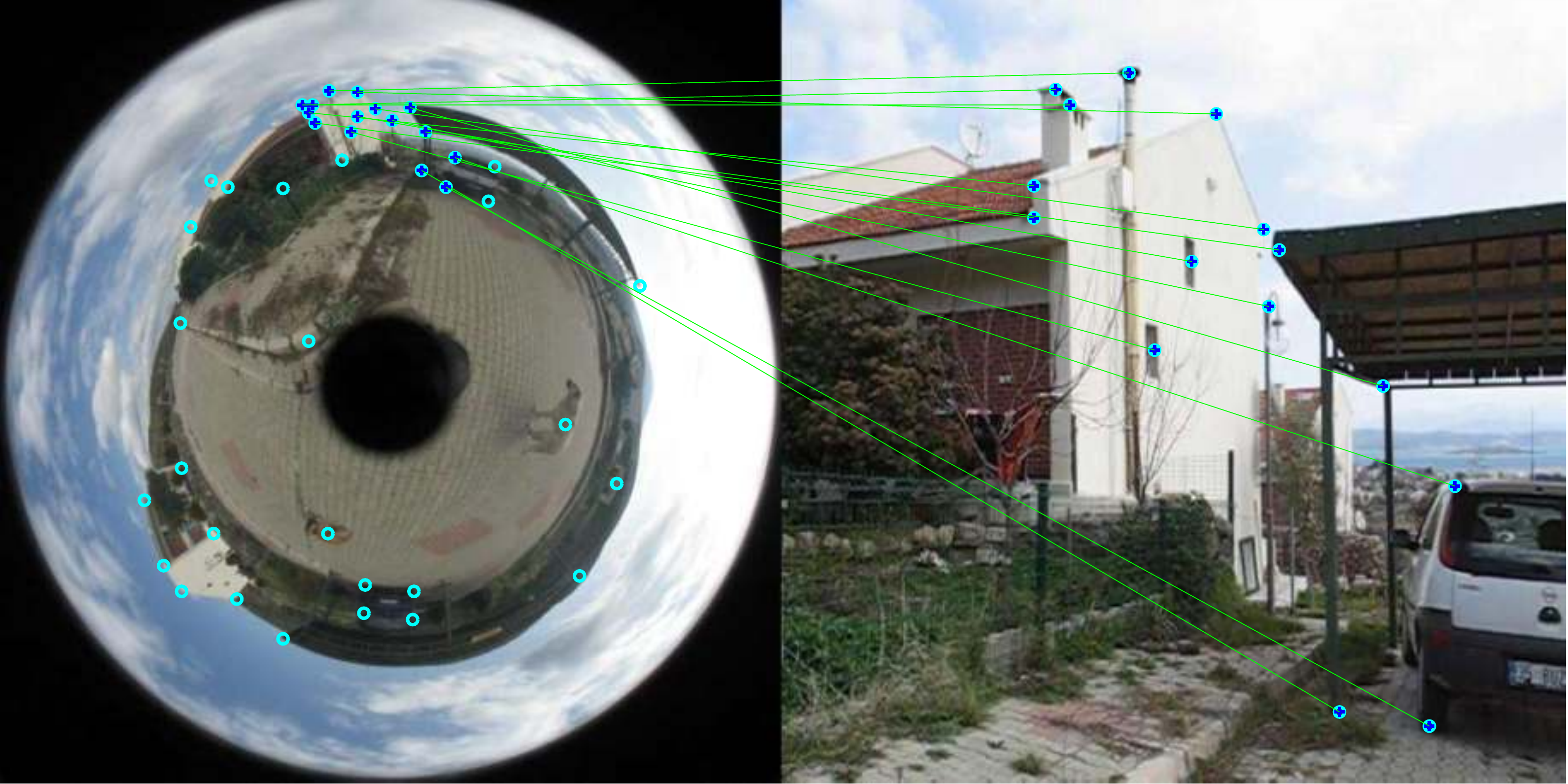}}%
	  \qquad
	  \subfigure[]{%
	  \label{fig:HorseShear}%
	    \includegraphics[trim={0 0 0 0},clip,width=55mm,height=28mm]{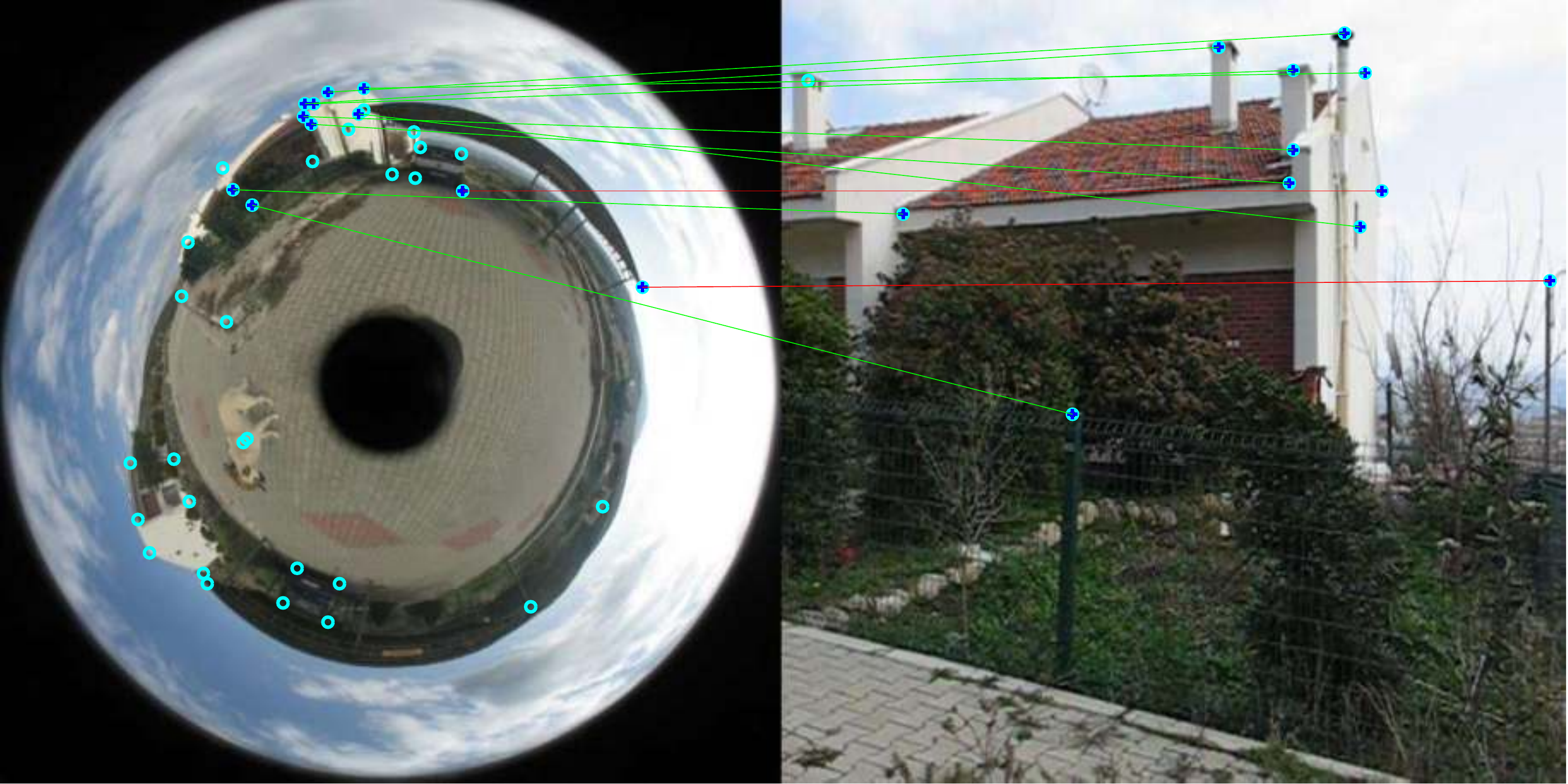}}%
	  }
	  \vspace{-2mm}
	  \qquad
	  \makebox[\linewidth]{
	  \centering
	  \hspace{-17mm}
	  \subfigure[]{%
	  \label{fig:Car}%
	    \includegraphics[width=55mm,height=28mm]{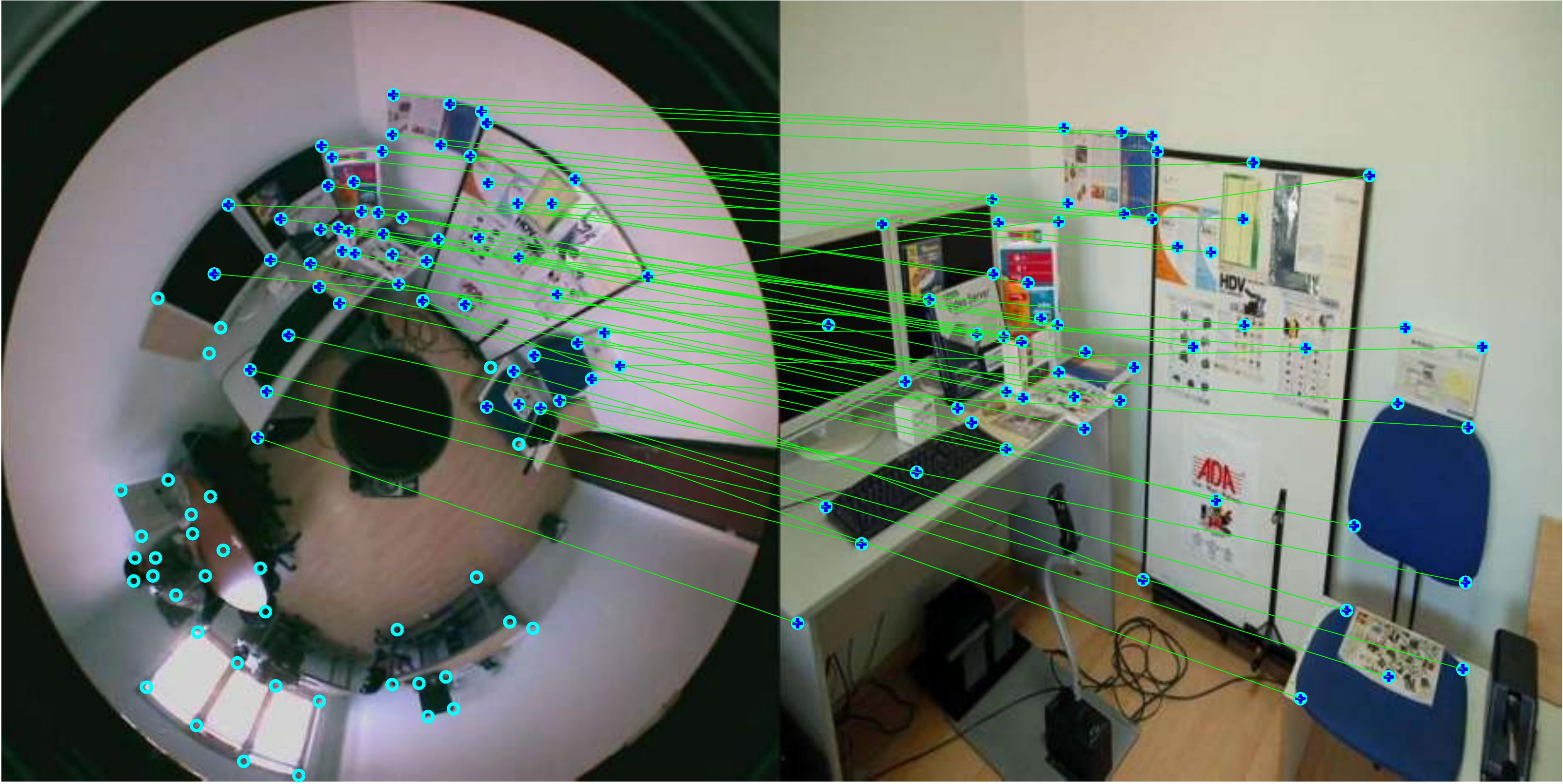}}%
	  \qquad
	  \subfigure[]{%
	  \label{fig:Bike}%
	    \includegraphics[width=55mm,height=28mm]{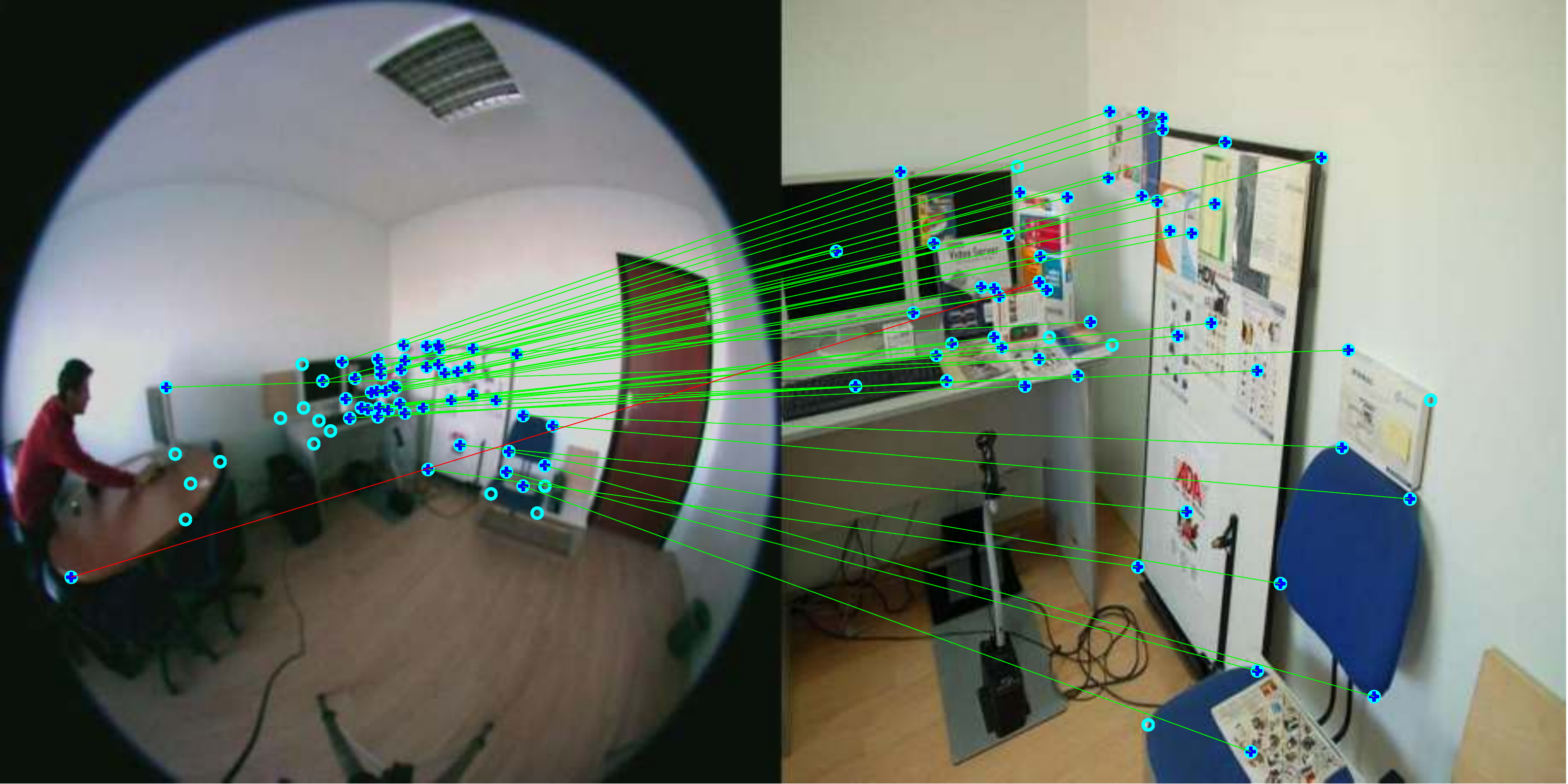}}%
	  }
	  \vspace{-2mm}
	  \qquad
	  \hspace{-9mm}
	  \makebox[\linewidth]{
	  \centering
	  \subfigure[]{%
	  \label{fig:Bldg}%
	    \includegraphics[trim={0 0 0 0},clip,width=55mm,height=28mm]{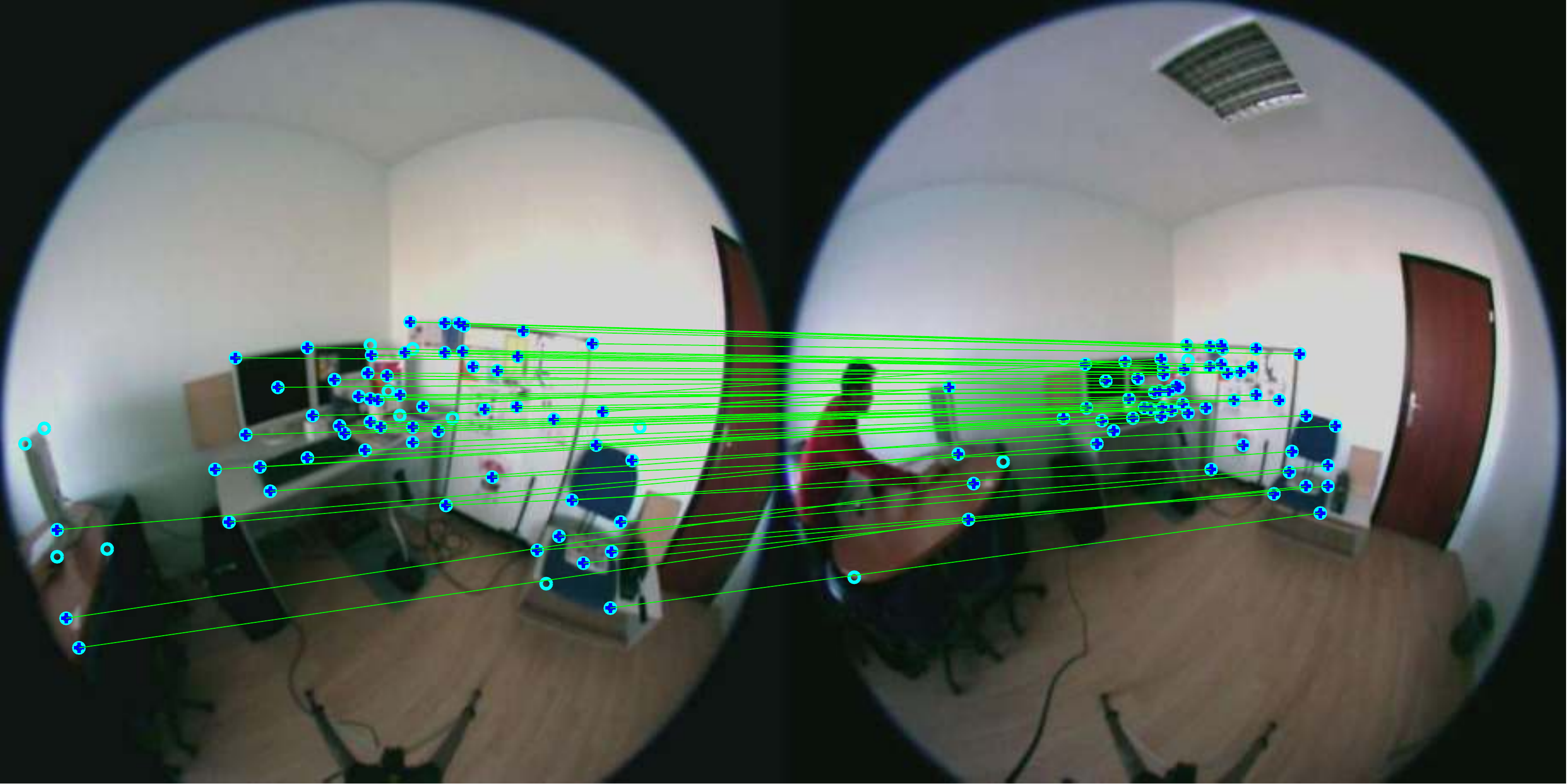}}%
	  \qquad
	  \subfigure[]{%
	  \label{fig:Book}%
	    \includegraphics[trim={0 0 0 0},clip,width=55mm,height=28mm]{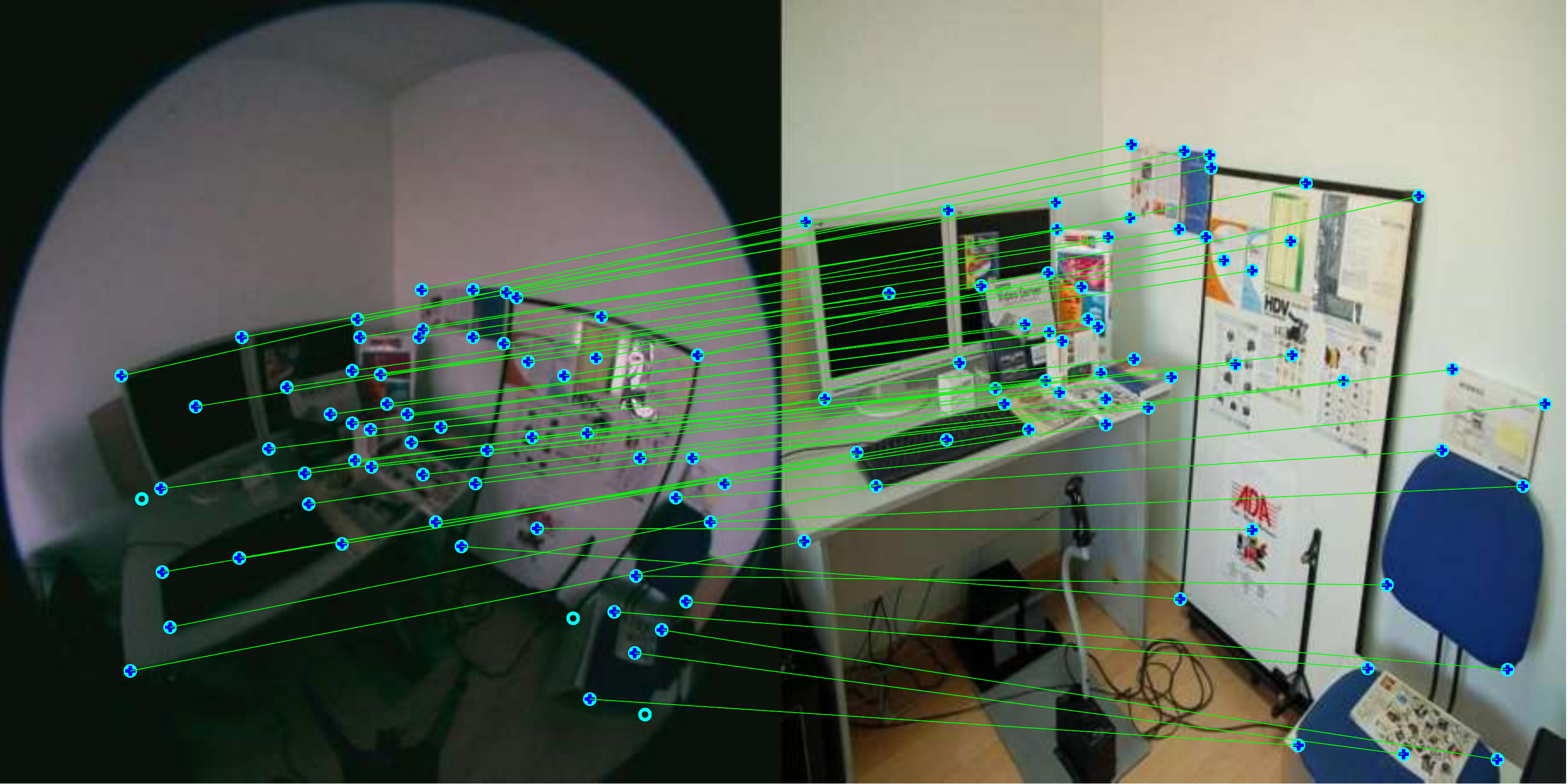}}%
	  }
	  \vspace{-2mm}
	  \qquad
	  \hspace{-9mm}
	  \makebox[\linewidth]{
	  \centering
	  \subfigure[]{%
	  \label{fig:Bldg}%
	    \includegraphics[trim={0 0 0 0},clip,width=55mm,height=28mm]{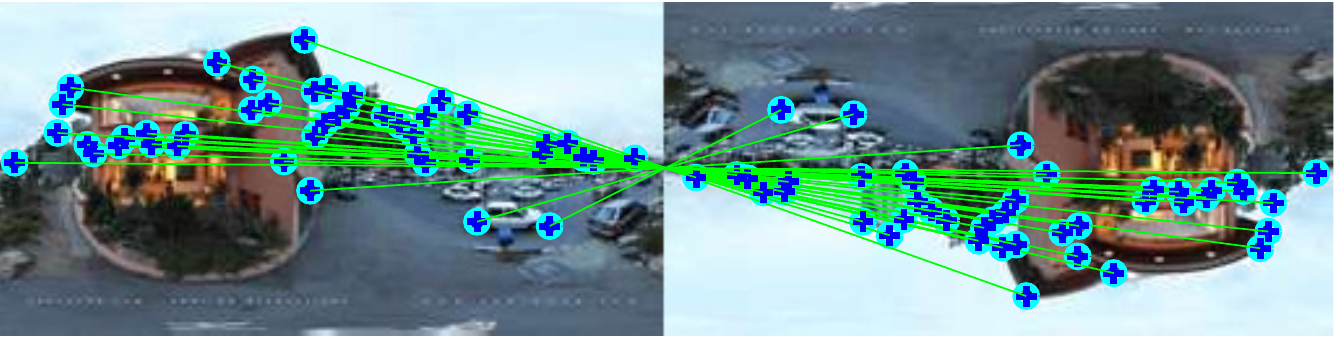}}%
	  \qquad
	  \subfigure[]{%
	  \label{fig:Book}%
	    \includegraphics[trim={0 0 3.5cm 0},clip,width=55mm,height=28mm]{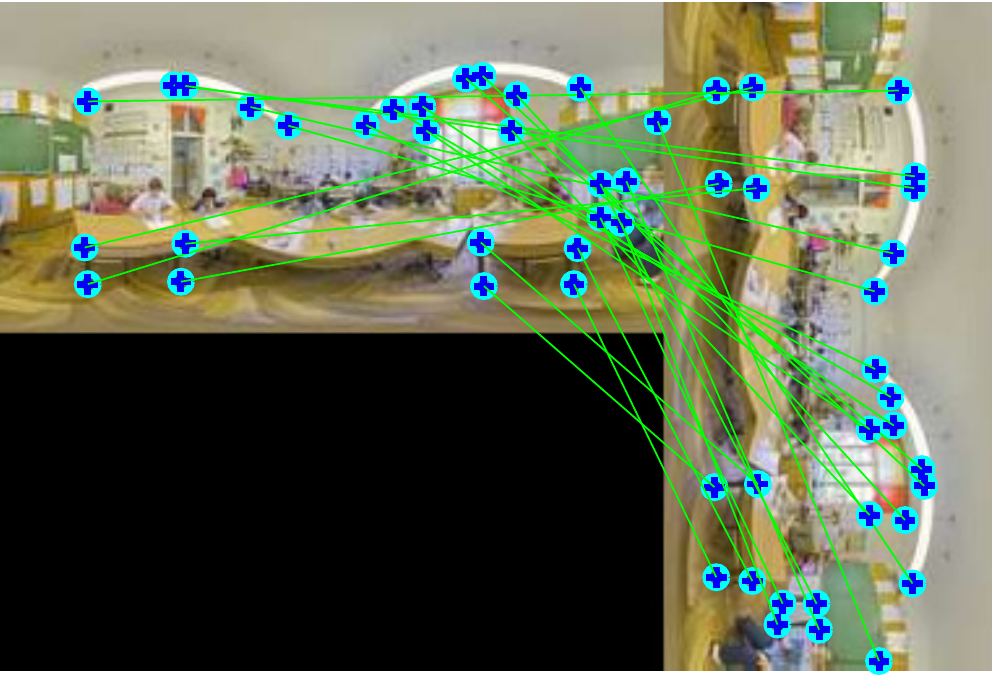}}%
	  }
	  %\vspace{-2mm}
	  \caption{Instances of matchings of spherical-spherical ($(g),(i),(j)$) and spherical-planar ($(a)-(f),(h)$) images of $(a)-(b)$ Desktop, $(c)-(d)$ Parking, $(e)-(h)$ Table and $(i)-(j)$ SUN360 datasets. Green/red lines show correct/incorrect matches respectively and isolated points show no matches.}
	  \label{fig:MatchingSP}
\end{figure*}

\begin{figure*}
	\makebox[\linewidth]{
	\centering
	\vspace{-10mm}
	  \subfigure[]{%
	  \label{fig:KamaishiMatch}%
	    \includegraphics[trim={0 0 0 0},clip,width=55mm,height=30mm]{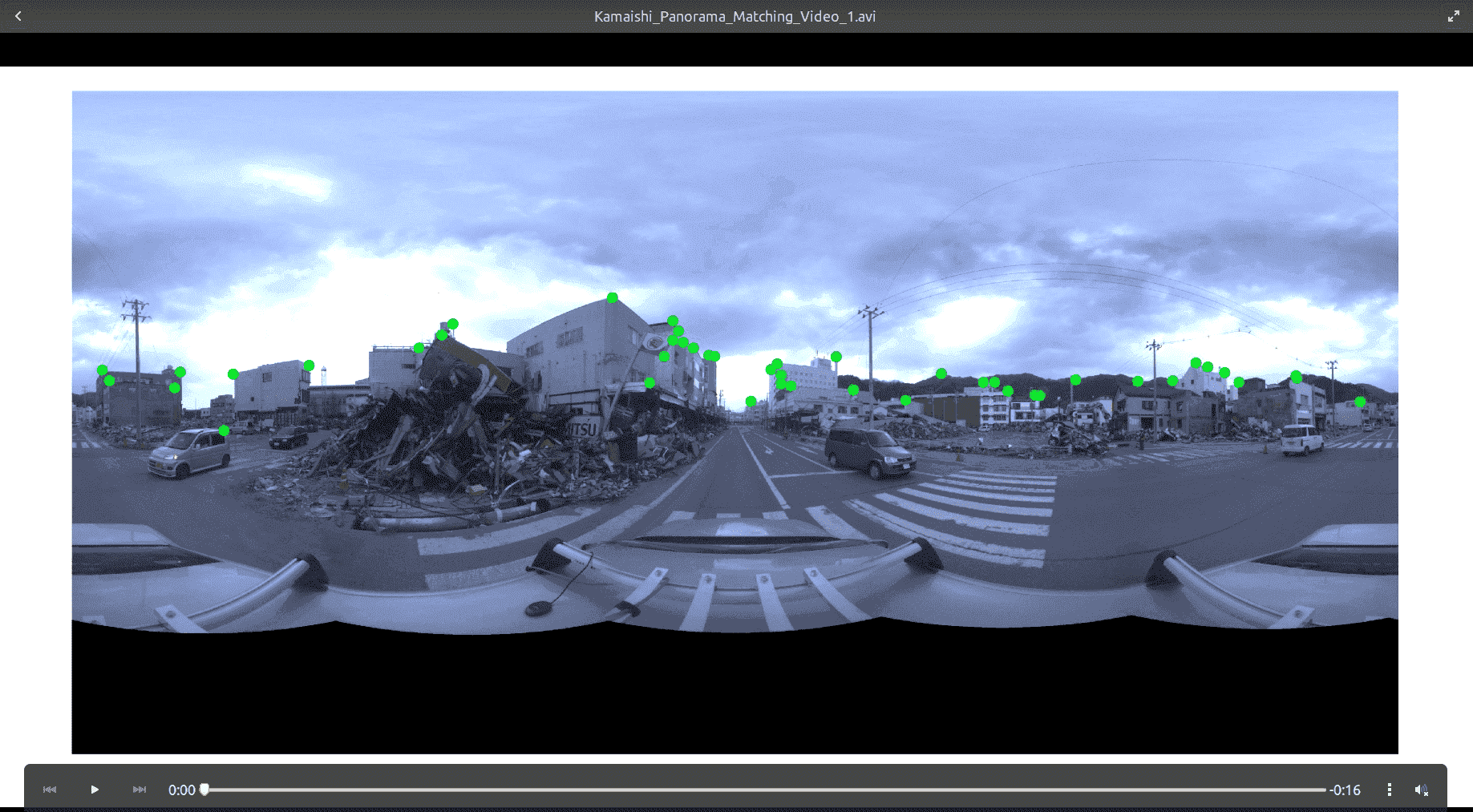}}%
	  %\caption{a) Matchings for house frame $1$ (left) with frame $55$ (right), b) Matchings for house frame $1$ (left) with frame $110$ (right), Error: $0.0\%$.}
	  \qquad
	  %\hspace{0.5mm}
	  \subfigure[]{%
	  \label{fig:KamaishiMatch2}%
	    \includegraphics[trim={0 0 0 0},clip,width=55mm,height=30mm]{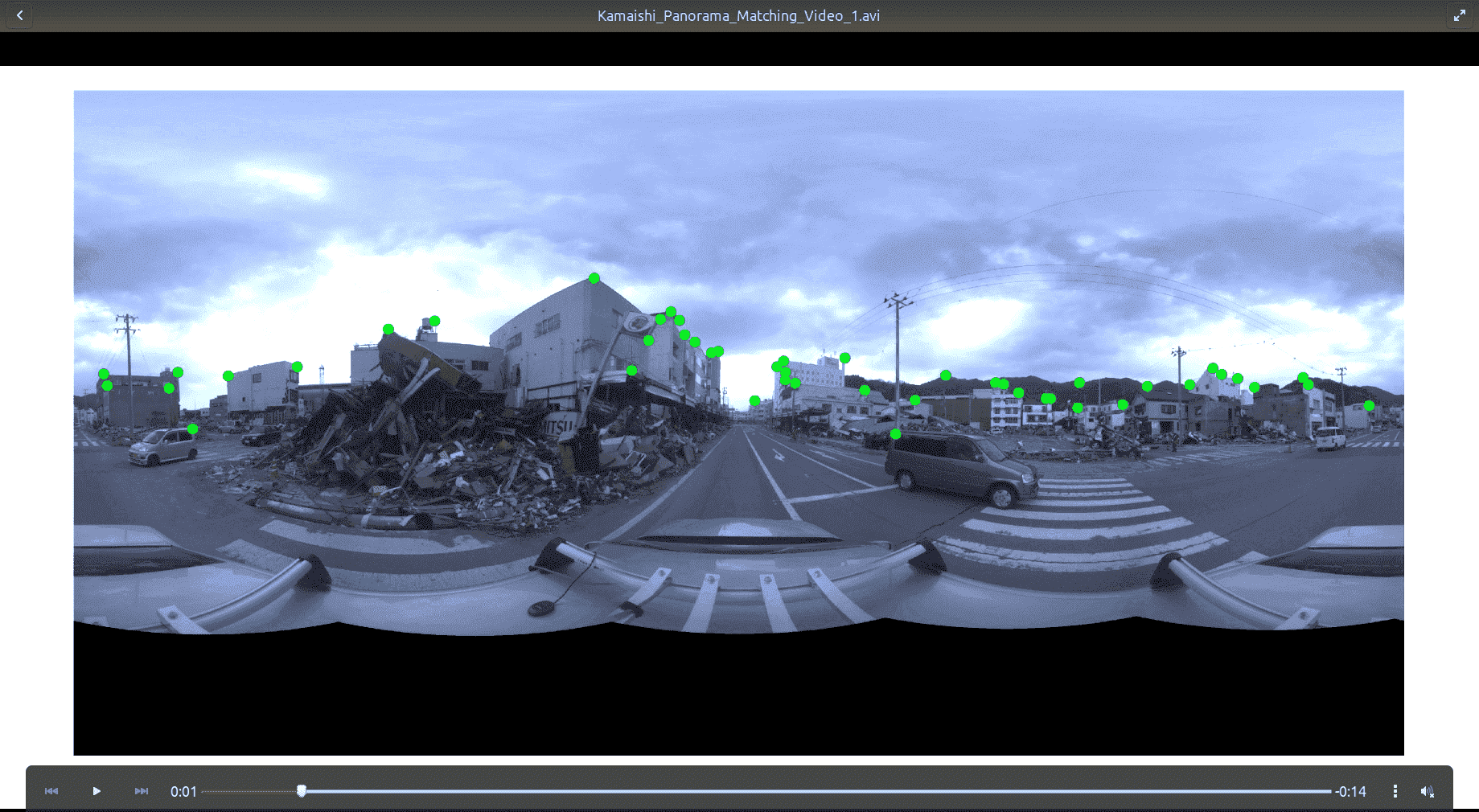}}%
	    %\vspace{-5cm}
	  }
	  \vspace{-2mm}
	  \qquad
	  \makebox[\linewidth]{
	  \centering
	  \hspace{-17mm}
	  \subfigure[]{%
	  \label{fig:HorseRotate}%
	    \includegraphics[trim={0 0 0 0},clip,width=55mm,height=30mm]{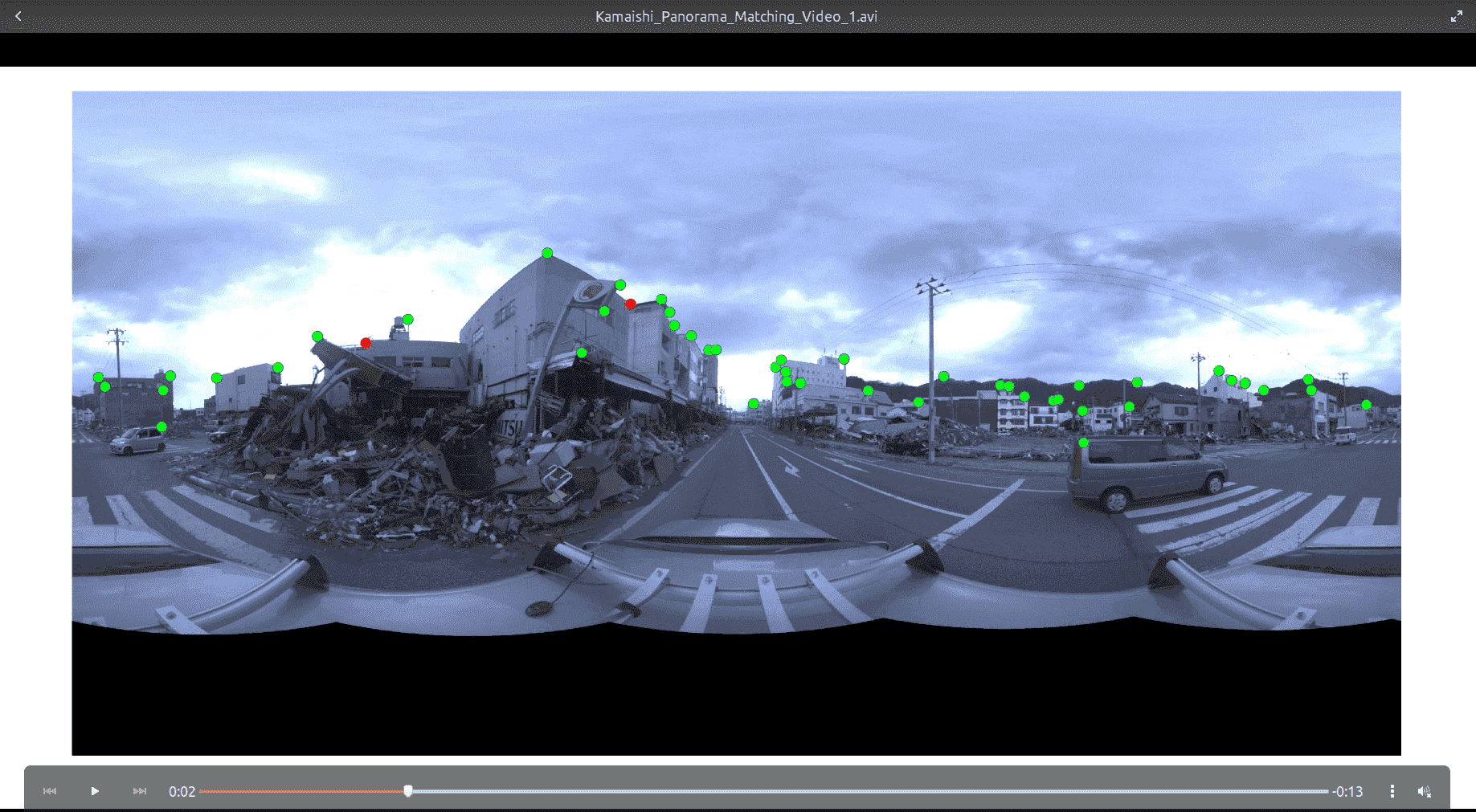}}%
	  \qquad
	  \subfigure[]{%
	  \label{fig:HorseShear}%
	    \includegraphics[trim={0 0 0 0},clip,width=55mm,height=30mm]{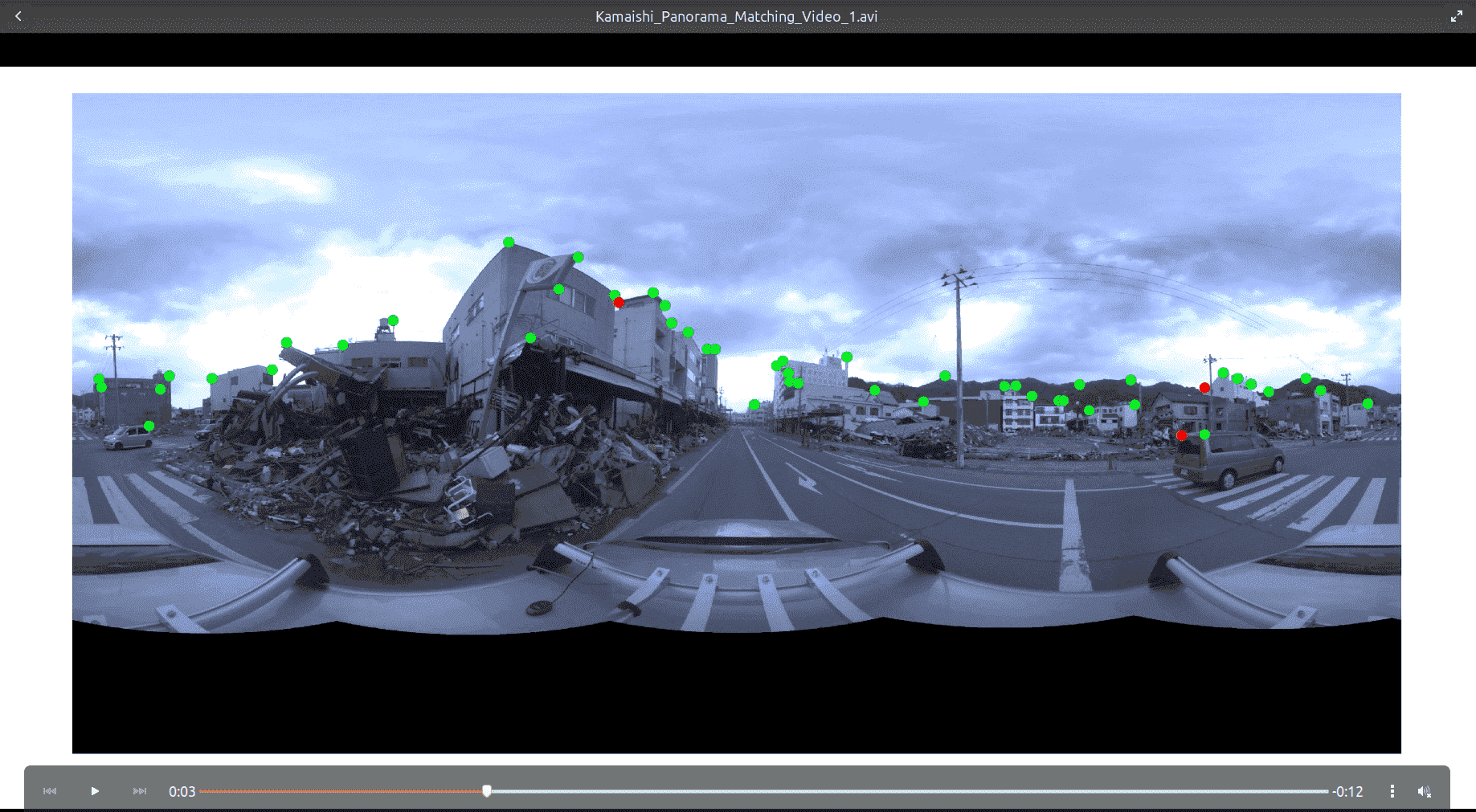}}%
	  }
	  \vspace{-2mm}
	  \qquad
	  \makebox[\linewidth]{
	  \centering
	  \hspace{-17mm}
	  \subfigure[]{%
	  \label{fig:Car}%
	    \includegraphics[width=55mm,height=30mm]{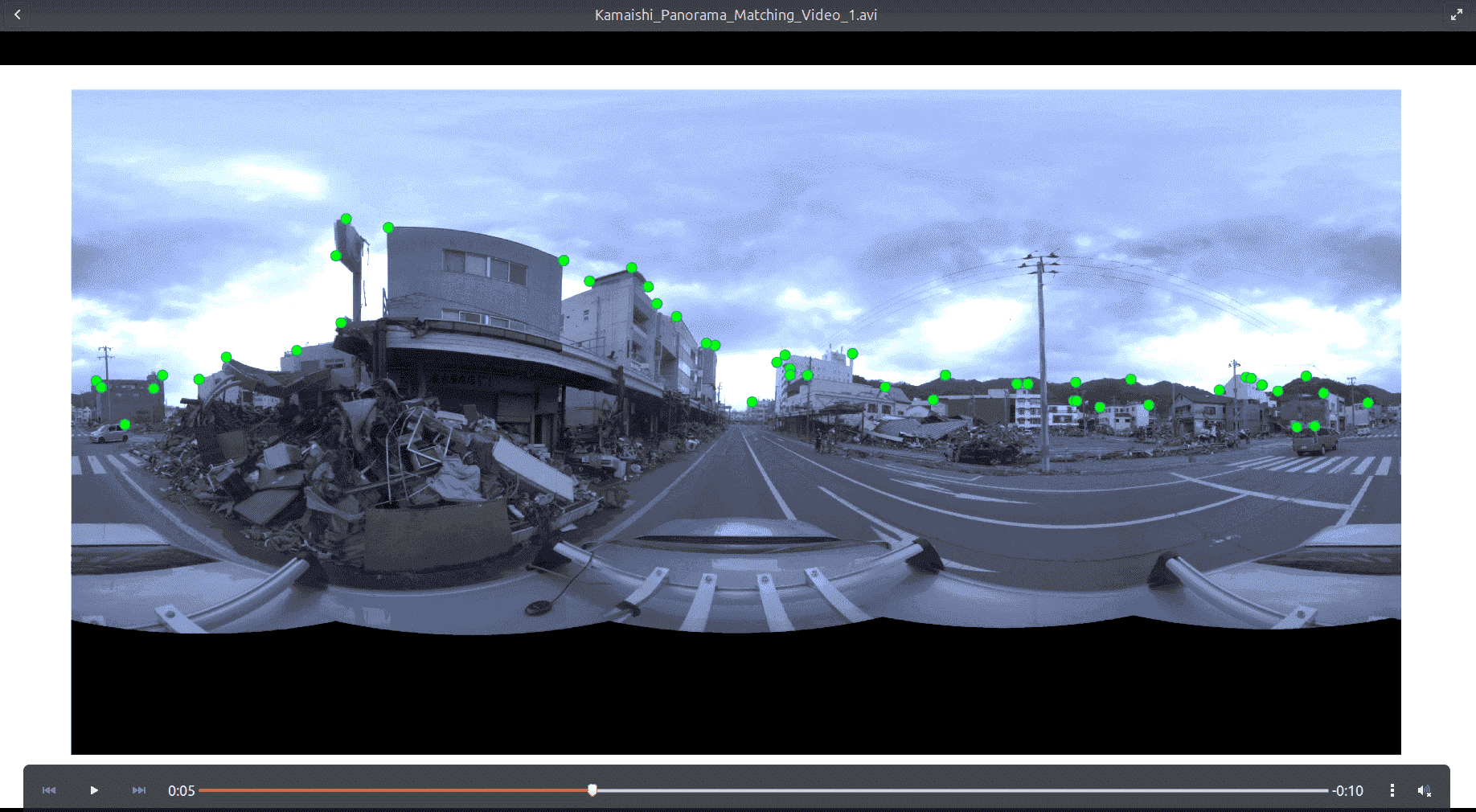}}%
	  \qquad
	  \subfigure[]{%
	  \label{fig:Bike}%
	    \includegraphics[width=55mm,height=30mm]{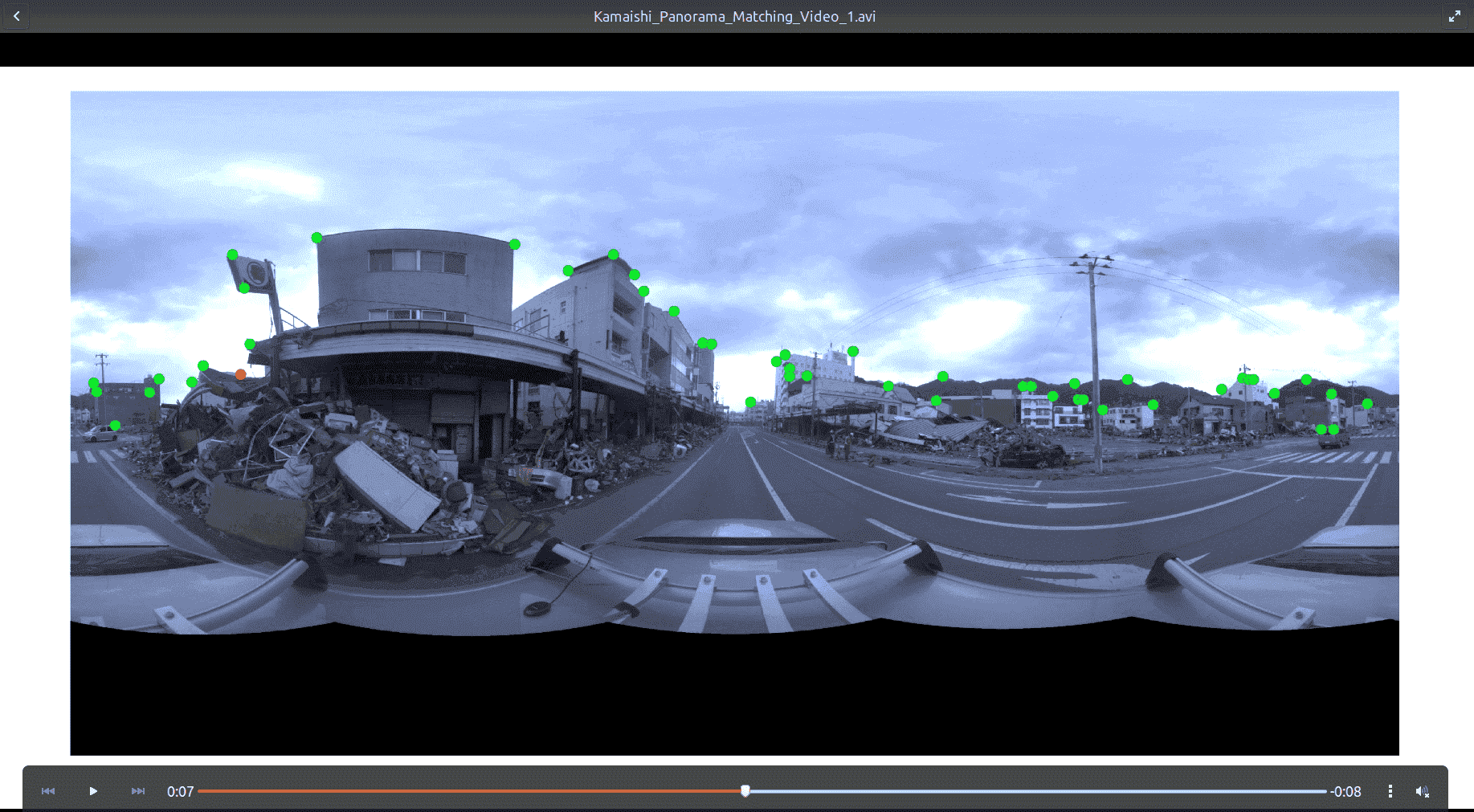}}%
	  }
	  \vspace{-2mm}
	  \qquad
	  \hspace{-9mm}
	  \makebox[\linewidth]{
	  \centering
	  \subfigure[]{%
	  \label{fig:Bldg}%
	    \includegraphics[trim={0 0 0 0},clip,width=55mm,height=30mm]{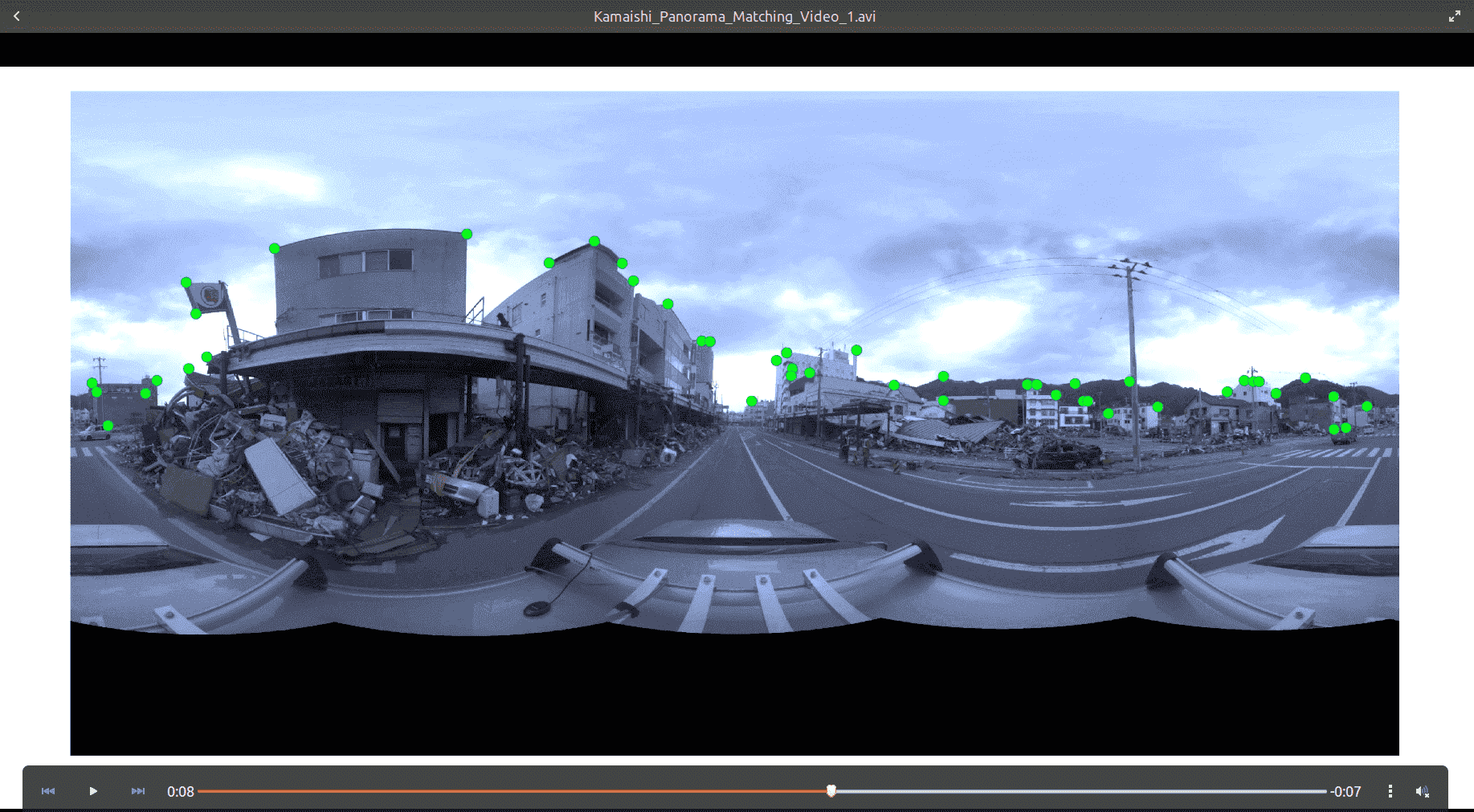}}%
	  \qquad
	  \subfigure[]{%
	  \label{fig:Book}%
	    \includegraphics[trim={0 0 0 0},clip,width=55mm,height=30mm]{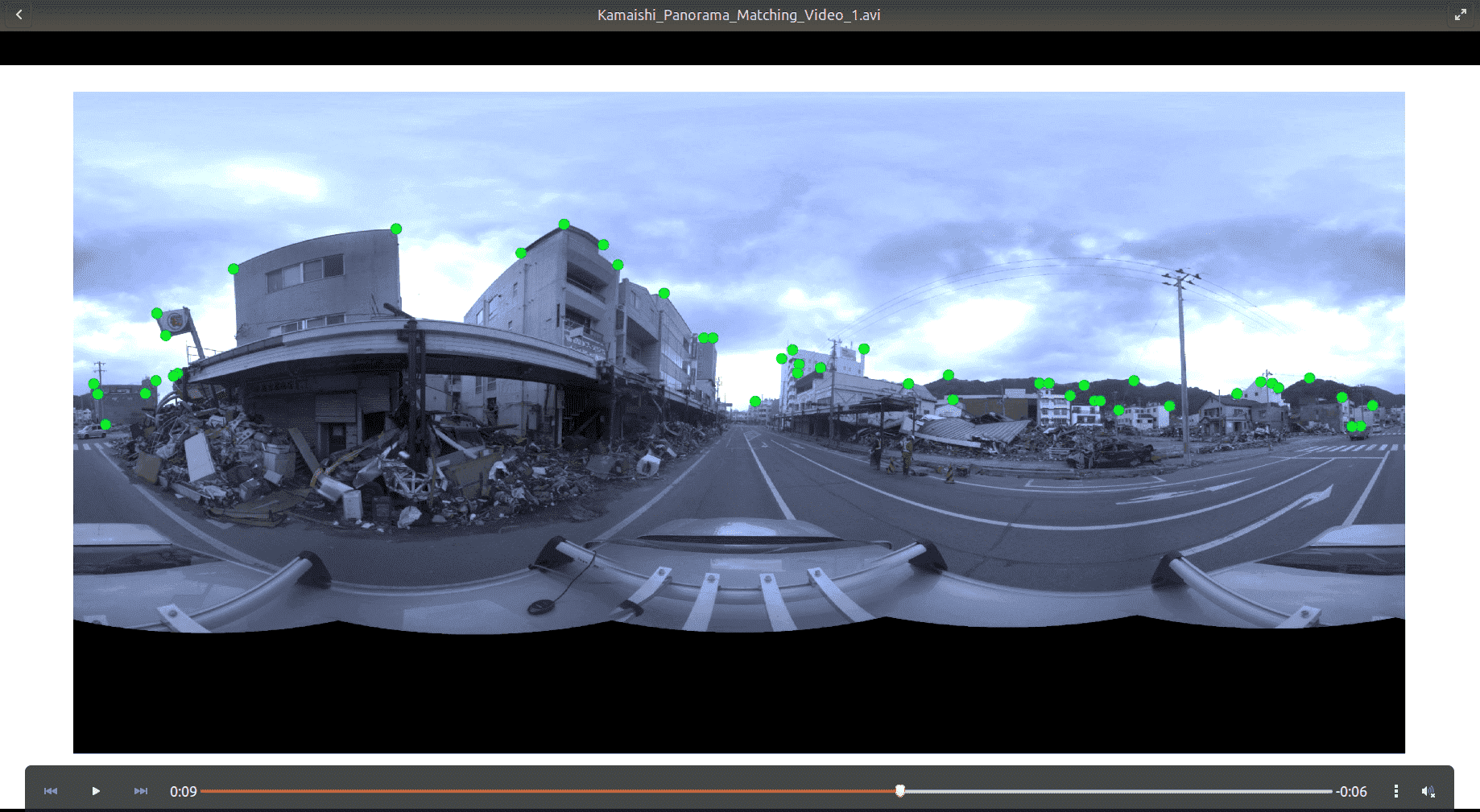}}%
	  }
	  \vspace{-2mm}
	  \qquad
	  \hspace{-9mm}
	  \makebox[\linewidth]{
	  \centering
	  \subfigure[]{%
	  \label{fig:Bldg}%
	    \includegraphics[trim={0 0 0 0},clip,width=55mm,height=30mm]{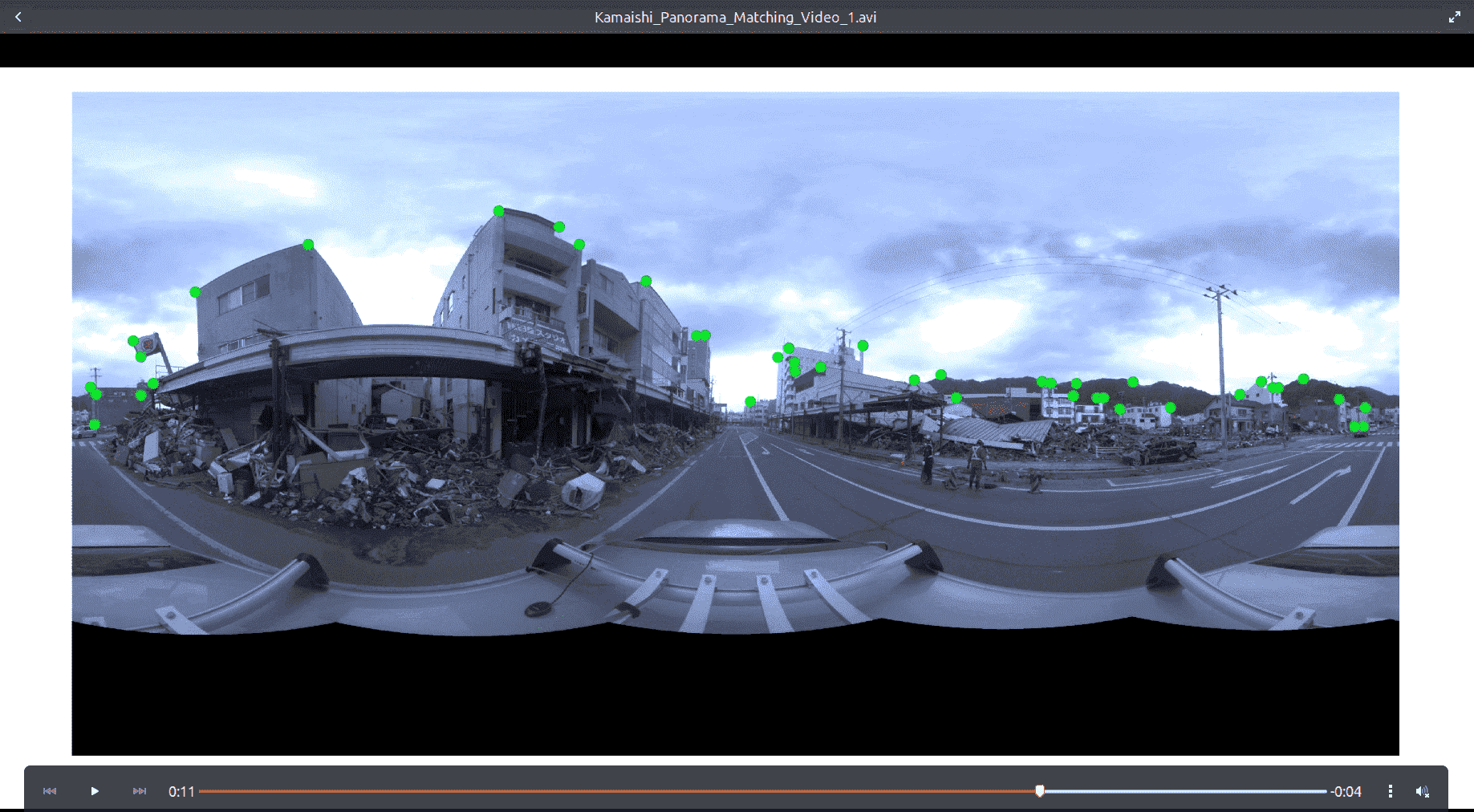}}%
	  \qquad
	  \subfigure[]{%
	  \label{fig:Book}%
	    \includegraphics[trim={0 0 0 0},clip,width=55mm,height=30mm]{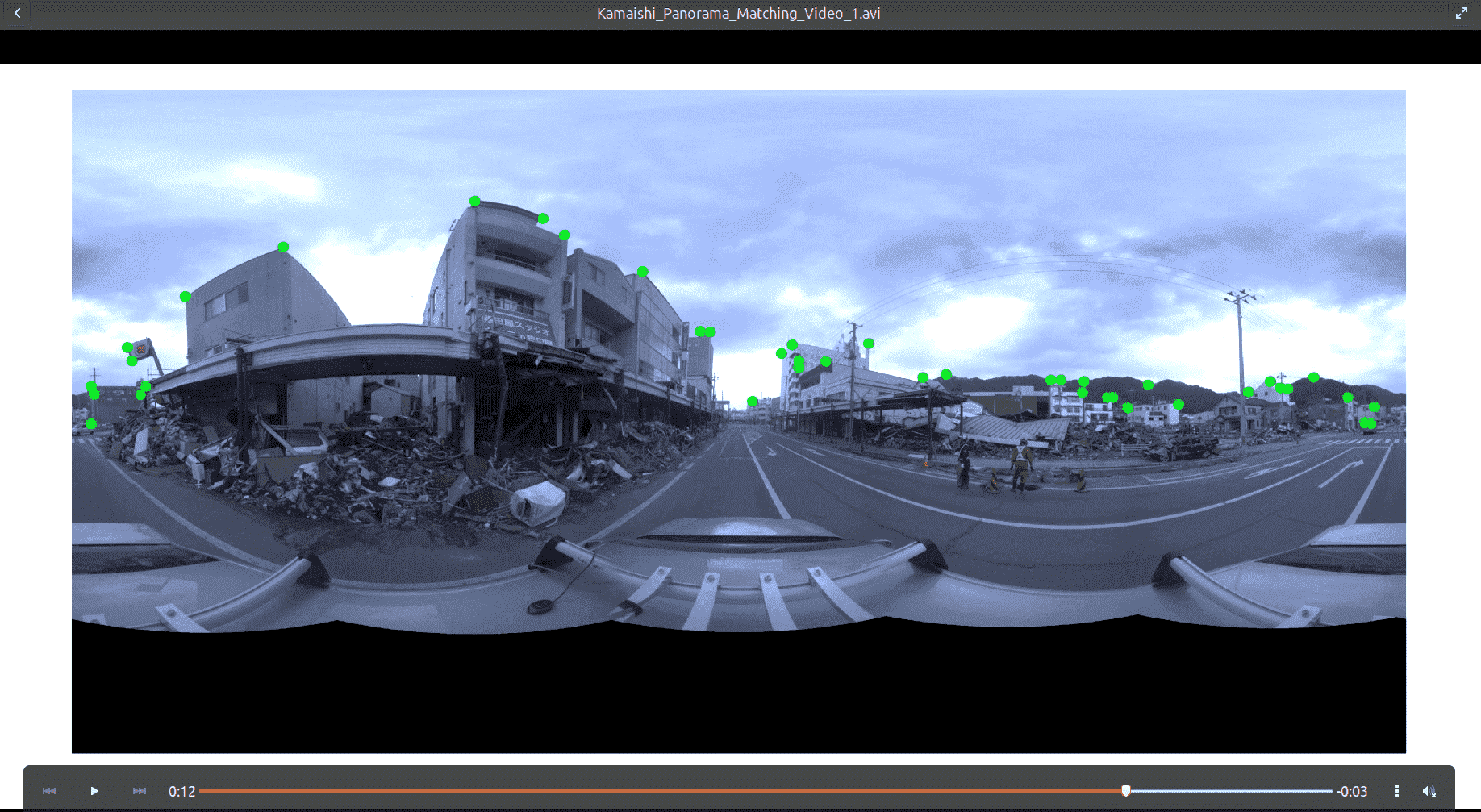}}%
	  }
	  %\vspace{-2mm}
	  \caption{Instances of a video of $Kamaishi$ dataset with correct matches as green landmark points and incorrect or zero matches as red landmark points.}
	  \label{fig:Video}
\end{figure*}

\end{document}